\documentclass[runningheads]{llncs}

\usepackage{eccv}
\usepackage{eccvabbrv}

\usepackage{graphicx}
\usepackage{booktabs}
\usepackage{xcolor}
\definecolor{creamyblue}{RGB}{120,170,210}
\definecolor{maskedcol}{RGB}{230,240,255}
\definecolor{lightblue}{RGB}{220,235,255}
\usepackage{multirow}
\usepackage{wrapfig}
\usepackage{makecell}
\usepackage{colortbl}
\usepackage{microtype}

\usepackage[accsupp]{axessibility}

\usepackage[breaklinks,colorlinks,citecolor=blue]{hyperref}
\usepackage{orcidlink}

\begin{document}

\title{Temporally Aware Densification for Dynamic 3D Gaussian Splatting}

\author{Vikram Sandu\inst{1}\orcidlink{0009-0006-6512-3695} \and
Mayurdeep Pathak\inst{1}\orcidlink{0000-0001-9883-7738} \and
Rajiv Soundararajan\inst{1}\orcidlink{0000-0001-5767-5373}}

\authorrunning{V. Sandu et al.}

\institute{
Indian Institute of Science, Bengaluru, India\\
\email{\{vikramsandu,mayurdeepp,rajivs\}@iisc.ac.in}
}

    \maketitle
    \vspace{-15pt}
    \begin{figure}[htbp]
    \centering
    \begin{subfigure}[t]{0.46\linewidth}
        \centering
        \includegraphics[
            width=\linewidth,
            height=0.25\textheight,
            keepaspectratio
        ]{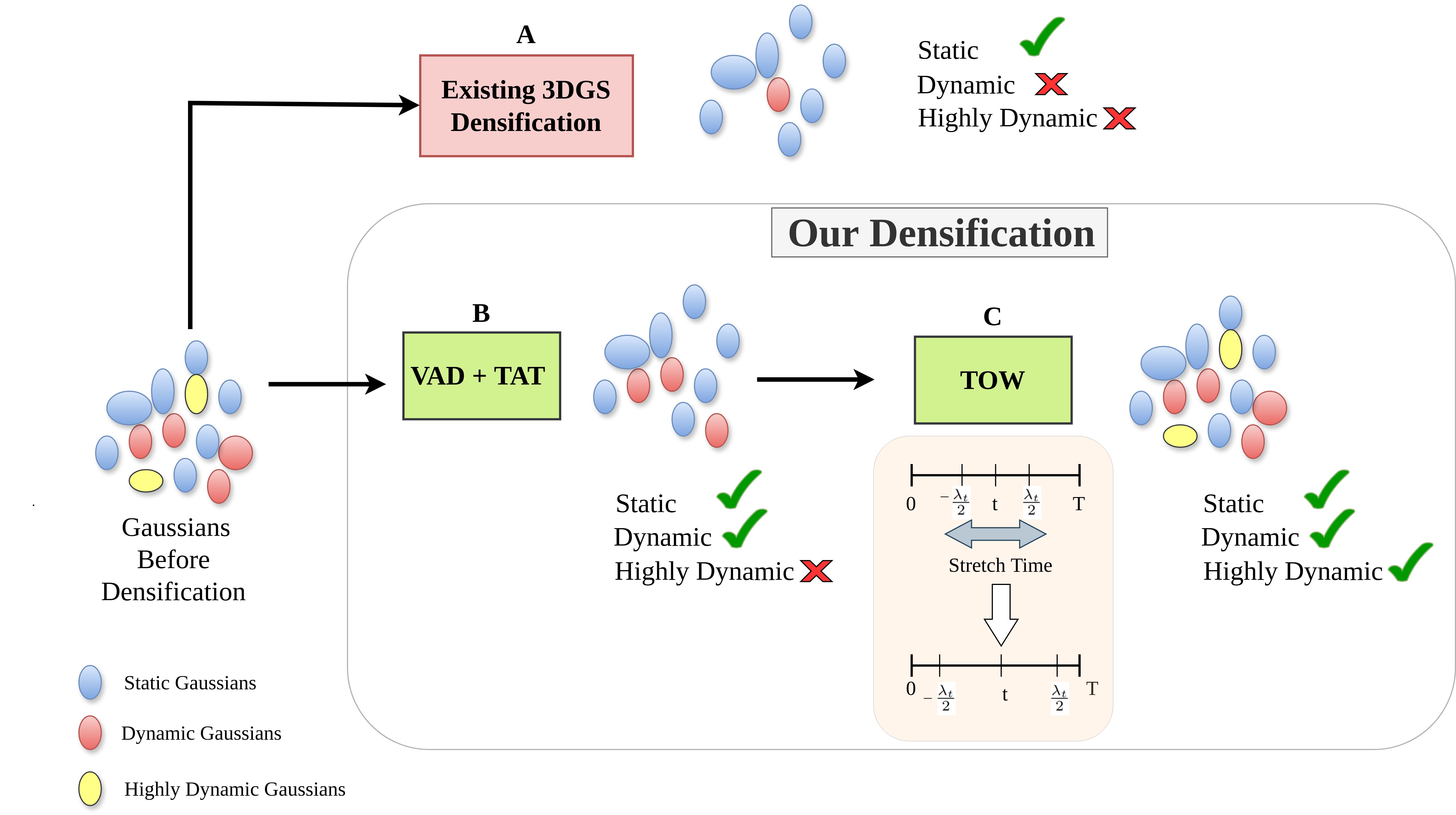}
    \end{subfigure}
    \hfill
    \begin{subfigure}[t]{0.50\linewidth}
        \centering
        \includegraphics[
            width=\linewidth,
            height=0.28\textheight,
            keepaspectratio
        ]{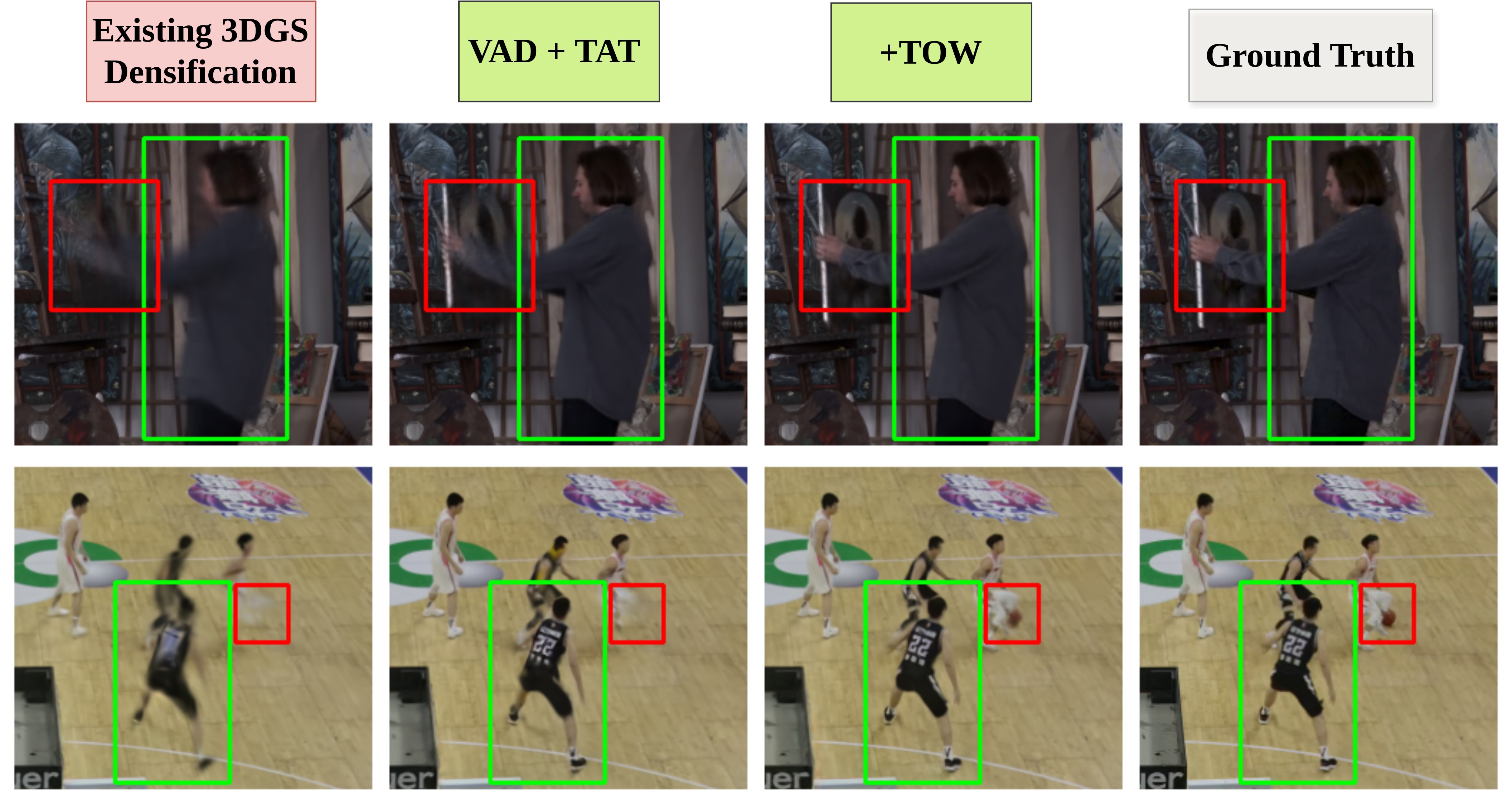}
    \end{subfigure}

    \vspace{-5pt}
    \caption{
    \textbf{Left)}
    (A) Existing 3DGS densification fails to densify short-lived dynamic Gaussians.
    (B) VAD with visibility-weighted gradients and TAT with lifespan-aware thresholds enable densification of short-lived dynamic Gaussians.
    (C) TOW warps temporal coordinates around center $t$ within window $\lambda_t$, boosting deformation capacity and densifying highly dynamic Gaussians.
    \textbf{Right)}
    Improvement over the densification baseline as VAD, TAT, and TOW are incrementally incorporated, resulting in sharper and more consistent dynamic reconstructions. VAD and TAT enhance moderately dynamic regions shown in \textcolor{green}{green} box, while TOW further improves reconstruction in highly dynamic regions such as the hand (top image) and basketball (bottom image) as shown in \textcolor{red}{red} box.
    }
    \label{fig:teaser}
    \vspace{-30pt}
\end{figure}

\begin{abstract}
  Despite modeling temporal motion, dynamic 3D Gaussian Splatting (3DGS) methods still inherit a static densification strategy ill-suited for dynamic scenes. This neglect of temporal behavior leads to under-reconstructed and blurry dynamic regions, as short-lived Gaussians receive sparse supervision and fail to densify effectively. We propose a \textit{Visibility-Aware Densification} (VAD) framework that integrates temporal visibility into the densification process, ensuring that Gaussians are refined based on their actual temporal presence. A \textit{Temporally-Adaptive Thresholding} (TAT) mechanism further adjusts each Gaussian's densification threshold according to its temporal lifespan, promoting balanced refinement of both static and dynamic regions. Finally, a \textit{Temporal Offset Warping} (TOW) design enhances deformation capacity around temporal centers, extending the lifespan of highly dynamic Gaussians and facilitating more effective densification. Our approach achieves substantial improvements in the visual quality of dynamic regions, outperforming existing methods across three dynamic multi-view benchmark datasets. Moreover, the proposed VAD module generalizes across diverse dynamic 3DGS methods, consistently improving dynamic reconstruction as a plug-and-play component.
\end{abstract}

\section{Introduction}
\label{sec:intro}

    Novel view synthesis (NVS) is a fundamental problem in 3D vision that aims to generate realistic views of a scene from unseen viewpoints by reasoning about its underlying geometry, appearance, and illumination, making it essential to modern 3D vision, graphics, and robotics. Since the advent of Neural Radiance Fields (NeRF) \cite{mildenhall2020nerf}, view synthesis has witnessed rapid progress, with numerous follow-up works \cite{yu2022plenoxels, barron2021mip, SunSC22, Chen2022ECCV, somraj2023VipNeRF, somraj2023simplenerf, mueller2022instant} enhancing reconstruction quality and efficiency, yet still constrained by slow rendering. More recently, 3D Gaussian Splatting (3DGS) \cite{kerbl3Dgaussians} has emerged as a compelling alternative, enabling real-time rendering while maintaining photorealistic fidelity. Building upon this foundation, several approaches have extended 3DGS to dynamic scenes - by introducing deformation fields \cite{Wu_2024_CVPR, li2023spacetime, lin2024gaussianflow}, modeling Gaussians directly in 4D space \cite{yang2023gs4d, duan:2024:4drotorgs}, or leveraging temporal keyframe interpolation \cite{lee2024ex4dgs}. While these methods differ in how they represent scene dynamics, they all fundamentally rely on the same densification strategy as the original 3DGS, which was primarily designed for static scenes. Densification enhances scene representation by introducing new Gaussians through splitting or cloning existing ones in regions where reconstruction lacks detail. Gaussians are selected for densification when their accumulated positional gradients, averaged over a predefined interval, exceed a fixed threshold.

    However, we observe that the existing densification strategy overlooks the temporal behavior of Gaussians, making it suboptimal for dynamic 3DGS frameworks and often leading to under-reconstructed or blurry dynamic regions in rendered views. Notably, dynamic Gaussians typically exhibit shorter temporal lifespans to effectively represent complex scene dynamics \cite{yan20244d, yuan20251000fps4dgaussian}. Consequently, their visibility, parametrized by the opacity of each Gaussian, is limited to only a few frames. This sparse visibility yields equally sparse supervision and gradient updates during training, which prevents these Gaussians from accumulating the positional gradients required for densification, leaving them permanently under-densified. This motivates a densification strategy that explicitly accounts for temporal visibility.

    In this work, we propose a \textit{visibility-aware densification} (VAD) criterion that integrates Gaussian visibility into the densification process by accumulating visibility-weighted positional gradients and normalizing them by total visibility. This ensures that the effective densification signal for each Gaussian is primarily influenced by the frames in which it is actually visible. To further promote the densification of short-lived Gaussians, we introduce a \textit{temporally-adaptive thresholding} (TAT) mechanism that dynamically adjusts each Gaussian's densification threshold in proportion to its temporal lifespan, facilitating densification of such Gaussians. We further observe that certain Gaussians exhibit extremely short temporal lifespans, typically those representing highly dynamic regions—due to the deformation field's limited ability to model abrupt or complex motion. This results in near-zero sampling probability of such Gaussians during the densification process. To alleviate this issue, we adopt multiple temporal centers, inspired by \cite{li2023spacetime, lee2024ex4dgs}, and propose a \textit{temporal offset warping} (TOW) mechanism that allocates greater deformation capacity near each Gaussian's temporal center. This design enhances the modeling of complex motion and effectively extends the temporal lifespan of highly dynamic Gaussians, thereby improving densification in such regions.

    Together, these components establish a unified framework for visibility-driven Gaussian densification in dynamic 3DGS models as shown in Fig~\ref{fig:teaser}. The key contributions of our work are summarized below.

    \begin{enumerate}
        \item \textit{Visibility-aware densification} criterion that incorporates per-frame temporal visibility into the densification process, seamlessly integrating with existing dynamic 3DGS frameworks to improve reconstruction quality in dynamic regions.
        \item \textit{Temporally-adaptive thresholding} mechanism that adjusts the densification gradient threshold based on each Gaussian's temporal lifespan, enabling effective densification of short-lived Gaussians.
        \item \textit{Temporal offset warping} that allocates greater deformation capacity near temporal centers, enhancing the modeling of complex motions and effectively improving densification in highly dynamic regions.
    \end{enumerate}

\section{Related Work}
\label{sec:related_work}

    \textbf{Static 3DGS.}
    Since the introduction of 3DGS~\cite{kerbl3Dgaussians}, numerous extensions have aimed to enhance its rendering quality, generalization, and efficiency across diverse scenarios~\cite{Huang2DGS2024, zhang2024pixelgs, zhu2024fsgs, zhang2024cor, park2025dropgaussian}. These works primarily focus on improving static scene representations through better appearance modeling, adaptive Gaussian management, or geometric regularization.

    \noindent\textbf{Dynamic 3DGS.}
    To handle dynamic scenes, several approaches extend 3DGS by incorporating temporal modeling~\cite{Wu_2024_CVPR, yang2023gs4d, li2023spacetime, lee2024ex4dgs, lin2024gaussianflow, stearns2024marbles, yang2023deformable3dgs, shaw2024swings}. Existing methods can be broadly categorized as:
    (1) 4D Gaussian representations, and
    (2) deformable 3D Gaussians.
    4D-based methods, such as 4DGS \cite{yang2023gs4d} and 4DRotorGS \cite{duan:2024:4drotorgs}, represent spatiotemporal variations directly in a unified 4D space. 4DGS leverages 4D primitives and spherical harmonics to jointly encode view- and time-dependent appearance, whereas 4DRotorGS models frame-wise dynamics through temporally varying Gaussian rotations.

    Deformable Gaussian methods, on the other hand, maintain a canonical 3D representation and learn a deformation field to capture motion over time. Examples include 4DGaussians~\cite{Wu_2024_CVPR}, which employs multi-resolution voxel planes for deformation; STG~\cite{li2023spacetime}, which parameterizes trajectories with low-order polynomials; SplineGS~\cite{Park_2025_CVPR} uses B-spline instead for motion trajectories and GaussianFlow~\cite{lin2024gaussianflow}, which uses a dual deformation field combining polynomial and Fourier representations. Deformable-3DGS~\cite{yang2023deformable3dgs} and SWinGS~\cite{shaw2024swings} adopt MLP-based temporal deformation for motion modeling. More recently, Ex-4DGS~\cite{lee2024ex4dgs} employs keyframe interpolation for dynamic modeling. Other approaches, such as Swift4D~\cite{wu2025swift4d} and LongVolCap~\cite{xu2024longvolcap}, focus on improving computational efficiency and scalability for long-term dynamic reconstruction.

    \noindent\textbf{Densification.} A substantial amount of work has been done on densification of 3D Gaussians for static scenarios. Kim et al.~\cite{Kim_2024_CVPR} introduce additional color gradients into the densification process to achieve more efficient scene modeling. PGD-GS~\cite{10889028} progressively adjusts the densification threshold to improve reconstruction in sparse input settings. Revising Densification~\cite{rota2024revising} proposes error-based densification, which distributes pixel-level errors to individual Gaussians and decides densification accordingly. Pixel-GS~\cite{zhang2024pixelgs} weighs positional gradients by pixel coverage area to achieve better reconstruction in regions with poor initialization. Grubert et al.~\cite{Grubert_2025} employ exponential scheduling of a predefined gradient threshold for faster convergence and improved densification in complex regions. HDA-GS~\cite{10887967} classifies Gaussians into multiple categories based on their size and applies different densification strategies to each. More recently, AD-GS~\cite{Patle2025ADGS} presents a regulated densification process that alternates between high and low densification phases to improve reconstruction under sparse-view settings.

    However, only a limited number of works have addressed densification in dynamic 3DGS. SaRO-GS~\cite{yan20244d} identifies the issue of short-lived Gaussians and introduces an adaptive optimization scheme that moderates each Gaussian's learning rate and densification threshold based on a temporally integrated state function to improve reconstruction in dynamic regions. STG~\cite{li2023spacetime} employs guided sampling by introducing Gaussians in regions with high reconstruction error to enhance scene reconstruction. More recently, Anchored-4DGS~\cite{10.1145/3757377.3763898} guides densification in dynamic regions by adapting the scaling threshold according to each Gaussian's temporal coverage to optimize anchors for efficient storage. While these methods offer improvements, they do not fully address the fundamental mismatch between static densification strategies and dynamic scene requirements. The core issue of temporal visibility bias in gradient accumulation remains unexamined, and existing solutions lack a unified framework that jointly optimizes densification criteria, threshold adaptation, and deformation capacity allocation.

    In contrast, our work provides a comprehensive analysis of densification challenges in dynamic settings and proposes a detailed solution along with a generalized VAD module that can be seamlessly integrated into any existing dynamic 3DGS framework to improve dynamic reconstruction.

\section{Preliminaries}
\label{sec:preliminaries}

    \textit{3D Gaussian Splatting} (3DGS) represents a scene as a set of Gaussian primitives $\mathcal{G}=\{G_i\}$, where each primitive \( G_i \) is parameterized by
    \(\theta_i = (\mu_i, \Sigma_i, \sigma_i, f_i)\). Here, \(\mu_i \in \mathbb{R}^3\) denotes the 3D position, \(\Sigma_i \in \mathbb{R}^{3\times3}\) the covariance, \(\sigma_i\in [0,1]\) the opacity, and \(f_i\) the spherical harmonics coefficients (or color feature vector). The covariance \(\Sigma_i\) is further decomposed as
    \[
    \Sigma_i = R_i S_i S_i^{\top} R_i^{\top},
    \]
    where \(R_i\) represents the rotation and \(S_i \in \mathbb{R}^{3 \times 3}\) is a diagonal scale matrix. This decomposition ensures that \(\Sigma_i\) remains positive semi-definite, allowing for stable and smooth optimization during training.

    For rendering, the Gaussians are first sorted in front-to-back order, and the final color at pixel $p$ in camera plane $\pi$ is computed as
    \begin{equation}
    f_{p} = \sum_{i \in \mathcal{G}_p} \phi(f_i, \mathbf{d}_\pi) \, \alpha_i^{2D} \prod_{j \in \mathcal{G}_p, j < i} \left( 1 - \alpha_j^{2D} \right),
    \end{equation}
    where $\mathcal{G}_p$ is the set of Gaussians influencing pixel $p$, and
    \begin{equation}
    \alpha_i^{2D} = \sigma_i \, G_i^\pi(p),
    \end{equation}
    with $G_i^\pi(p)$ denoting the value of the 2D Gaussian function at pixel $p$, derived from its 3D mean $\mu_i$ and covariance $\Sigma_i$ under camera projection. The function \(\phi(f_i, \mathbf{d}_\pi)\) maps the spherical harmonics coefficients $f_i$ of Gaussian $i$ to an RGB color along the viewing direction $\mathbf{d}_\pi$.

   \begin{wrapfigure}{r}{0.5\textwidth}
        \centering
        \vspace{-25pt}
        \includegraphics[width=\linewidth]{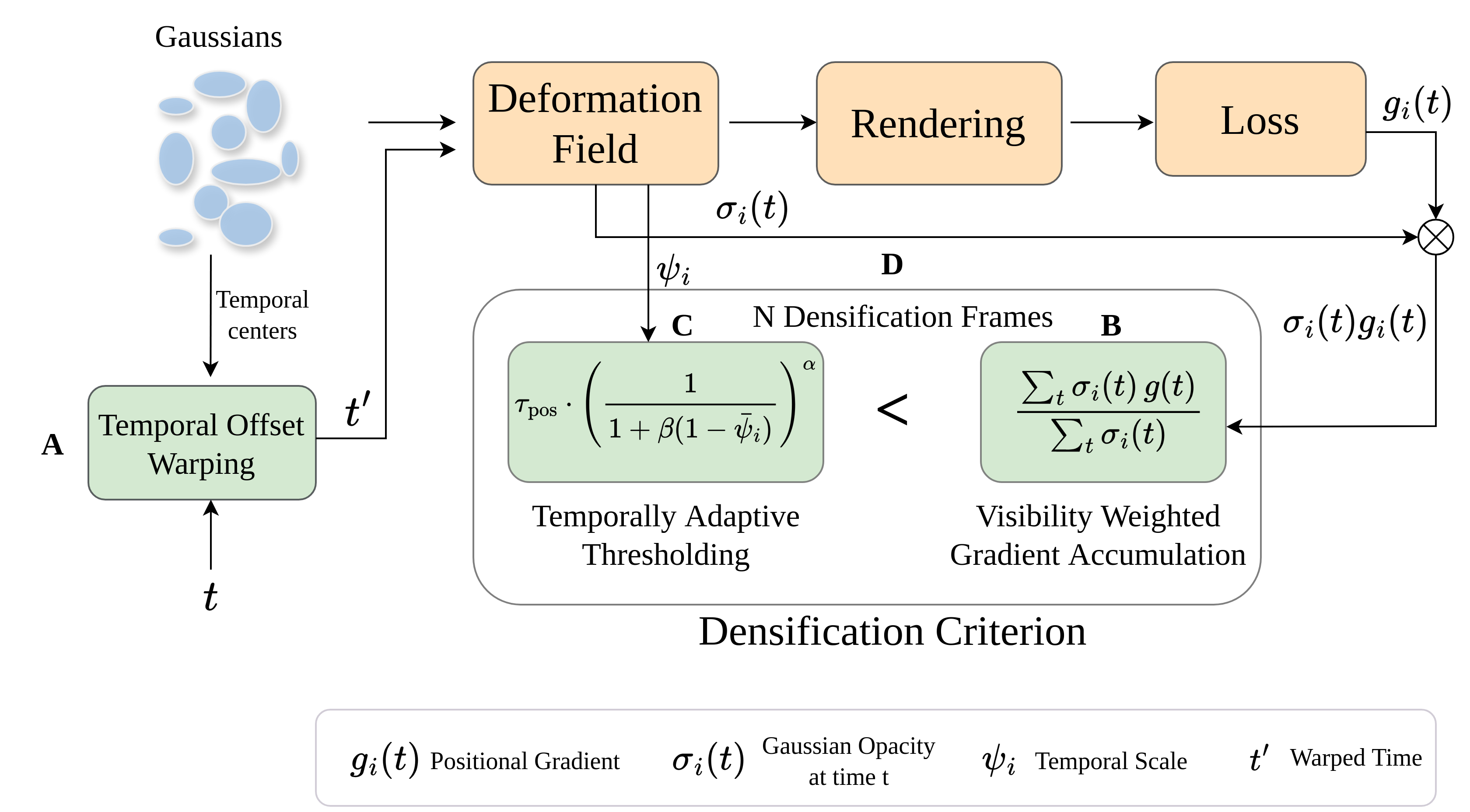}
        \vspace{-15pt}
        \caption{\textbf{Overview of our proposed densification framework.}
        (A) Temporal Offset Warping adaptively warps the input time around each Gaussian's temporal center, stretching regions near the center and compressing those farther away.
        (B) We accumulate the visibility weighted gradient signal for the $N$ densification frames.
        (C) We dynamically adjust the densification threshold based on temporal scale $\psi$ averaged over the $N$ densification frames.
        (D) A Gaussian is densified when its visibility-weighted gradient accumulation exceeds the temporally adaptive threshold.}
        \vspace{-20pt}
        \label{fig:method-diagram}
        \vspace{5pt}
    \end{wrapfigure}

    \noindent\textit{Densification} is a key mechanism in 3DGS that introduces new Gaussians in under-reconstructed regions, typically by splitting or cloning existing ones. Regions requiring densification are identified using the screen-space positional gradient, defined as $g_i = ||\nabla_{\mu_i^{2D}} L||_2$, where $L$ denotes the photometric reconstruction loss and $\mu_i^{2D}$ is the 2D projection of $\mu_i$. Densification is performed at regular intervals of $N$ iterations based on the accumulated gradient signal. Specifically, a Gaussian $G_{i}$ is eligible for densification if
    \begin{equation}
    \frac{1}{N} \sum_{n=1}^{N} g_i^{(n)} > \tau_{\text{pos}},
    \label{eq:3dgs_densification}
    \end{equation}
    where $N$ is set to 100 iterations and $\tau_{\text{pos}}$ to $0.0002$ in the 3DGS implementation.

\section{Method}

    We propose a framework that consists of three critical components designed to enhance the densification process in dynamic 3DGS models as shown in Fig~\ref{fig:method-diagram}, thus improving reconstruction fidelity in dynamic regions:
    \textbf{(i)} VAD,
    \textbf{(ii)} TAT, and
    \textbf{(iii)} TOW.
    The first two components, VAD and TAT, reformulate the existing 3DGS densification strategy by incorporating temporal variations, while the third component, TOW, indirectly facilitates densification by improving deformation modeling, allowing for longer temporal lifespan of Gaussians and supervision from more frames. We first describe the deformation model on which we impose our densification ideas.

    \subsection{Deformation Model}\label{sec:deformationmodel}

    We model the temporal evolution of each Gaussian through a Fourier-based deformation for motion, a temporal radial basis function (RBF) for opacity, and polynomial functions for rotation and scale, while view- and time-dependent color variations are represented using spherical harmonics.

    \noindent\textbf{Temporal RBF for Opacity.}
    We employ a temporal radial basis function, similar to \cite{li2023spacetime}, to model time-dependent opacity. The opacity at time $t$ is defined as
    \begin{equation}
    \sigma_i(t) = \sigma_i^s \, e^{-\psi_i (t - t_i)^2},
    \label{eq:opacity-deform}
    \end{equation}
    where $\sigma_i^s$ denotes the time-independent spatial opacity, and $t_i$ and $\psi_i$ represent the temporal center and temporal scale of the Gaussian, corresponding to the time instant of maximum visibility and its effective temporal lifespan, respectively. This formulation naturally provides the temporal lifespan of each Gaussian, which is later utilized in Section~\ref{sec:tat} for temporally adaptive thresholding.

    \vspace{1em}
    \noindent\textbf{Fourier-based Motion.}
    Inspired by \cite{lin2024gaussianflow}, for each Gaussian $G_i$, the Gaussian mean at time $t$ is defined as
    \begin{equation}
    \begin{aligned}
    {\mu}_i(t) &= \mu_i +
    \sum_{f=1}^{K}
    \Big(
        \mathbf{M}_{i,f}^{(c)} \cos(\Delta t_{i,f})
        +
        \mathbf{M}_{i,f}^{(s)} \sin(\Delta t_{i,f})
    \Big),
    \end{aligned}
    \label{eq:fourier-asym}
    \end{equation}
    where the phase term $\Delta t_{i,f}$ is given by
    \[
    \Delta t_{i,f} = 2\pi f (t - t_i) + \phi_i.
    \]
    Here, $\mu_i$ denotes the canonical Gaussian mean. The Fourier coefficients $\mathbf{M}_{i,f}$ model motion as a sum of smooth periodic oscillations, while the phase offsets $\phi_i$ introduce beneficial asymmetry around each Gaussian's temporal center $t_i$, enabling the modeling of complex, non-uniform motion. All parameters $\mu_i$, $\mathbf{M}_{i,f}$, and $\phi_i$ are learnable. We set $K = 4$ in our experiments.

    \vspace{1em}
    \noindent\textbf{Polynomial Rotation and Scale.}
    We parameterize rotation using quaternions and employ a polynomial formulation, following prior works \cite{li2023spacetime, bond2025gaussianvideo}, to represent time-dependent rotation and scale. The quaternion and scale at time $t$ are defined as
    \begin{align}
    q_i(t) &= \sum_{k=0}^{N_q} c_{ik} (t - t_i)^k, \quad
    s_i(t) = \sum_{k=0}^{N_s} d_{ik} (t - t_i)^k,
    \label{eq:rot_scale_deform}
    \end{align}
    where $c_{ik}$ and $d_{ik}$ are learnable coefficients, and $N_q$ and $N_s$ are set to $1$ in our implementation. We use low-degree polynomials for rotation and scale as they typically exhibit smoother temporal variations compared to positional motion.

    Each Gaussian $G_i$ is associated with a temporal center $t_i$, which is fixed at initialization. We initialize Gaussians across multiple temporal centers similar to \cite{li2023spacetime, yan20244d}, offering two key benefits: (i) richer temporal coverage during initialization, and (ii) deformation learning that is localized and more stable, as it is performed relative to each Gaussian's own temporal center. A similar concept is adopted in \cite{lee2024ex4dgs}, where it is referred to as keyframe interpolation. We initialize the temporal centers uniformly, with the spacing empirically set to 20 frames in our implementation.

    \subsection{Visibility-Aware Densification}\label{sec:vad}

    We first identify a key limitation in the existing densification strategy (Eq.~\ref{eq:3dgs_densification}), and subsequently introduce a temporal visibility--aware formulation that promotes the densification of short-lived dynamic Gaussians.

    Consider a Gaussian $G_i$ that is visible in a subset $\mathcal{M}\subset\mathcal{T}$, where $\mathcal{T}$ is the entire set of frames. The cardinality of $\mathcal{M}$, $|\mathcal{M}|=M$. Also $|\mathcal{T}|=T$. Typically,  $M \ll T$, corresponding to a short-lived dynamic Gaussian within the total video duration of $T$ frames. Let $\mathcal{N}\subset\mathcal{T}$ correspond to set of $N$ frames considered for densification. As a result, $G_i$ contributes sparsely in $\mathcal{N}$.
    Let $g_i(t)$ be the gradient signal of frame $t$. The consolidated densification gradient signal $\bar{g}_i$ under the standard formulation can be decomposed as
    \begin{equation}
    \bar{g}_i = \frac{1}{N} \sum_{t \in \mathcal{M}\cap\mathcal{N}} g_{i}(t) +
    \frac{1}{N} \sum_{t \in \mathcal{M}^c\cap\mathcal{N}} g_{i}(t).
    \label{eq:d_grad_signal}
    \end{equation}
    Here, the first term contributes weakly, as the limited number of visible frames is penalized by the large denominator, while the second term vanishes since $g_{i}(t)$ is zero for frames where $G_i$ is not visible. Consequently, dynamic Gaussians receive disproportionately low gradient signals that rarely exceed the densification threshold $\tau_{pos}$, thereby suppressing their densification. This imbalance becomes more pronounced as the total video duration $T$ increases.

    To address this issue, we propose a simple yet effective densification strategy that reweights the gradient signal by the temporal visibility of each Gaussian. The consolidated gradient signal for densification is reformulated as
    \begin{equation}
    \bar{g}_i =
    \frac{\sum_{t\in \mathcal{N}} \sigma_{i}(t) \; g_{i}(t)}{\sum_{t\in\mathcal{N}} \sigma_{i}(t)},
    \label{eq:d_vad}
    \end{equation}
    where $\sigma_{i}(t)$ denotes the opacity of Gaussian $G_i$ in frame $t$.
    Note that if a Gaussian remains completely visible or opaque across all frames (i.e., $\sigma_{i}(t) = 1$ for all $t$, corresponding to the static case), our densification strategy naturally reduces to the original 3DGS formulation defined in Eq.~\ref{eq:3dgs_densification}.

    \subsection{Temporally Adaptive Thresholding}
    \label{sec:tat}

    Even after applying visibility-aware densification, we observe that certain short-lived dynamic Gaussians struggle to exceed the fixed densification threshold $\tau_{\text{pos}}$, originally tuned for static scenes in 3DGS.

    To address this, we introduce a temporally adaptive thresholding strategy that modulates the densification threshold according to each Gaussian's normalized temporal lifespan $\psi\in[0, 1]$. Specifically, we redefine the threshold for a Gaussian $G_i$ as
    \begin{equation}
    \tau_{\text{pos}}^{i} =
    \tau_{\text{pos}} \cdot
    \left(
        \frac{1}{1 + \beta (1 - \bar{\psi}_i)}
    \right)^{\alpha},
    \label{eq:adaptive_tau}
    \end{equation}
    where $\tau_{\text{pos}}$ denotes the base threshold, and
    $\bar{\psi}_i = \frac{1}{N} \sum_{n\in\mathcal{N}} \psi_i^{(n)}$
    represents the average temporal scale of Gaussian $G_i$ over the set of frames considered for densification. Note that $\psi_i^{(n)}$ corresponds to the temporal scale of the Gaussian when the $n^{th}$ frame is sampled.
    This adaptive relaxation of the threshold for short-lived Gaussians (small $\psi_i$), promotes more frequent densification, while maintaining a stricter criterion for long-lived (large $\psi_i$) ones. The hyperparameters $\alpha$ and $\beta$ control the curvature and sensitivity of this temporal modulation. We choose this function with sub-linear decay to achieve more aggressive decay close to life-span 1 and slower decay close to life-span 0. Thus, other polynomial or exponential functions with appropriate parameters may also be used instead.

    Since $\psi_i$ is a learnable parameter updated at every iteration, we use its averaged form $\bar{\psi}_i$ to determine the effective temporal lifespan. In practice, we set $\beta=0.3$ to provide moderate threshold relaxation and $\alpha=1.0$ for linear scaling, empirically balancing densification quality and stability. Notably, for Gaussians with full temporal lifespan ($\psi_i = 1$), the adaptive threshold simplifies to $\tau_{\text{pos}}^{i} = \tau_{\text{pos}}$, recovering the standard static 3DGS behavior.

    Incorporating both the visibility-aware densification and temporally adaptive thresholding, our final densification criterion for a Gaussian $G_i$ is expressed as

    \begin{equation}
    \frac{\sum_{t\in \mathcal{N}} \sigma_{i}(t) \; g_{i}(t)}{\sum_{t\in\mathcal{N}} \sigma_{i}(t)}> \tau_{\text{pos}} \cdot
    \left(
        \frac{1}{1 + \beta (1 - \bar{\psi}_i)}
    \right)^{\alpha}
    \label{eq:d_criterion}
    \end{equation}
     Here, the left-hand side aggregates visibility-weighted gradients, while the right-hand side relaxes the threshold for short-lived Gaussians, together facilitating their effective densification.

    \subsection{Temporal Offset Warping}\label{sec:tow}

    Despite our visibility-aware and temporally adaptive densification strategy, we observe that certain Gaussians, particularly those corresponding to highly dynamic regions, remain under-densified. This issue arises from the limited capacity of the deformation field to model complex or rapid motion. Consequently, these Gaussians exhibit extremely low temporal scales ($\psi_i$), which substantially reduces their sampling probability among the frames considered for densification.

    The core limitation is that a standard Fourier basis, with its globally-uniform frequencies, allocates modeling capacity uniformly across time. This is inefficient for dynamic scenes where motion complexity is non-uniform, often requiring high-frequency modeling near a Gaussian's peak activity and only low-frequency modeling elsewhere.

    To address this, we introduce Temporal Offset Warping (TOW), a simple input warping scheme that enables a fixed set of Fourier basis functions to achieve \textbf{adaptive frequency allocation}. By non-linearly warping the input offset, we concentrate the model's capacity to represent high-frequency details near each Gaussian's temporal center while representing motion at lower frequencies farther away. The warped time $t'$ is defined as:
    \begin{equation}
    t' = t_i + \mathcal{W}(t - t_i; \lambda_t, \rho_t),
    \label{eq:tow_main}
    \end{equation}
    where $\lambda_t \in (0,1)$ denotes the normalized focus window size (as a fraction of the total temporal range $T$), and $\rho_t \in (0,1)$ represents the fraction of temporal capacity allocated within the focus window. The piecewise-linear warping function $\mathcal{W}$ is formulated as:
    \begin{equation}
    \mathcal{W}(\Delta t) =
    \begin{cases}
    s_{\text{near}} \cdot \Delta t, & |\Delta t| \le \tfrac{\lambda_t}{2}, \\[6pt]
    s_{\text{far}} \cdot \Delta t, & \text{otherwise},
    \end{cases}
    \label{eq:tow_warp}
    \end{equation}
    where $\Delta t = t - t_i$, and the near- and far-region scaling factors are defined as:
    \begin{equation}
    s_{\text{near}} = \frac{\rho_t}{\lambda_t}, \quad
    s_{\text{far}} = \frac{1 - \rho_t}{1 - \lambda_t}.
    \label{eq:tow_scales}
    \end{equation}

    This formulation guarantees that exactly a fraction $\rho_t$ of the total temporal deformation capacity is allocated within the focus window.

    \begin{wrapfigure}{r}{0.5\textwidth}
        \centering
        \vspace{-20pt}
        \includegraphics[width=\linewidth]{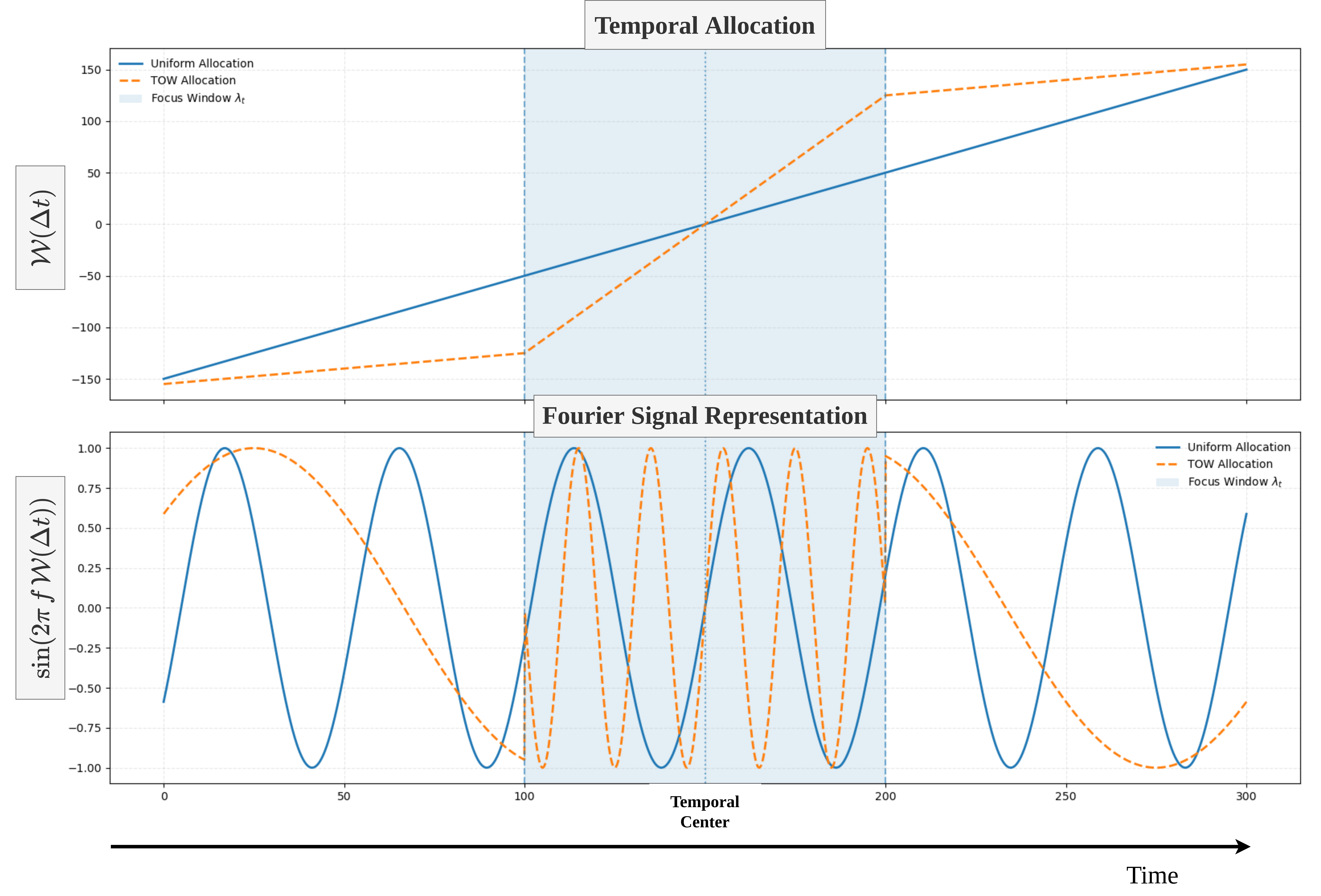}
        \vspace{-20pt}
        \caption{\textbf{Temporal Offset Warping (TOW).}
        \textbf{Top:} Warping function $\mathcal{W}(\Delta t)$ vs.\ time.
        Uniform mapping has constant unit slope ($s=1$), whereas TOW applies a piecewise-linear reparameterization with slope $s_{\text{near}}>1$ inside the focus window and $s_{\text{far}}<1$ outside, expanding near-center offsets and compressing distant ones while preserving the total temporal span.
        \textbf{Bottom:} Signals generated under uniform and warped temporal offsets. TOW enables finer temporal variation within the focus window while enforcing smoother evolution in distant regions.}
        \vspace{-30pt}
        \label{fig:tow-diagram}
        \vspace{10pt}
    \end{wrapfigure}

    \noindent\textbf{Adaptive Frequency Interpretation.}
    The key insight is that warping the input time effectively retunes the frequencies of the Fourier basis relative to physical time. The derivative of the warp function, $d\mathcal{W}/d(\Delta t)$, defines the local frequency scaling factor. Within the focus window ($|\Delta t| \le \lambda_t/2$), the scaling factor $s_{\text{near}} > 1$ stretches the input, causing the fixed Fourier frequencies to represent higher effective frequencies in physical time, enabling the modeling of complex, rapid motion. Outside the window, $s_{\text{far}} < 1$ compresses the input, causing the same basis functions to represent lower effective frequencies, suitable for smoother motion as shown in Fig.~\ref{fig:tow-diagram}. Unlike adding Fourier coefficients, which increases the parameter count and amplifies high-frequency noise, TOW achieves adaptive resolution without new parameters.

    This warped time $t'$ replaces $t$ in our deformation model (Equations~\ref{eq:opacity-deform}, \ref{eq:fourier-asym}, and \ref{eq:rot_scale_deform}). The enhanced deformation capability, particularly the ability to model high-frequency motion near $t_i$, prevents the collapse of temporal scales to extremely low values. This enables Gaussians in highly dynamic regions to maintain a longer effective temporal lifespan, thereby increasing their sampling probability during densification and leading to superior reconstruction.
    In practice, for a sequence of $T$ frames, we set $\lambda_t = \frac{50}{T}$ and $\rho_t = 0.75$, thereby concentrating $75\%$ of the temporal modeling capacity within the $50$-frame neighborhood around $t_i$.

\section{Experiments}

    We utilize three benchmark multiview datasets for our experiments:
    \textbf{Neural 3D Video} (N3DV) \cite{li2022neural} dataset comprises six indoor scenes captured by 17--21 time synchronized cameras recorded at $2704\times2028$ resolution, each featuring intricate motion confined to a small region of the frame. We use the central camera for evaluation.
    \textbf{Interdigital} \cite{Sabater2017} consists of five indoor scenes captured by a $4\times4$ camera rig at $2048\times1088$ resolution. In contrast to Neural 3D Video, these sequences exhibit long-range motion and abrupt object appearances and disappearances. We use the camera in the second row and the second column for evaluation.
    \textbf{VRU Basketball} \cite{wu2025swift4d} features two real-world basketball scenes captured with 34 cameras at $1920 \times 1080$ resolution. The scenes include fast human and ball movements, making them highly challenging for accurately modeling complex non-rigid motion and interactions. We evaluate on the camera at the center. For all experiments, we use videos downsampled to half of their original resolution.

    \textbf{Setup.}
    Two experimental setups are commonly used across dynamic 3D Gaussian Splatting methods. Approaches such as~\cite{li2023spacetime, lee2024ex4dgs} train on shorter temporal segments (e.g., 50 frames) and report results aggregated over the full 300-frame sequence. In contrast, most methods like~\cite{Wu_2024_CVPR, wu2025swift4d, yan20244d, yang2023gs4d} train a single model over all 300 frames. We adopt the latter setting, as densification challenges become more pronounced when training on longer continuous sequences due to sparser temporal sampling, whereas short-segment training increases total training time and can introduce flicker at segment boundaries. For completeness, we further benchmark our method against the short-segment training configuration in the supplementary.

    \subsection{Implementation Details \& Evaluation Metrics}
    Please refer to the supplementary for detailed discussions on benchmarking protocols and hyperparameter settings. All our method specific hyperparameters are kept fixed across all datasets and scenes. These datasets span a wide range of motion regimes, including localized rapid motion (N3DV), mixed slow and fast motion (Interdigital), and high-speed real-world sports motion (VRU).

    For evaluation, we report PSNR and SSIM~\cite{1284395} to measure overall rendering quality. However, since dynamic regions typically occupy small portions of the frame, global metrics may obscure improvements in motion areas. To specifically evaluate dynamic reconstruction quality, we introduce Masked PSNR (M-PSNR) and Masked SSIM (M-SSIM), computed exclusively within dynamic regions identified using RAFT optical flow~\cite{raft_optical_flow} between frames separated by 5 timesteps. Additionally, we report LPIPS~\cite{zhang2018perceptual} to assess perceptual image quality.

    \begin{table}[t]
        \footnotesize
        \centering
        \setlength{\tabcolsep}{3pt}
        \renewcommand{\arraystretch}{1.05}
        \caption{
        \textbf{Quantitative comparisons on the Neural 3D Video (N3DV) dataset.}
        We report PSNR, M-PSNR, M-SSIM, LPIPS, rendering speed (FPS), Train Time (TT) in minutes, and Model Size in MB, measured on NVIDIA A4000 16GB GPU.
        \textsuperscript{$(\dagger)$} indicates results obtained from official pretrained models released by the respective authors. Our model significantly improves reconstruction quality in dynamic regions, as reflected by the Masked metrics (M-PSNR and M-SSIM). Notably, our method achieves the highest FPS since it avoids heavy components like MLP decoders and multi-planar voxels used by other methods.
        }
        \label{tab:n3dv_results}
        \vspace{-2mm}
        \begin{tabular}{l c >{\columncolor{maskedcol}}c >{\columncolor{maskedcol}}c c c c c}
        \toprule
        \scriptsize\textbf{Method} &
        \scriptsize\textbf{PSNR}$\uparrow$ &
        \scriptsize\textbf{M-PSNR}$\uparrow$ &
        \scriptsize\textbf{M-SSIM}$\uparrow$ &
        \scriptsize\textbf{LPIPS}$\downarrow$ &
        \scriptsize\textbf{FPS}$\uparrow$ &
        \scriptsize\textbf{TT}$\downarrow$ &
        \scriptsize\textbf{Size$\downarrow$} \\
        \midrule
        \scriptsize 4DGaussian~\cite{Wu_2024_CVPR}
        & 31.21 & 22.67 & 0.784 & 0.071 & 61  & \colorbox{orange!30}{50} & \colorbox{red!30}{42 MB} \\

        \scriptsize STG~\cite{li2023spacetime}
        & 31.40 & 22.61 & 0.792 & 0.069 & \colorbox{orange!30}{93} & 120 & \colorbox{orange!30}{62 MB}\\

        \scriptsize Ex4DGS$^{\dagger}$~\cite{lee2024ex4dgs}
        & 31.45 & 23.40 & 0.814 & 0.078 & 29  & 144 & 213 MB\\

        \scriptsize SaroGS$^{\dagger}$~\cite{yan20244d}
        & 32.08 & 23.62 & 0.821 & 0.064 & 39  & 189 & 310 MB\\

        \scriptsize Swift4D$^{\dagger}$~\cite{wu2025swift4d}
        & \colorbox{orange!30}{32.12}
        & \colorbox{orange!30}{23.74}
        & \colorbox{orange!30}{0.835}
        & \colorbox{orange!30}{0.061}
        & 53  & \colorbox{red!30}{41} & 147 MB\\

        \midrule
        \scriptsize\textbf{Ours}
        & \colorbox{red!30}{32.42}
        & \colorbox{red!30}{24.68}
        & \colorbox{red!30}{0.863}
        & \colorbox{red!30}{0.059}
        & \colorbox{red!30}{146}
        & 62 & 204 MB\\
        \bottomrule
        \end{tabular}
        \vspace{-3mm}
    \end{table}

    \begin{wrapfigure}{r}{0.5\textwidth}
    \centering
    \vspace{-28pt}
    \makeatletter\def\@captype{table}
    \caption{\textbf{Quantitative comparison on the Interdigital dataset.}
    We report PSNR, Masked PSNR (M-PSNR), Masked SSIM (M-SSIM), and LPIPS~$\downarrow$.
    \colorbox{red!30}{Red} and \colorbox{orange!30}{Orange} highlight the best and second-best results, respectively. }
    \vspace{2mm}
    \renewcommand{\arraystretch}{1.2}
    \resizebox{\linewidth}{!}{
    \begin{tabular}{lcccc}
    \toprule
    Method &
    PSNR~$\uparrow$ &
    \cellcolor{maskedcol} M-PSNR~$\uparrow$ &
    \cellcolor{maskedcol} M-SSIM~$\uparrow$ &
    LPIPS~$\downarrow$ \\
    \midrule
    4DGaussian~\cite{Wu_2024_CVPR} & 26.74 & \cellcolor{maskedcol} 18.09 & \cellcolor{maskedcol} 0.520 & 0.173 \\
    STG~\cite{li2023spacetime} & \colorbox{orange!30}{33.45} & \cellcolor{maskedcol}\colorbox{orange!30}{27.14} & \cellcolor{maskedcol}\colorbox{orange!30}{0.860} & \colorbox{orange!30}{0.060} \\
    Ex4DGS~\cite{lee2024ex4dgs} & 32.44 & \cellcolor{maskedcol} 26.15 & \cellcolor{maskedcol} 0.844 & 0.070 \\
    Swift4D~\cite{wu2025swift4d} & 31.62 & \cellcolor{maskedcol} 23.01 & \cellcolor{maskedcol} 0.741 & 0.072 \\
    \midrule
    Ours & \colorbox{red!30}{34.14} & \cellcolor{maskedcol}\colorbox{red!30}{28.87} & \cellcolor{maskedcol}\colorbox{red!30}{0.901} & \colorbox{red!30}{0.044} \\
    \bottomrule
    \end{tabular}
    }
    \label{tab:main_results_interdigital}
    \vspace{-20pt}
    \end{wrapfigure}

    \subsection{Results}
    We compare against 4DGaussian \cite{Wu_2024_CVPR}, Ex4DGS \cite{lee2024ex4dgs}, STG \cite{li2023spacetime}, Swift4D \cite{wu2025swift4d}, and SaroGS \cite{yan20244d}. \cref{tab:n3dv_results} and \cref{tab:main_results_interdigital} report averaged results across all scenes for the N3DV and Interdigital datasets. SaroGS does not report results on Interdigital in their paper. We encountered CUDA runtime errors while attempting to run this model on Interdigital. Our method consistently outperforms all other methods across all datasets, particularly in dynamic regions. We observe that Swift4D and 4DGS perform relatively well when motion is localized to small regions, as in N3DV, whereas Ex4DGS and STG achieve better results in scenarios with slower, large-scale motion, such as the Interdigital dataset. As shown in Table~\ref{tab:n3dv_results}, our training time (TT) is comparable to 4DGaussian and Swift4D, while the resulting model size remains competitive with the top three performing methods.
    Please see the supplementary for VRU Basketball benchmark and more details on scene-wise comparisons on all datasets.

    \begin{figure*}
    \vspace{-6pt}
        \captionsetup[subfigure]{labelformat=empty}
        \centering
        \begin{tabular}{@{}c@{\hspace{2pt}}c@{\hspace{2pt}}c@{\hspace{2pt}}c@{\hspace{2pt}}c@{}}
        \subfloat[\textbf{\textcolor{red}{Ours}}]
            {\includegraphics[width=0.18\linewidth]{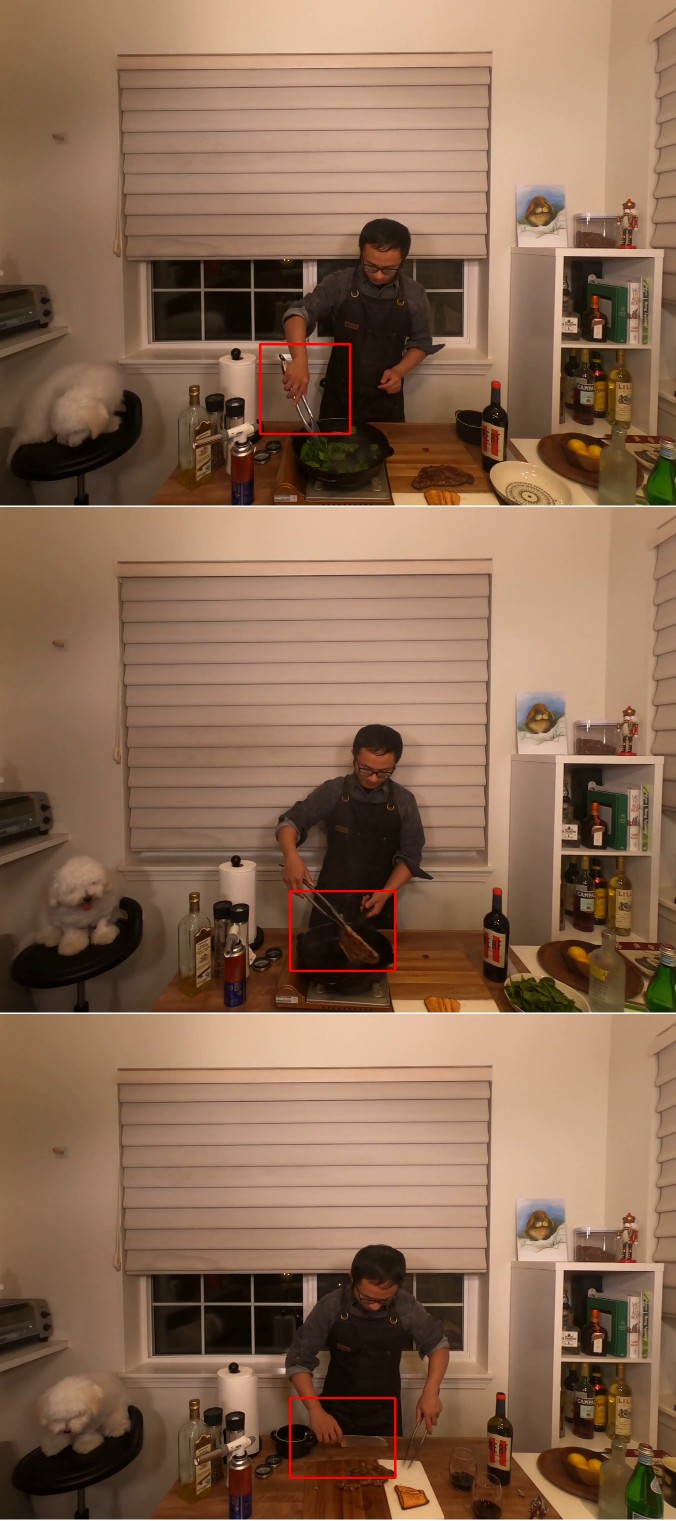}} &
        \subfloat[4DGaussian]
            {\includegraphics[width=0.18\linewidth]{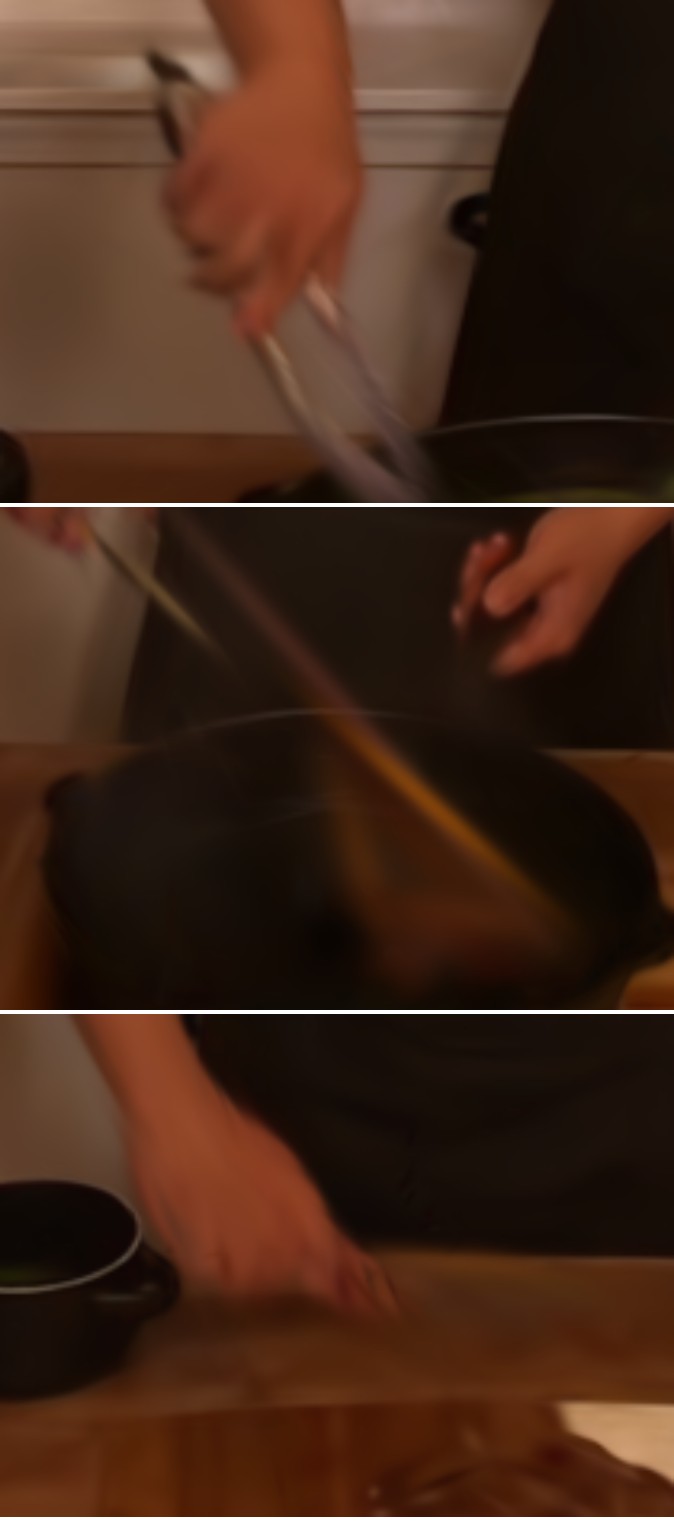}} &
        \subfloat[Swift4D]
            {\includegraphics[width=0.18\linewidth]{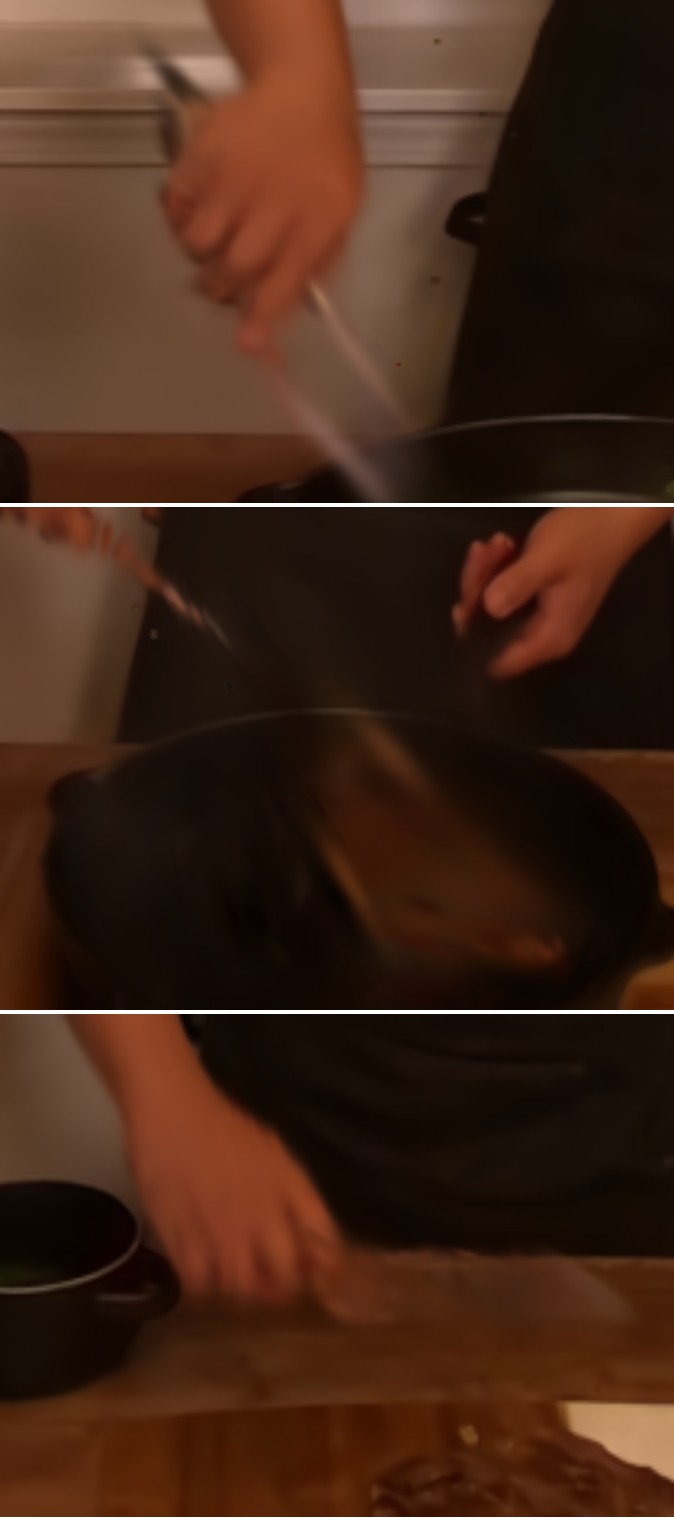}} &
        \subfloat[\textbf{\textcolor{red}{Ours}}]
            {\includegraphics[width=0.18\linewidth]{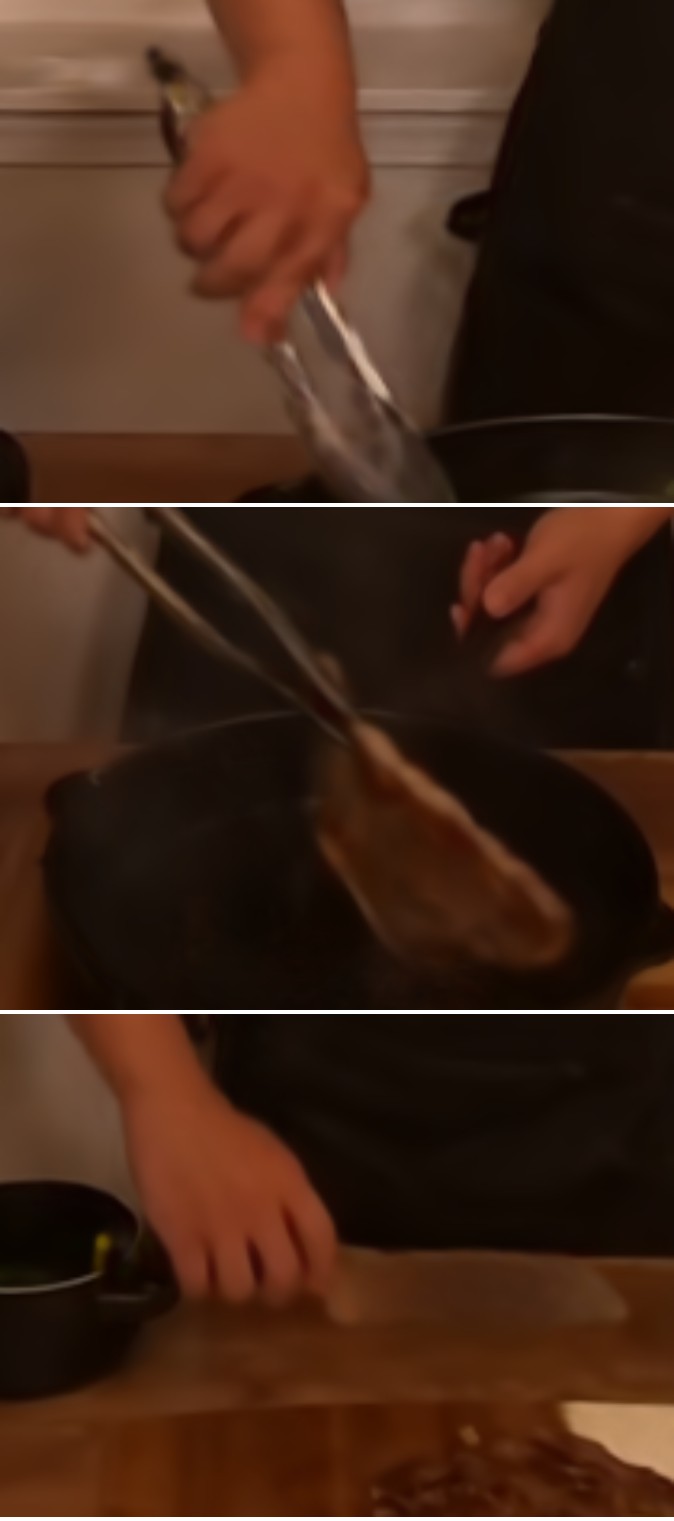}} &
        \subfloat[Ground truth]
            {\includegraphics[width=0.18\linewidth]{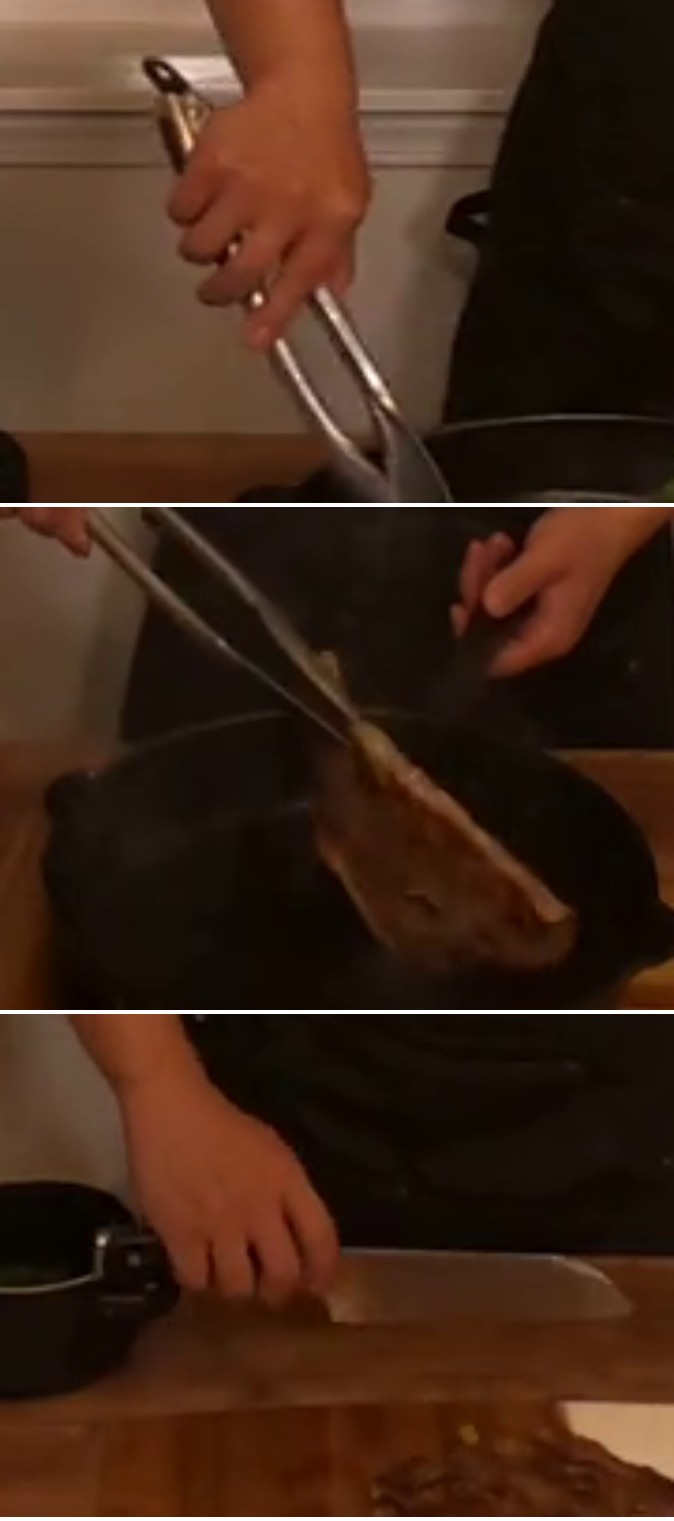}}
        \end{tabular}
        \caption{Qualitative comparison on Neural 3D Video~\cite{li2022neural}. Scene: i) \textit{cook spinach} ii) \textit{sear steak} iii) \textit{cut roasted beef}}
        \label{fig:n3d_main_figure}
    \end{figure*}

    \begin{figure*}
        \captionsetup[subfigure]{labelformat=empty}
        \centering
        \begin{tabular}{@{}c@{\hspace{4pt}}c@{\hspace{4pt}}c@{\hspace{4pt}}c@{}}
            \subfloat[STG]{\includegraphics[width=0.19\textwidth, keepaspectratio]{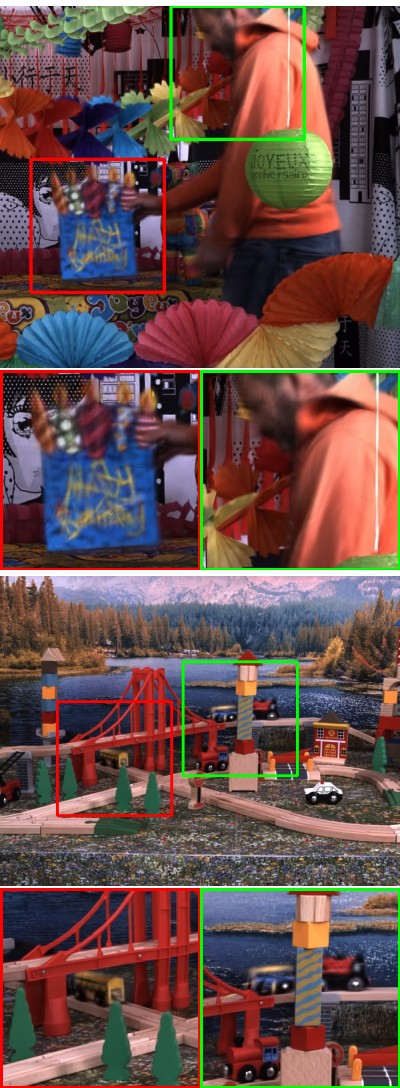}} &
            \subfloat[Ex4DGS]{\includegraphics[width=0.19\textwidth, keepaspectratio]{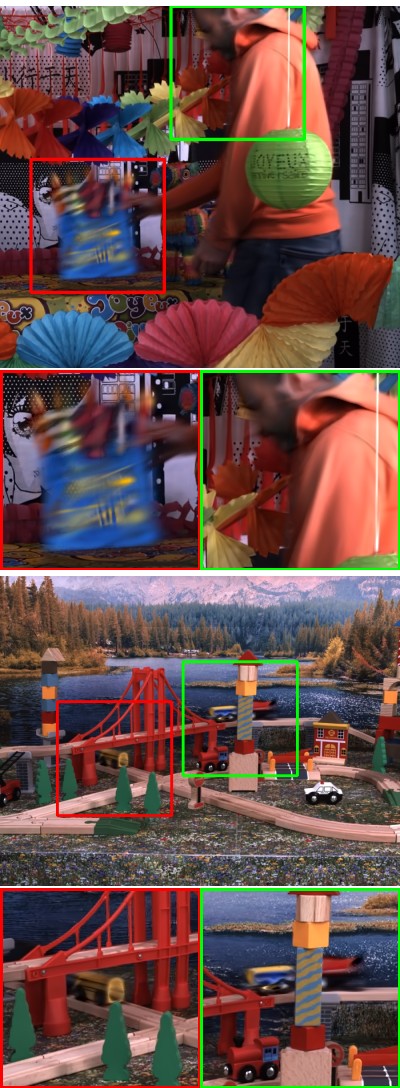}} &
            \subfloat[\textbf{\textcolor{red}{Ours}}]{\includegraphics[width=0.19\textwidth, keepaspectratio]{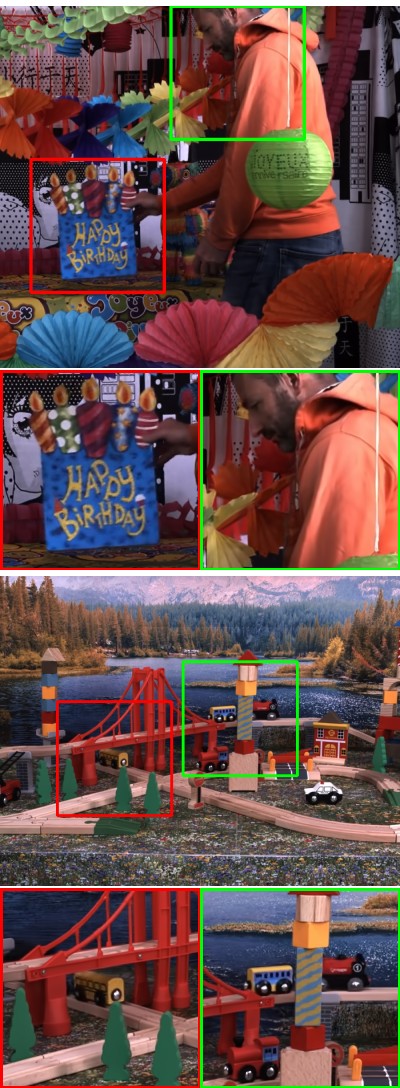}} &
            \subfloat[Ground truth]{\includegraphics[width=0.19\textwidth, keepaspectratio]{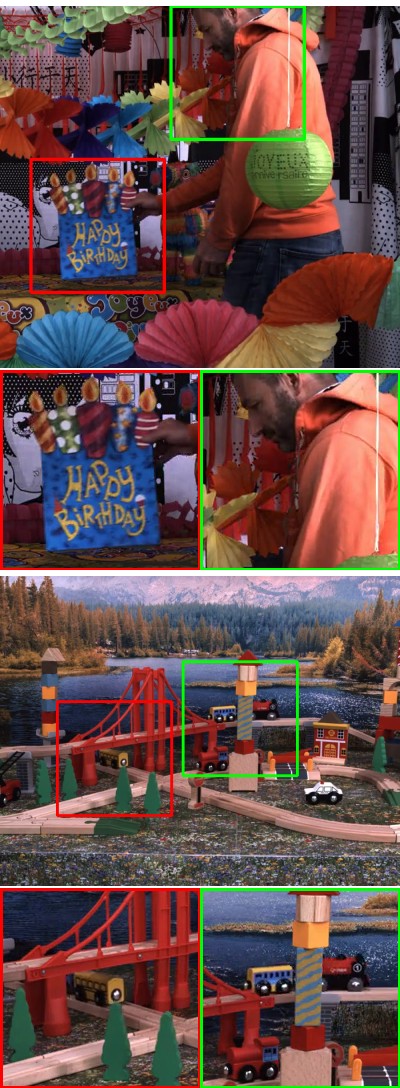}}
        \end{tabular}
        \caption{Qualitative results on the Interdigital Dataset~\cite{Sabater2017}. Scenes shown top to bottom: i) \textit{Birthday}, ii) \textit{Train}: Note that our method preserves finer details in the moving train region.}
        \label{fig:interdigital_main_figure}
    \end{figure*}

    \begin{figure*}
    \captionsetup[subfigure]{labelformat=empty}
    \centering
    \begin{tabular}{@{}c@{\hspace{2pt}}c@{\hspace{2pt}}c@{\hspace{2pt}}c@{\hspace{2pt}}c@{}}
    \subfloat[\textbf{\textcolor{red}{Ours}}]{\includegraphics[width=0.18\textwidth]{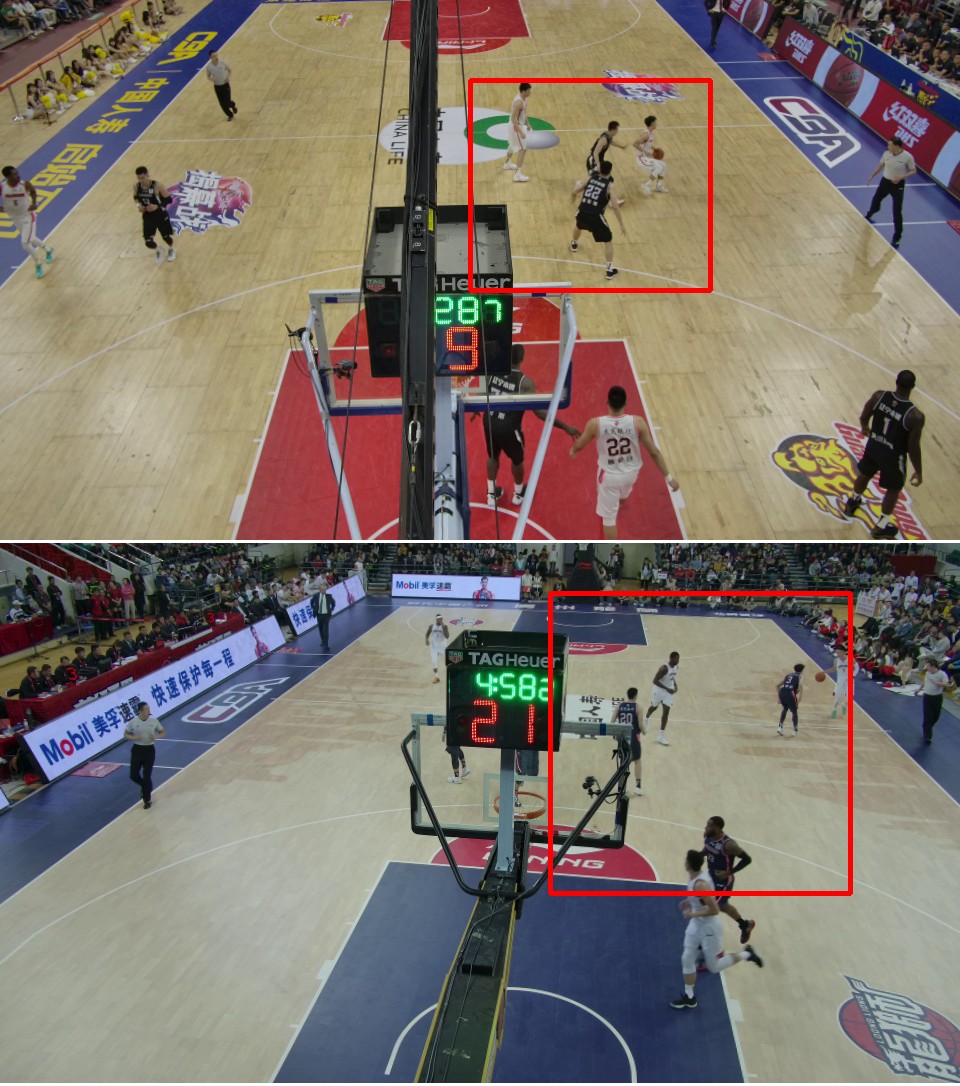}} &
    \subfloat[Swift4D]{\includegraphics[width=0.18\textwidth]{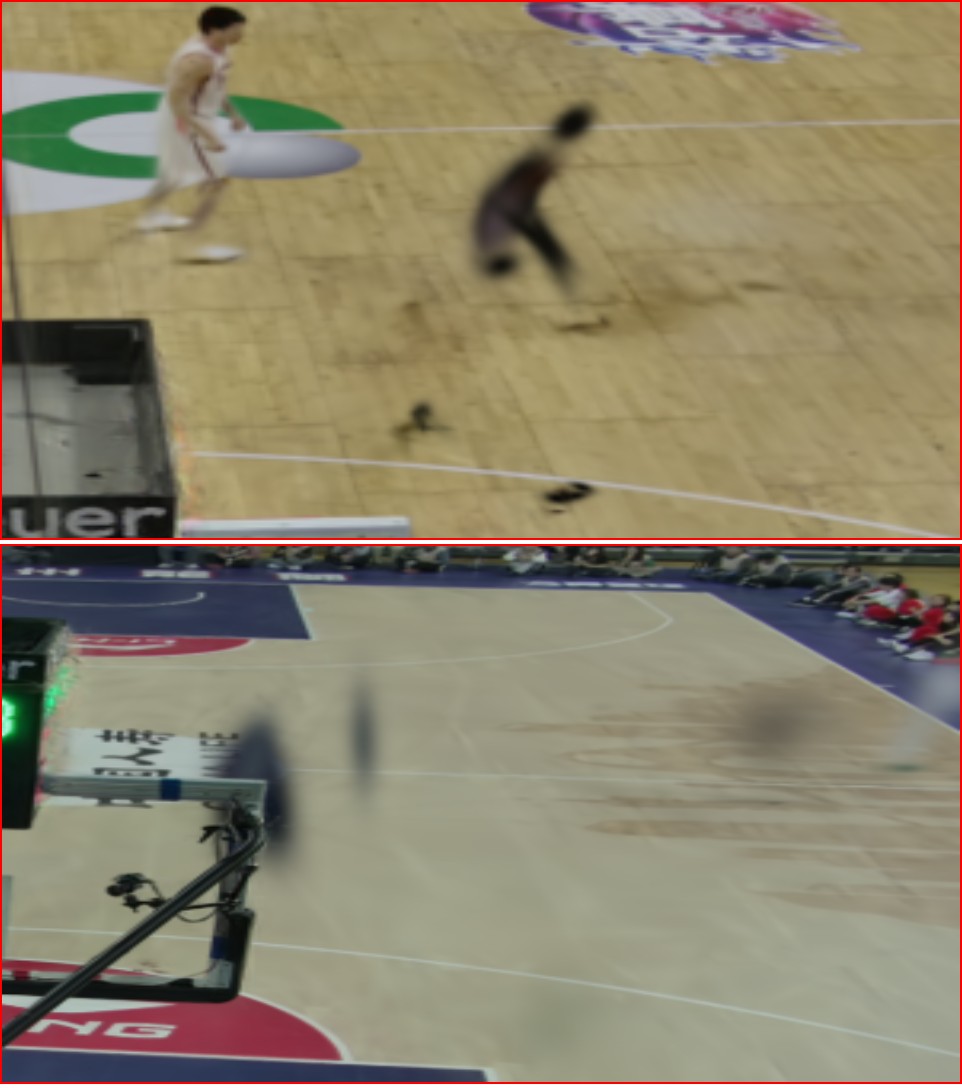}} &
    \subfloat[STG]{\includegraphics[width=0.18\textwidth]{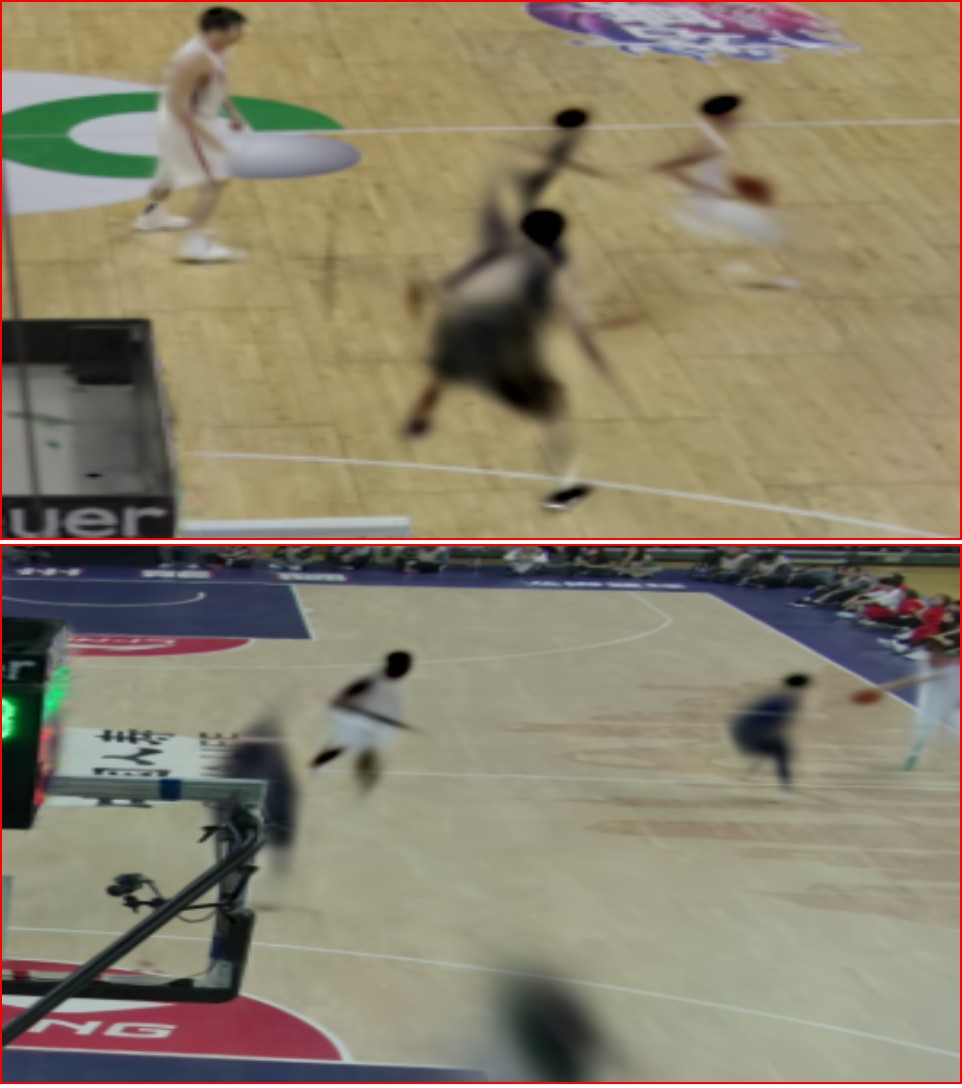}} &
    \subfloat[\textbf{\textcolor{red}{Ours}}]{\includegraphics[width=0.18\textwidth]{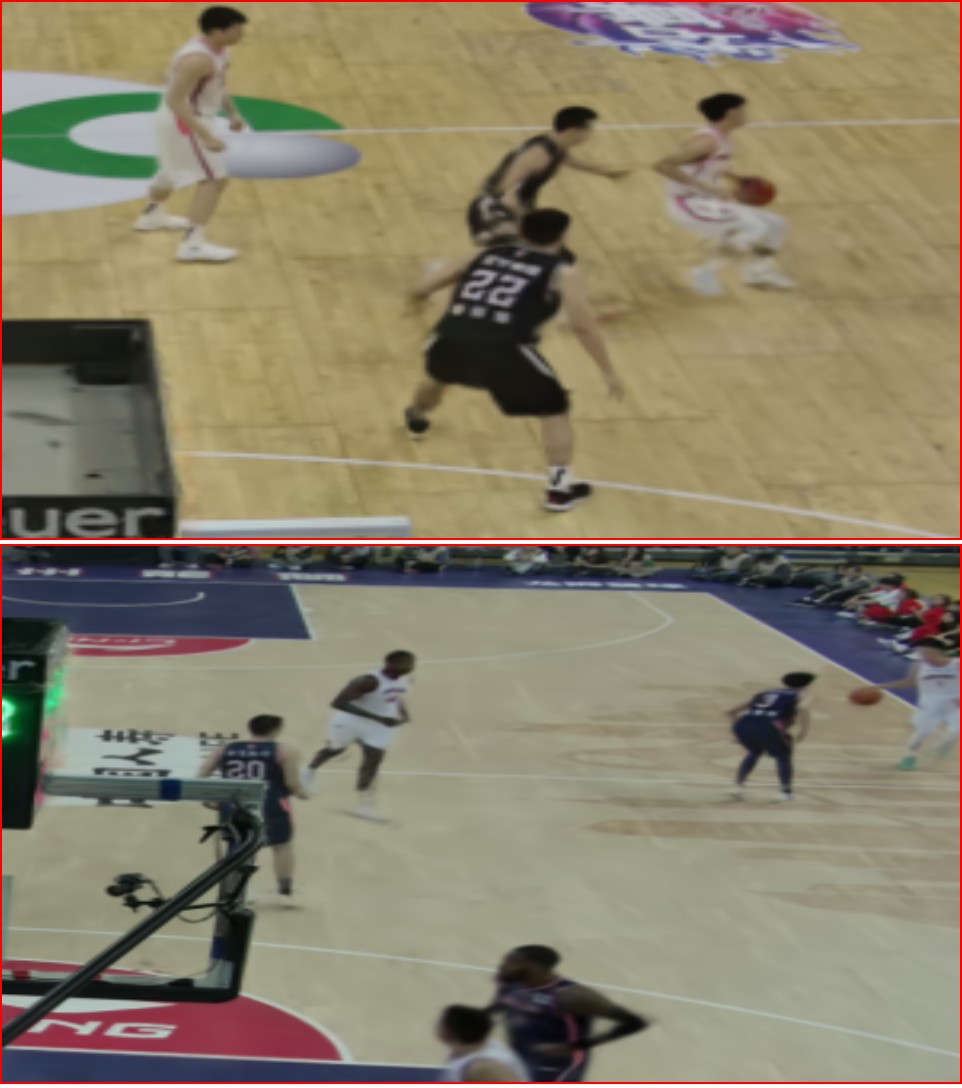}} &
    \subfloat[Ground truth]{\includegraphics[width=0.18\textwidth]{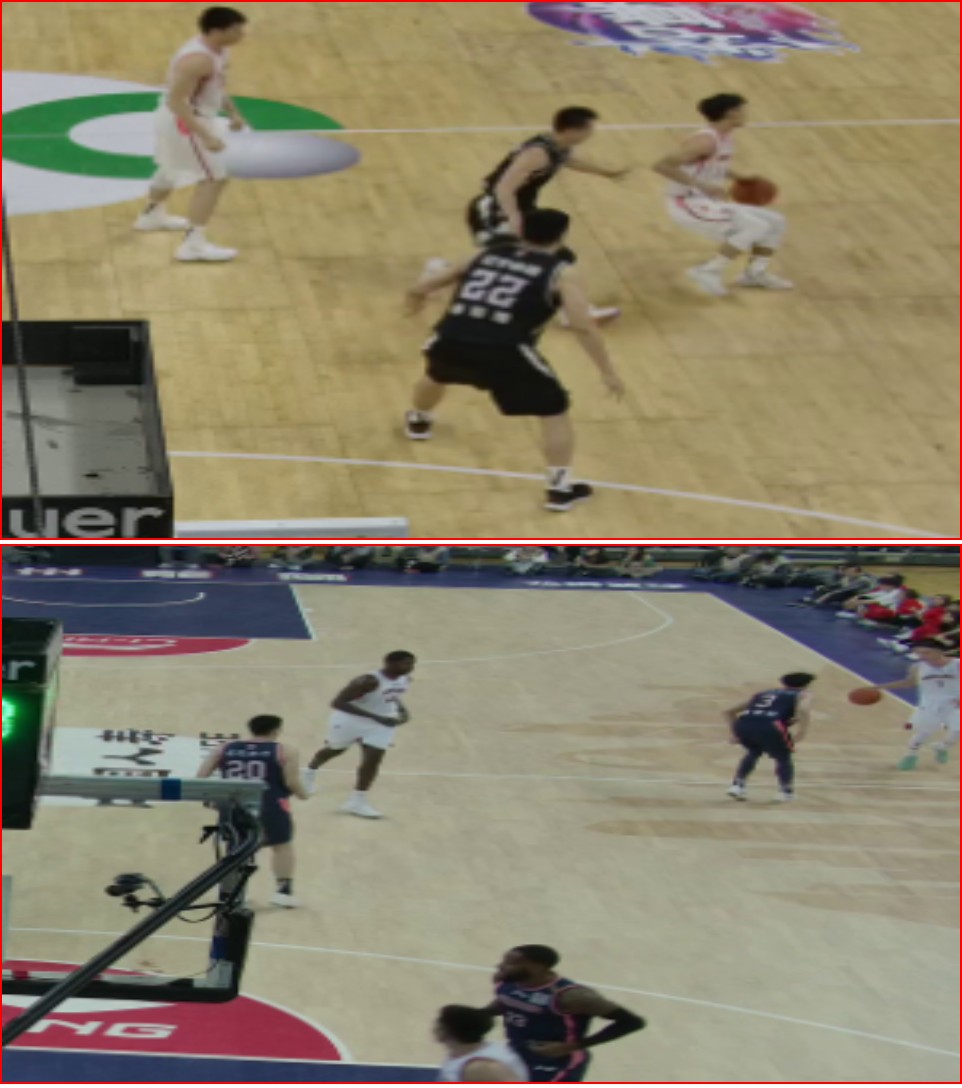}}
    \end{tabular}
    \caption{Qualitative comparison on VRU Basketball ~\cite{wu2025swift4d}. Scenes top to bottom: i) \textit{DG}, ii) \textit{GZ}}
    \label{fig:vru_figureoly}
    \end{figure*}

     We present qualitative results in Figures \ref{fig:n3d_main_figure}, \ref{fig:interdigital_main_figure}, and \ref{fig:vru_figureoly}.
     Notably, 4DGaussian and Swift4D exhibit noticeable loss of detail in regions with rapid or complex motion. Ex4DGS and STG exhibit blurring artifacts in dynamic regions when trained on longer continuous sequence. Our videos in the supplementary indicate the superior reconstruction of dynamic regions without any flicker artifacts.

    \subsection{Ablations}
    \textbf{Ablation study of the proposed components:} Table \ref{tab:ablation-combined} analyzes the contribution of our three core components: VAD, TAT, and TOW. Each component provides consistent gains across both Interdigital and N3DV datasets, with the full model achieving the highest PSNR, M-PSNR, and M-SSIM.
    \newline

\begin{figure}[t]
    \centering
    \begin{minipage}{0.48\textwidth}
        \centering
        \vspace{3mm}
        \includegraphics[width=\linewidth]{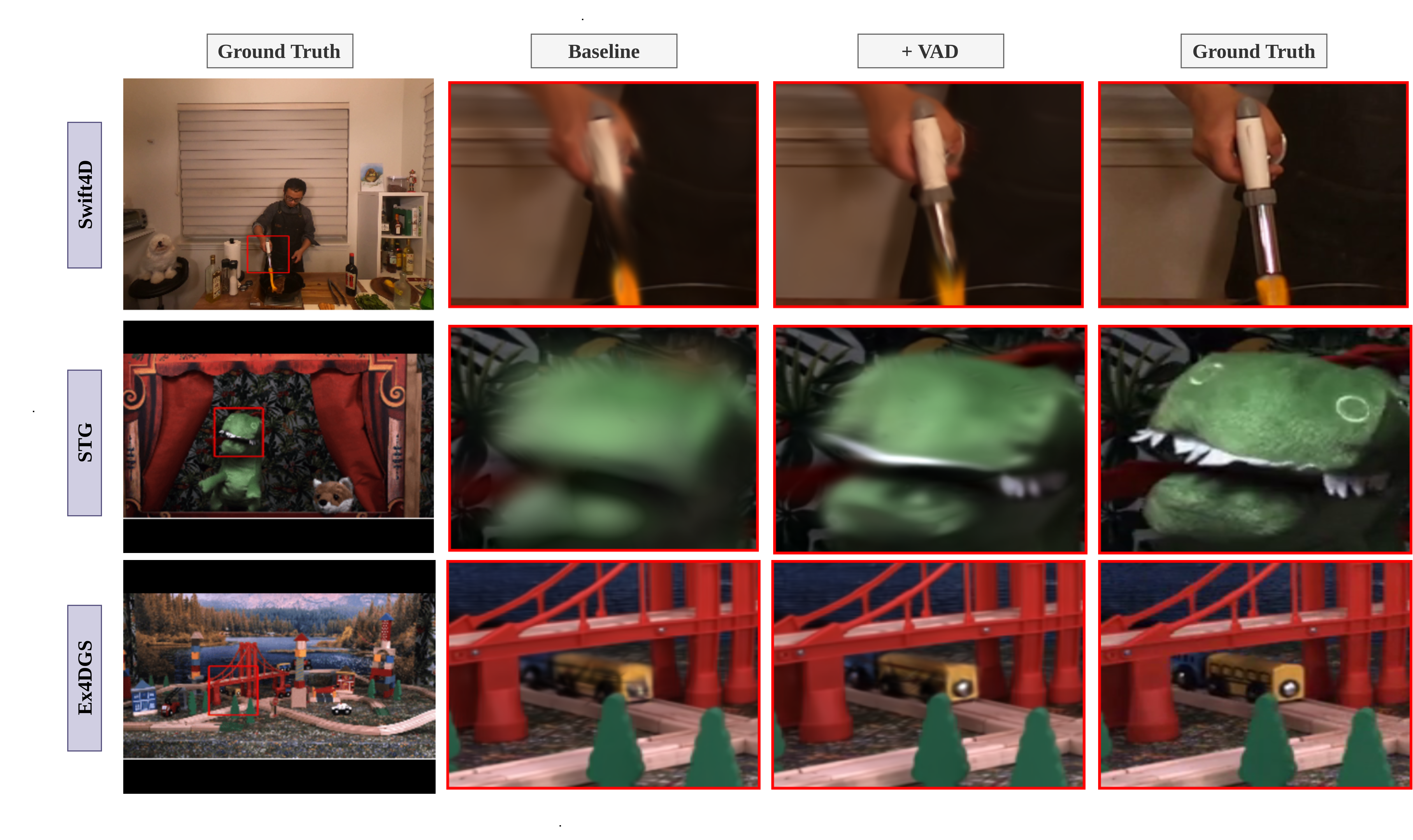}
        \vspace{-2mm}
        \caption{\textbf{Generalization of the proposed VAD module across diverse baselines.}}
        \label{fig:ablation-2-fig}
    \end{minipage}
    \hfill
    \begin{minipage}{0.48\textwidth}
        \centering
        \vspace{-2mm}
        \includegraphics[width=\linewidth]{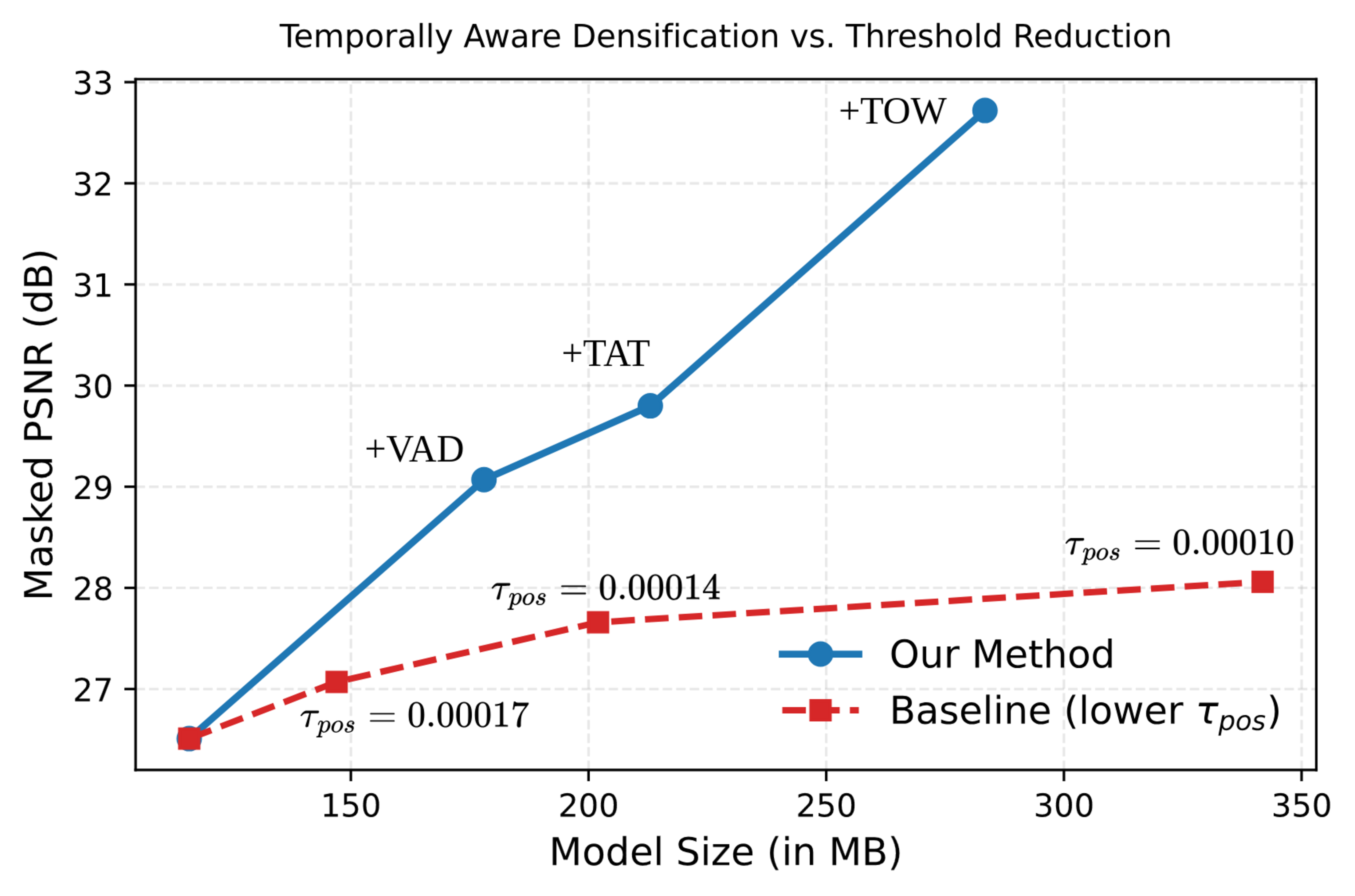}
        \vspace{-4mm}
        \caption{\textbf{Ablation on densification efficiency. (Scene: Painter)}}
        \label{fig:densification_vs_threshold}
    \end{minipage}
    \vspace{-2mm}
    \end{figure}

    \noindent\textbf{Generalization of VAD:} Table \ref{tab:ablation_vad} and Figure \ref{fig:ablation-2-fig} demonstrate that integrating the proposed VAD consistently improves reconstruction especially in dynamic regions across multiple dynamic 3DGS baselines on both Interdigital and N3DV datasets showing its plug-and-play capability. We observe larger improvements in dynamic regions for methods such as~\cite{li2023spacetime} and ours that employ multiple anchors, which are more likely to have short-lived Gaussians and thus VAD becomes important. Additionally, TAT can be naturally extended to methods that model temporal opacity using Gaussian RBFs, such as~\cite{li2023spacetime}. TOW, however, is intuitively designed for approaches that rely on a Fourier basis for motion modeling.
    \newline

    \begin{center}
    \begin{minipage}{\textwidth}

    \begin{minipage}[t]{0.48\textwidth}
    \makeatletter\def\@captype{table}
    \caption{Ablation study showing the impact of VAD, TAT, and TOW on model performance across the \textbf{Interdigital} and \textbf{N3DV} datasets. ``baseline'' refers to training with our deformation model only.}
    \small
    \setlength{\tabcolsep}{4.0pt}
    \renewcommand{\arraystretch}{1.0}
    \resizebox{\linewidth}{!}{
    \begin{tabular}{@{} c c c c c c c @{}}
    \toprule
    \textbf{VAD} & \textbf{TAT} & \textbf{TOW} & \textbf{PSNR} & \textbf{M-PSNR} & \textbf{M-SSIM} & \textbf{LPIPS} \\
    \midrule
    \multicolumn{7}{c}{\textbf{Interdigital}} \\
    \midrule
     & baseline  &  & 32.80 & 24.75 & 0.793 & 0.065 \\
    $\boldsymbol{\checkmark}$ &  &  & 33.59 & 27.32 & 0.869 & 0.049 \\
    $\boldsymbol{\checkmark}$ & $\boldsymbol{\checkmark}$ &  & 33.65 & 27.51 & 0.873 & 0.048 \\
    $\boldsymbol{\checkmark}$ & $\boldsymbol{\checkmark}$ & $\boldsymbol{\checkmark}$ & \textbf{34.14} & \textbf{28.87} & \textbf{0.901} & \textbf{0.044} \\
    \midrule
    \multicolumn{7}{c}{\textbf{Neural 3D Video}} \\
    \midrule
     & baseline  &  & 31.98 & 21.96 & 0.781 & 0.066 \\
    $\boldsymbol{\checkmark}$ &  &  & 32.14 & 23.40 & 0.832 & 0.062 \\
    $\boldsymbol{\checkmark}$ & $\boldsymbol{\checkmark}$ &  & 32.17 & 23.62 & 0.837 & 0.061 \\
    $\boldsymbol{\checkmark}$ & $\boldsymbol{\checkmark}$ & $\boldsymbol{\checkmark}$ & \textbf{32.42} & \textbf{24.68} & \textbf{0.863} & \textbf{0.059} \\
    \bottomrule
    \end{tabular}
    }
    \label{tab:ablation-combined}
    \end{minipage}
    \hfill
    \begin{minipage}[t]{0.48\textwidth}
    \makeatletter\def\@captype{table}
    \caption{\textbf{Ablation of the VAD module.}
    The dynamic metrices M-PSNR and M-SSIM show substantial gains over existing dynamic 3DGS baselines.
    }
    \vspace{2mm}
    \resizebox{\linewidth}{!}{
    \renewcommand{\arraystretch}{1.20}
    \begin{tabular}{ll
    c
    >{\columncolor{maskedcol}}c
    >{\columncolor{maskedcol}}c
    c}
    \toprule
    \textbf{Dataset} & \textbf{Method}
    & \textbf{PSNR}$\uparrow$
    & \textbf{M-PSNR}$\uparrow$
    & \textbf{M-SSIM}$\uparrow$
    & \textbf{LPIPS}$\downarrow$ \\
    \midrule

    \multirow{6}{*}{Interdigital}

    & STG~\cite{li2023spacetime}
    & 33.45 & 27.14 & 0.860 & 0.060 \\
    & \textbf{+VAD}
    & \textbf{33.67} & \textbf{27.91} & \textbf{0.884} & \textbf{0.057} \\
    \cmidrule(lr){2-6}

    & Ex4DGS~\cite{lee2024ex4dgs}
    & 32.44 & 26.15 & 0.844 & 0.070 \\
    & \textbf{+VAD}
    & \textbf{32.51} & \textbf{26.42} & \textbf{0.853} & \textbf{0.066} \\
    \cmidrule(lr){2-6}

    & Swift4D~\cite{wu2025swift4d}
    & 31.62 & 23.01 & 0.741 & 0.072 \\
    & \textbf{+VAD}
    & \textbf{31.79} & \textbf{23.31} & \textbf{0.762} & \textbf{0.070} \\

    \midrule

    \multirow{6}{*}{Neural 3D Video}

    & STG~\cite{li2023spacetime}
    & 31.40 & 22.61 & 0.792 & 0.069 \\
    & \textbf{+VAD}
    & \textbf{31.71} & \textbf{24.01} & \textbf{0.842} & \textbf{0.065} \\
    \cmidrule(lr){2-6}

    & Ex4DGS~\cite{lee2024ex4dgs}
    & 31.45 & 23.40 & 0.814 & 0.078 \\
    & \textbf{+VAD}
    & \textbf{31.62} & \textbf{23.79} & \textbf{0.830} & \textbf{0.076} \\
    \cmidrule(lr){2-6}

    & Swift4D~\cite{wu2025swift4d}
    & 32.12 & 23.74 & 0.835 & 0.061 \\
    & \textbf{+VAD}
    & \textbf{32.21} & \textbf{24.04} & \textbf{0.846} & \textbf{0.061} \\

    \bottomrule
    \end{tabular}
    }
    \label{tab:ablation_vad}
    \vspace{-2mm}
    \end{minipage}

    \end{minipage}
    \vspace{10pt}
    \end{center}

    \noindent\textbf{Temporally Aware Densification vs. Threshold Reduction:} To validate the effectiveness of our method, we compare it with a baseline that simply reduces the densification threshold $\tau_{pos}$. As illustrated in Fig.~\ref{fig:densification_vs_threshold}, lower thresholds lead to a rapid increase in model size but small performance gain. Our method, on the other hand, achieves higher Masked PSNR at a better model size.
    \newline

\begin{wrapfigure}{r}{0.48\textwidth}
        \centering
        \vspace{-25pt}
        \includegraphics[width=\linewidth]{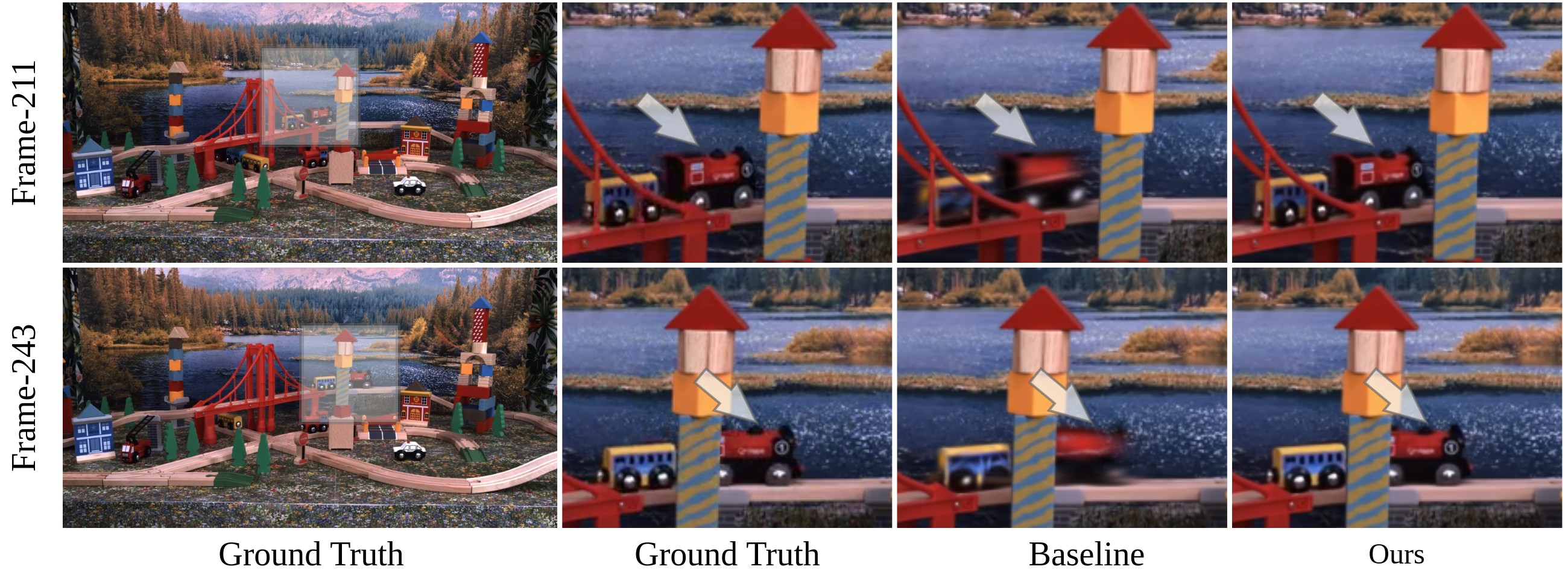}
        \vspace{-15pt}
        \caption{\textbf{Occlusion and reappearance.} \textbf{Train} scene. Baseline uses only deformation model, while ours better preserves occluded objects upon reappearance.}
        \label{fig:occlusion}
        \vspace{-20pt}
    \end{wrapfigure}

    \noindent\textbf{Handling Occlusion:} Our method is robust to occlusions due to both its temporal initialization and temporally aware densification strategy. By initializing Gaussians across multiple temporal centers, we ensure broader temporal coverage, increasing the likelihood that occluded objects are captured during initialization. Furthermore, VAD, TAT, and TOW jointly help densify Gaussians that are visible for only a few frames due to occlusion or late entry, which would otherwise receive insufficient gradients because of limited visibility. Fig~\ref{fig:occlusion} illustrates a scenario in which an object re-emerges after occlusion. We also present such cases in the supplementary videos, notably in the Painter (sudden object appearance) and Train scenes.

    \noindent Additional ablations on the VRU dataset, choice of hyperparameters, and impact of TOW on life spans of Gaussians are provided in the supplementary material.

    \section{Conclusion}
    We revisit the densification strategy in dynamic 3DGS and identify its limitations when modeling fast complex motion. To address this, we propose a deformation formulation complemented by our three key contributions. Together, these modules significantly improve reconstruction in dynamic regions, and VAD module can generalize to other models.

    \noindent\textbf{Limitations and Future Work.} Although effective, our method has not yet been tested on longer video sequences spanning several minutes; evaluating how well our method scales to such durations is an important next step. Additionally, our current focus is multiview capture, and exploring our approach to monocular or sparse-view settings remains a practical next step. We also plan to explore applying TAT and TOW as plug-and-play modules with minimal modifications to other baseline architectures.

\section*{Acknowledgements}
This work is supported in part by a grant from Anusandhan National Research Foundation, India under grant number ~ANRF/ARG/2025/002993/ENS and the Kotak IISc AI ML Center.

\bibliographystyle{splncs04}
\bibliography{main}

@String(CVPR  = {IEEE Conf. Comput. Vis. Pattern Recog.})

@String(ECCV  = {Eur. Conf. Comput. Vis.})

@String(ICLR  = {Int. Conf. Learn. Represent.})

@String(CVPRW = {IEEE Conf. Comput. Vis. Pattern Recog. Worksh.})

@String(ICASSP=	{ICASSP})

@String(TOG   = {ACM Trans. Graph.})

@String(CVPR  = {CVPR})

@String(ECCV  = {ECCV})

@String(ICLR  = {ICLR})

@String(CVPRW = {CVPRW})

@String(TOG   = {ACM TOG})

@inproceedings{mildenhall2020nerf,
 title={NeRF: Representing Scenes as Neural Radiance Fields for View Synthesis},
 author={Ben Mildenhall and Pratul P. Srinivasan and Matthew Tancik and Jonathan T. Barron and Ravi Ramamoorthi and Ren Ng},
 year={2020},
 booktitle={Proceedings of the European Conference on Computer Vision (ECCV)},
}

@article{li2023spacetime,
  title={Spacetime Gaussian Feature Splatting for Real-Time Dynamic View Synthesis},
  author={Li, Zhan and Chen, Zhang and Li, Zhong and Xu, Yi},
  journal={Proceedings of the IEEE/CVF Conference on Computer Vision and Pattern Recognition (CVPR)},
  year={2024}
}

@inproceedings{Sabater2017,
    Title = {Dataset and Pipeline for Multi-View Light-Field Video},
    Author = {Sabater, Neus and Boisson, Guillaume and Vandame, Benoit and Kerbiriou, Paul and Babon, Frederic and Hog, Matthieu and Langlois, Tristan and Gendrot, Remy and Bureller, Olivier and Schubert, Arno and Allie, Valerie},
    booktitle = {Proceedings of the IEEE/CVF Conference on Computer Vision and Pattern Recognition Workshops (CVPRW)},
    Year = {2017}
}

@inproceedings{zhang2018perceptual,
  title={The Unreasonable Effectiveness of Deep Features as a Perceptual Metric},
  author={Zhang, Richard and Isola, Phillip and Efros, Alexei A and Shechtman, Eli and Wang, Oliver},
booktitle={Proceedings of the IEEE/CVF Conference on Computer Vision and Pattern Recognition (CVPR)},
  year={2018}
}

@article{raft_optical_flow,
    title = {RAFT: Recurrent All-Pairs Field Transforms for Optical Flow},
    author = {Teed, Zachary and Deng, Jun-Yan},
    journal = {Proceedings of the European Conference on Computer Vision (ECCV)},
    year = {2020}
}

@article{1284395,
  author={Zhou Wang and Bovik, A.C. and Sheikh, H.R. and Simoncelli, E.P.},
  journal={IEEE Transactions on Image Processing}, 
  title={Image quality assessment: from error visibility to structural similarity}, 
  year={2004},
  volume={13},
  number={4},
  pages={600-612},
  doi={10.1109/TIP.2003.819861}}

@inproceedings{yang2023deformable3dgs,
    title={Deformable 3D Gaussians for High-Fidelity Monocular Dynamic Scene Reconstruction},
    author={Yang, Ziyi and Gao, Xinyu and Zhou, Wen and Jiao, Shaohui and Zhang, Yuqing and Jin, Xiaogang},
    booktitle={Proceedings of the IEEE/CVF Conference on Computer Vision and Pattern Recognition (CVPR)},
    year={2024}
}

@inproceedings{stearns2024marbles,
  title={Dynamic Gaussian Marbles for Novel View Synthesis of Casual Monocular Videos},
  author={Stearns, Colton and Harley, Adam W and Uy, Mikaela and Dubost, Florian and Tombari, Federico and Wetzstein, Gordon and Guibas, Leonidas},
  booktitle={Proceedings of the ACM Special Interest Group on Computer Graphics and Interactive Techniques - Asia (SIGGRAPH-Asia)},
  pages={1--11},
  year={2024}
}

@inproceedings{lee2024ex4dgs,
  title={Fully Explicit Dynamic Guassian Splatting},
  author={Lee, Junoh and Won, ChangYeon and Jung, Hyunjun and Bae, Inhwan and Jeon, Hae-Gon},
  booktitle={Proceedings of the Neural Information Processing Systems},
  year={2024}
}

@inproceedings{Wu_2024_CVPR,
                        author    = {Wu, Guanjun and Yi, Taoran and Fang, Jiemin and Xie, Lingxi and Zhang, Xiaopeng and Wei, Wei and Liu, Wenyu and Tian, Qi and Wang, Xinggang},
                        title     = {4D Gaussian Splatting for Real-Time Dynamic Scene Rendering},
                        booktitle = {Proceedings of the IEEE/CVF Conference on Computer Vision and Pattern Recognition (CVPR)},
                        month     = {June},
                        year      = {2024},
                        pages     = {20310-20320}
                    }

@inproceedings{yang2023gs4d,
  title={Real-time Photorealistic Dynamic Scene Representation and Rendering with 4D Gaussian Splatting},
  author={Yang, Zeyu and Yang, Hongye and Pan, Zijie and Zhang, Li},
  booktitle = {Proceedings of the International Conference on Learning Representations (ICLR)},
  year={2024}
}

@inproceedings{Chen2022ECCV,
  author = {Anpei Chen and Zexiang Xu and Andreas Geiger and Jingyi Yu and Hao Su},
  title = {TensoRF: Tensorial Radiance Fields},
  booktitle = {Proceedings of the European Conference on Computer Vision (ECCV)},
  year = {2022}
}

@inproceedings{duan:2024:4drotorgs,
   author = "Yuanxing Duan and Fangyin Wei and Qiyu Dai and Yuhang He and Wenzheng Chen and Baoquan Chen",
   title = "{4D-Rotor Gaussian Splatting: Towards Efficient Novel View Synthesis for Dynamic Scenes}",
   booktitle = "Proceedings of the ACM Special Interest Group on Computer Graphics and Interactive Techniques (SIGGRAPH)",
   year = "2024",
   month = {July}
}

@article{kerbl3Dgaussians,
      author       = {Kerbl, Bernhard and Kopanas, Georgios and Leimk{\"u}hler, Thomas and Drettakis, George},
      title        = {{3D Gaussian Splatting for Real-Time Radiance Field Rendering}},
      journal      = {ACM Transactions on Graphics (TOG)},
      number       = {4},
      volume       = {42},
      month        = {July},
      year         = {2023},
}

@inproceedings{li2022neural,
  title={Neural {3D} video synthesis from multi-view video},
  author={Li, Tianye and Slavcheva, Mira and Zollhoefer, Michael and Green, Simon and Lassner, Christoph and Kim, Changil and Schmidt, Tanner and Lovegrove, Steven and Goesele, Michael and Newcombe, Richard and others},
  booktitle={Proceedings of the IEEE/CVF Conference on Computer Vision and Pattern Recognition (CVPR)},
  pages={5521--5531},
  year={2022}
}

@inproceedings{li2020neural,
  title={Neural Scene Flow Fields for Space-Time View Synthesis of Dynamic Scenes},
  author={Li, Zhengqi and Niklaus, Simon and Snavely, Noah and Wang, Oliver},
  booktitle = {Proceedings of the IEEE/CVF Conference on Computer Vision and Pattern Recognition (CVPR)},
  year={2021}
}

@article{xu2024longvolcap,
  author  = {Xu, Zhen and Xu, Yinghao and Yu, Zhiyuan and Peng, Sida and Sun, Jiaming and Bao, Hujun and Zhou, Xiaowei},
  title   = {Representing Long Volumetric Video with Temporal Gaussian Hierarchy},
  journal = {ACM Transactions on Graphics (TOG)},
  number  = {6},
  volume  = {43},
  month   = {November},
  year    = {2024},
}

@inproceedings{shaw2024swings,
  title={Swings: sliding windows for dynamic 3D Gaussian Splatting},
  author={Shaw, Richard and Nazarczuk, Michal and Song, Jifei and Moreau, Arthur and Catley-Chandar, Sibi and Dhamo, Helisa and P{\'e}rez-Pellitero, Eduardo},
  booktitle = {Proceedings of the European Conference on Computer Vision (ECCV)},
  pages={37--54},
  year={2024},
  organization={Springer}
}

@inproceedings{lin2024gaussianflow,
    author={Youtian Lin and Zuozhuo Dai and Siyu Zhu and Yao Yao},
    title={Gaussian-Flow: 4D Reconstruction with Dynamic 3D Gaussian Particle},
      booktitle={Proceedings of the IEEE/CVF Conference on Computer Vision and Pattern Recognition (CVPR)},
    pages={21136--21145},
    year={2024}
}

@inproceedings{yan20244d,
      title={4D Gaussian Splatting with Scale-aware Residual Field and Adaptive Optimization for Real-time rendering of temporally complex dynamic scenes},
      author={Yan, Jinbo and Peng, Rui and Tang, Luyang and Wang, Ronggang},
      booktitle={ACM Multimedia 2024},
      year={2024}
    }

@inproceedings{wu2025swift4d,
  title={Swift4D: Adaptive divide-and-conquer Gaussian Splatting for compact and efficient reconstruction of dynamic scene},
  author={Wu, Jiahao and Peng, Rui and Wang, Zhiyan and Xiao, Lu and Tang, Luyang and Yan, Jinbo and Xiong, Kaiqiang and Wang, Ronggang},
    booktitle = {Proceedings of the International Conference on Learning Representations (ICLR)},
  year={2025}
}

@inproceedings{Park_2025_CVPR,
      author    = {Park, Jongmin and Bui, Minh-Quan Viet and Bello, Juan Luis Gonzalez and Moon, Jaeho and Oh, Jihyong and Kim, Munchurl},
      title     = {SplineGS: Robust Motion-Adaptive Spline for Real-Time Dynamic 3D Gaussians from Monocular Video},
      booktitle={Proceedings of the IEEE/CVF Conference on Computer Vision and Pattern Recognition (CVPR)},
      month     = {June},
      year      = {2025},
      pages     = {26866-26875}
  }

@inproceedings{Huang2DGS2024,
    title={2D Gaussian Splatting for Geometrically Accurate Radiance Fields},
    author={Huang, Binbin and Yu, Zehao and Chen, Anpei and Geiger, Andreas and Gao, Shenghua},
    publisher = {Association for Computing Machinery},
    booktitle = "Proceedings of the ACM Special Interest Group on Computer Graphics and Interactive Techniques (SIGGRAPH)",
    year      = {2024},
    doi       = {10.1145/3641519.3657428}
}

@inproceedings{zhang2024cor,
      title={CoR-GS: Sparse-View 3D Gaussian Splatting via Co-Regularization},
      author={Zhang, Jiawei and Li, Jiahe and Yu, Xiaohan and Huang, Lei and Gu, Lin and Zheng, Jin and Bai, Xiao},
      booktitle = {Proceedings of the European Conference on Computer Vision (ECCV)},
      year={2024}
    }

@inproceedings{park2025dropgaussian,
  title={Dropgaussian: Structural regularization for sparse-view Gaussian Splatting},
  author={Park, Hyunwoo and Ryu, Gun and Kim, Wonjun},
  booktitle={Proceedings of the IEEE/CVF Conference on Computer Vision and Pattern Recognition (CVPR)},
  year={2025}
}

@inproceedings{zhang2024pixelgs,
  title     = {Pixel-GS: Density Control with Pixel-aware Gradient for 3D Gaussian Splatting},
  author    = {Zhang, Zheng and Hu, Wenbo and Lao, Yixing and He, Tong and Zhao, Hengshuang},
  booktitle = {Proceedings of the European Conference on Computer Vision (ECCV)},
  year      = {2024}
}

@inproceedings{zhu2024fsgs,
  title={FSGS: Real-time few-shot view synthesis using Gaussian Splatting},
  author={Zhu, Zehao and Fan, Zhiwen and Jiang, Yifan and Wang, Zhangyang},
  booktitle = {Proceedings of the European Conference on Computer Vision (ECCV)},
  year={2024},
}

@inproceedings{rota2024revising,
  title={Revising densification in Gaussian Splatting},
  author={Rota Bul{\`o}, Samuel and Porzi, Lorenzo and Kontschieder, Peter},
  booktitle = {Proceedings of the European Conference on Computer Vision (ECCV)},
  year={2024},
}

@article{bond2025gaussianvideo,
  title={Gaussianvideo: Efficient video representation via hierarchical Gaussian Splatting},
  author={Bond, Andrew and Wang, Jui-Hsien and Mai, Long and Erdem, Erkut and Erdem, Aykut},
  journal={arXiv preprint arXiv:2501.04782},
  year={2025}
}

@article{kingma2014adam,
  title={Adam: A method for stochastic optimization},
  author={Kingma, Diederik P},
  journal={arXiv preprint arXiv:1412.6980},
  year={2014}
}

@misc{yuan20251000fps4dgaussian,
      title={1000+ FPS 4D Gaussian Splatting for Dynamic Scene Rendering}, 
      author={Yuheng Yuan and Qiuhong Shen and Xingyi Yang and Xinchao Wang},
      year={2025},
      eprint={2503.16422},
      archivePrefix={arXiv},
      primaryClass={cs.CV},
      url={https://arxiv.org/abs/2503.16422}, 
}

@inproceedings{yu2022plenoxels,
      title={Plenoxels: Radiance Fields without Neural Networks}, 
      author={Sara Fridovich-Keil and Alex Yu and Matthew Tancik and Qinhong Chen and Benjamin Recht and Angjoo Kanazawa},
      year={2022},
      booktitle={Proceedings of the IEEE/CVF Conference on Computer Vision and Pattern Recognition (CVPR)},
}

@inproceedings{barron2021mip,
  title={Mip-nerf: A multiscale representation for anti-aliasing neural radiance fields},
  author={Barron, Jonathan T and Mildenhall, Ben and Tancik, Matthew and Hedman, Peter and Martin-Brualla, Ricardo and Srinivasan, Pratul P},
  booktitle={Proceedings of the IEEE/CVF Conference on Computer Vision and Pattern Recognition (CVPR)},
  pages={5855--5864},
  year={2021}
}

@inproceedings{SunSC22,
  author    = {Cheng Sun and Min Sun and Hwann{-}Tzong Chen},
  title     = {Direct Voxel Grid Optimization: Super-fast Convergence for Radiance Fields Reconstruction},
  booktitle={Proceedings of the IEEE/CVF Conference on Computer Vision and Pattern Recognition (CVPR)},
  year      = {2022},
}

@inproceedings{somraj2023VipNeRF,
    title = {{ViP-NeRF}: Visibility Prior for Sparse Input Neural Radiance Fields},
    author = {Somraj, Nagabhushan and Soundararajan, Rajiv},
    booktitle = "Proceedings of the ACM Special Interest Group on Computer Graphics and Interactive Techniques (SIGGRAPH)",
    month = {August},
    year = {2023},
    doi = {10.1145/3588432.3591539},
}

@inproceedings{somraj2023simplenerf,
    title = {{SimpleNeRF}: Regularizing Sparse Input Neural Radiance Fields with Simpler Solutions},
    author = {Somraj, Nagabhushan and Karanayil, Adithyan and Soundararajan, Rajiv},
    booktitle={Proceedings of the ACM Special Interest Group on Computer Graphics and Interactive Techniques - Asia (SIGGRAPH-Asia)},
    month = {December},
    year = {2023},
    doi = {10.1145/3610548.3618188}
}

@article{mueller2022instant,
    author = {Thomas M\"uller and Alex Evans and Christoph Schied and Alexander Keller},
    title = {Instant Neural Graphics Primitives with a Multiresolution Hash Encoding},
    journal = {ACM Transactions on Graphics (TOG)},
    issue_date = {July 2022},
    volume = {41},
    number = {4},
    month = jul,
    year = {2022},
    pages = {102:1--102:15},
    articleno = {102},
    numpages = {15},
    url = {https://doi.org/10.1145/3528223.3530127},
    doi = {10.1145/3528223.3530127},
    publisher = {ACM},
    address = {New York, NY, USA},
}

@inproceedings{Kim_2024_CVPR,
    author    = {Kim, Sieun and Lee, Kyungjin and Lee, Youngki},
    title     = {Color-cued Efficient Densification Method for 3D Gaussian Splatting},
    booktitle = {Proceedings of the IEEE/CVF Conference on Computer Vision and Pattern Recognition Workshops (CVPRW)},
    month     = {June},
    year      = {2024},
    pages     = {775-783}
}

@inproceedings{10889028,
  author={Huang, Haoyang and Zhang, Zhe and Wu, Guanhua and Wang, Ronggang},
  booktitle={ICASSP 2025 - 2025 IEEE International Conference on Acoustics, Speech and Signal Processing (ICASSP)}, 
  title={PGDGS: Improving Few-shot 3D Gaussian Splatting with Progressive Gaussian Densification}, 
  year={2025},
  volume={},
  number={},
  pages={1-5},
  doi={10.1109/ICASSP49660.2025.10889028}}

@inproceedings{Grubert_2025,
   title={Improving Adaptive Density Control for 3D Gaussian Splatting},
   url={http://dx.doi.org/10.5220/0013308500003912},
   DOI={10.5220/0013308500003912},
   booktitle={Proceedings of the 20th International Joint Conference on Computer Vision, Imaging and Computer Graphics Theory and Applications},
   publisher={SCITEPRESS - Science and Technology Publications},
   author={Grubert, Glenn and Barthel, Florian and Hilsmann, Anna and Eisert, Peter},
   year={2025},
   pages={610--621}
}

@inproceedings{10887967,
  author={Wang, Zhanke and Wu, Guanhua and Wang, Zhiyan and Xiao, Lu and Liu, Runling and Wu, Jiahao and Wang, Ronggang},
  booktitle={ICASSP 2025 - 2025 IEEE International Conference on Acoustics, Speech and Signal Processing (ICASSP)}, 
  title={HDA-GS: Hierarchical Density-Controlled for Anisotropic 3D Gaussian Splatting}, 
  year={2025},
  volume={},
  number={},
  pages={1-5},
  doi={10.1109/ICASSP49660.2025.10887967}}

@inproceedings{Patle2025ADGS,
    author = {Patle, Gurutva and Girgaonkar, Nilay and Somraj, Nagabhushan and Soundararajan, Rajiv},
    title = {{AD-GS: Alternating Densification for Sparse-Input 3D Gaussian Splatting}},
    booktitle={Proceedings of the ACM Special Interest Group on Computer Graphics and Interactive Techniques - Asia (SIGGRAPH-Asia)},
    year = {2025},
    doi = {10.1145/3757377.3763993},
}

@inproceedings{10.1145/3757377.3763898,
author = {Li, Yilong and Pang, Bo and Chen, Yisong and Wang, Guoping},
title = {Anchored 4D Gaussian Splatting for Dynamic Novel View Synthesis},
year = {2025},
isbn = {9798400721373},
publisher = {Association for Computing Machinery},
address = {New York, NY, USA},
url = {https://doi.org/10.1145/3757377.3763898},
doi = {10.1145/3757377.3763898},
booktitle = {Proceedings of the ACM Special Interest Group on Computer Graphics and Interactive Techniques - Asia (SIGGRAPH-Asia)}
}

\clearpage
\section*{Supplementary Material}

\noindent The contents of this supplement include
    \begin{enumerate}
        \item Implementation details
        \item VRU Results
        \item More Ablations
        \item Qualitative Comparisons
        \item Video comparisons
        \item Scene-wise quantitative comparisons.
    \end{enumerate}

\section{Implementation Details}
We implement our framework in PyTorch, building upon the 3DGS codebase and its differentiable rasterization pipeline. All experiments are conducted on a single NVIDIA A4000 16GB GPU. We employ the Adam optimizer~\cite{kingma2014adam} with an initial learning rate of $2.6\times10^{-4}$ for Gaussian mean parameters, applying exponential decay to a final rate of $2.6\times10^{-6}$. Training proceeds for 40K iterations, with densification applied every 200 iterations up to iteration 17k. We use a batch size of 1 for all experiments, and retain the default 3DGS parameters for all other optimization settings.

\noindent\textbf{Benchmarking Protocols.} We evaluate the competing methods on the Neural 3D Video~\cite{li2022neural}, Interdigital~\cite{Sabater2017}, and VRU~\cite{wu2025swift4d} datasets.

\noindent\textit{Neural 3D Video.} For Ex4DGS~\cite{lee2024ex4dgs}, SaroGS~\cite{yan20244d}, and Swift4D~\cite{wu2025swift4d}, we use the official pretrained models released by the respective authors. For 4DGaussian~\cite{Wu_2024_CVPR}, we train and evaluate the model using the default configuration recommended in the paper. Since STG~\cite{li2023spacetime} is designed for 50-frame sequences, we adapt it to the full 300-frame sequence by proportionally scaling the densification count (6×) and training for 40k iterations; matching our setup for a fair comparison. As also noted in Ex4DGS~\cite{lee2024ex4dgs}, STG exhibits performance degradation when trained on longer sequences.

\noindent\textit{Interdigital.} For STG and Ex4DGS, we train and evaluate on the full 300-frame sequence using the authors' default configurations, scaling the densification count proportionally and training for 40k iterations. Since 4DGaussian and Swift4D do not report results on this dataset, we train these methods for 40k iterations using their default multiview configuration.

\noindent\textit{VRU.} All benchmarked methods are trained and evaluated on the full 250-frame sequence using the default multiview configuration provided by the respective authors, for 40k iterations.

\noindent\textbf{Hyperparameters.} All method-specific hyperparameters are kept fixed across all datasets and scenes. Table~\ref{tab:hyperparams} summarizes the hyperparameters used in our method.

\begin{table}[h]
\centering
\caption{\textbf{Hyperparameter Table.}
In TOW, $T$ denotes the total number of frames in the video (i.e., the video duration).}
\vspace{1mm}

\setlength{\tabcolsep}{10pt}
\renewcommand{\arraystretch}{1.2}

\begin{tabular}{lccc}
\toprule
\textbf{Dataset} & \textbf{VAD} & \textbf{TAT} & \textbf{TOW} \\
\midrule
N3DV, Interdigital, VRU
& None
& $\alpha = 1.0,\; \beta = 0.3$
& $\lambda_t = \frac{50}{\text{T}},\; \rho_t = 0.75$ \\
\bottomrule
\end{tabular}

\label{tab:hyperparams}
\vspace{-2mm}
\end{table}

\section{VRU Results}
\label{sec:vru_results}
We benchmark our method on the VRU dataset against STG and Swift4D, as shown in Table~\ref{tab:main_results_vru}. Competing methods struggle to preserve fine details in highly dynamic regions, while our approach retains these details to a good extent as shown in Figure 6 in the main paper.

\begin{table}[t]
\centering

\begin{minipage}[t]{0.49\linewidth}
\makeatletter\def\@captype{table}
\caption{\textbf{Quantitative comparison on the VRU dataset.}
We report PSNR, Masked PSNR (M-PSNR), Masked SSIM (M-SSIM), and LPIPS~$\downarrow$.}
\vspace{2mm}

\small
\renewcommand{\arraystretch}{1.2}

\resizebox{\linewidth}{!}{
\begin{tabular}{lcccc}
\toprule
Method & PSNR~$\uparrow$ & M-PSNR~$\uparrow$ & M-SSIM~$\uparrow$ & LPIPS~$\downarrow$ \\
\midrule

STG~\cite{li2023spacetime}
& \colorbox{orange!30}{26.03}
& \colorbox{orange!30}{21.33}
& \colorbox{orange!30}{0.736}
& \colorbox{orange!30}{0.127} \\

Swift4D~\cite{wu2025swift4d}
& 25.72
& 19.52
& 0.705
& 0.129 \\

\midrule
\textbf{Ours}
& \colorbox{red!30}{28.28}
& \colorbox{red!30}{25.39}
& \colorbox{red!30}{0.881}
& \colorbox{red!30}{0.090} \\

\bottomrule
\end{tabular}
}

\label{tab:main_results_vru}
\end{minipage}
\hfill
\begin{minipage}[t]{0.49\linewidth}
\makeatletter\def\@captype{table}
\caption{Ablation study showing the impact of VAD, TAT, and TOW on model performance on the \textbf{VRU} dataset. ``baseline'' refers to training with our deformation model only.}

\small
\setlength{\tabcolsep}{4pt}
\renewcommand{\arraystretch}{1.2}

\resizebox{\linewidth}{!}{
\begin{tabular}{@{} c c c c c c c @{}}
\toprule
\textbf{VAD} & \textbf{TAT} & \textbf{TOW} & \textbf{PSNR} & \textbf{M-PSNR} & \textbf{M-SSIM} & \textbf{LPIPS} \\
\midrule

\multicolumn{7}{c}{\textbf{VRU Basketball}} \\
\midrule

 & baseline &  & 26.75 & 21.03 & 0.737 & 0.134 \\

$\boldsymbol{\checkmark}$ &  &  & 27.63 & 23.24 & 0.828 & 0.105 \\

$\boldsymbol{\checkmark}$ & $\boldsymbol{\checkmark}$ &  & 27.71 & 23.45 & 0.835 & 0.104 \\

$\boldsymbol{\checkmark}$ & $\boldsymbol{\checkmark}$ & $\boldsymbol{\checkmark}$
& \textbf{28.28}
& \textbf{25.39}
& \textbf{0.881}
& \textbf{0.090} \\

\bottomrule
\end{tabular}
}

\label{tab:ablation-vru}
\end{minipage}

\end{table}

\section{More Ablations}
\label{sec:more_ablations}

\noindent\textbf{Ablation of the proposed components on VRU Dataset:} Table~\ref{tab:ablation-vru} presents the ablation results on VRU, demonstrating the contribution of each core component, exhibiting trends consistent with those observed in Table 4 in the main paper.

\begin{figure}[t]
\centering

\begin{minipage}{0.48\textwidth}
    \centering
    \vspace{3mm}
    \includegraphics[width=\linewidth]{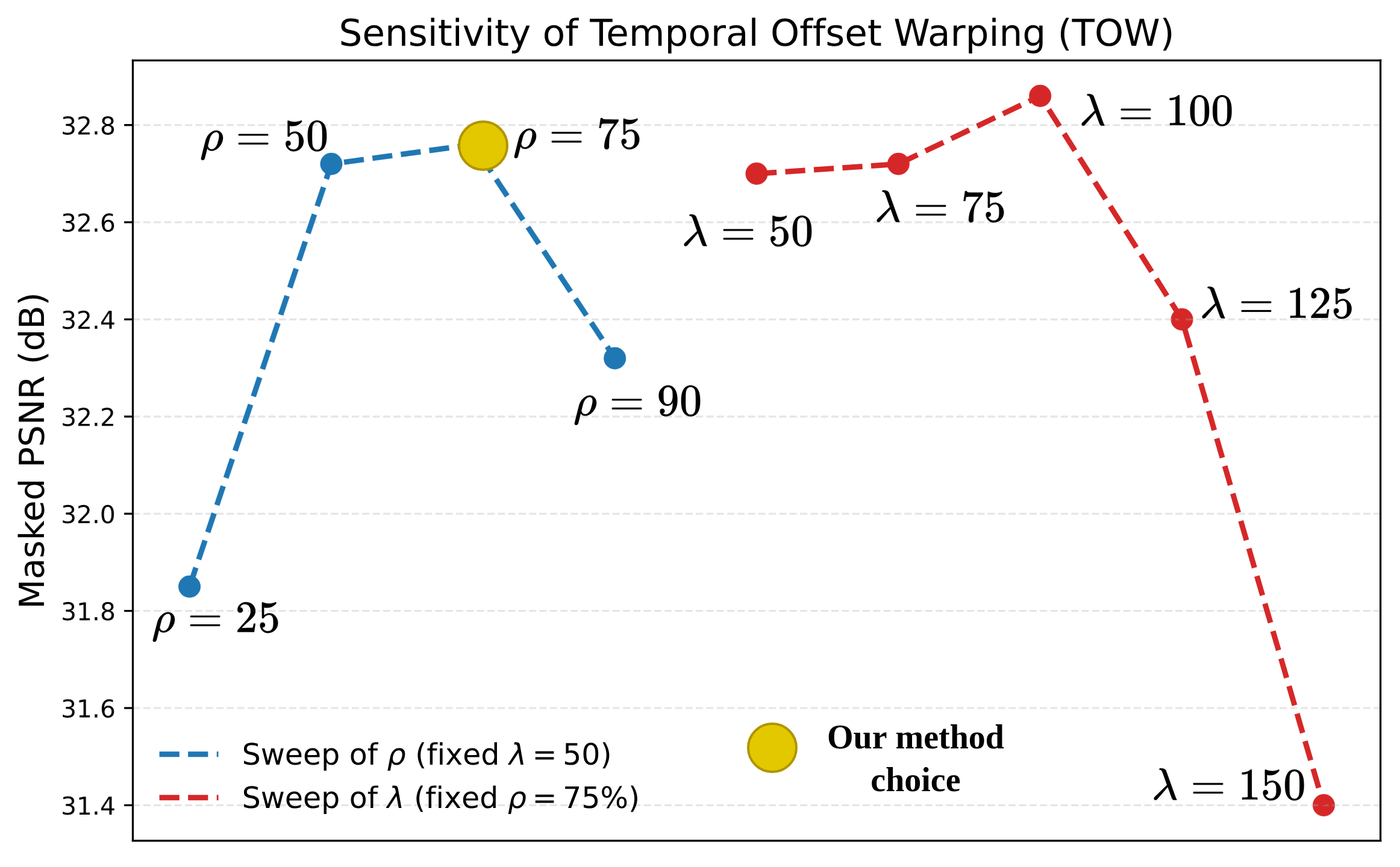}
    \vspace{1mm}
    \caption{\textbf{TOW hyperparameter sensitivity on the \textit{Painter} scene.}
    Moderate temporal capacity ($\rho$) and focus-window sizes ($\lambda$) yield the best Masked PSNR, while extreme settings degrade performance. For ease of interpretation, $\rho$ is reported as a percentage and $\lambda$ in number of frames. Our model adopts $\rho=75\%$ and $\lambda=50$, positioned near the optimal region.}
    \label{fig:tow_ablation}
\end{minipage}
\hfill
\begin{minipage}{0.48\textwidth}
    \centering
    \vspace{-2mm}
    \includegraphics[width=\linewidth]{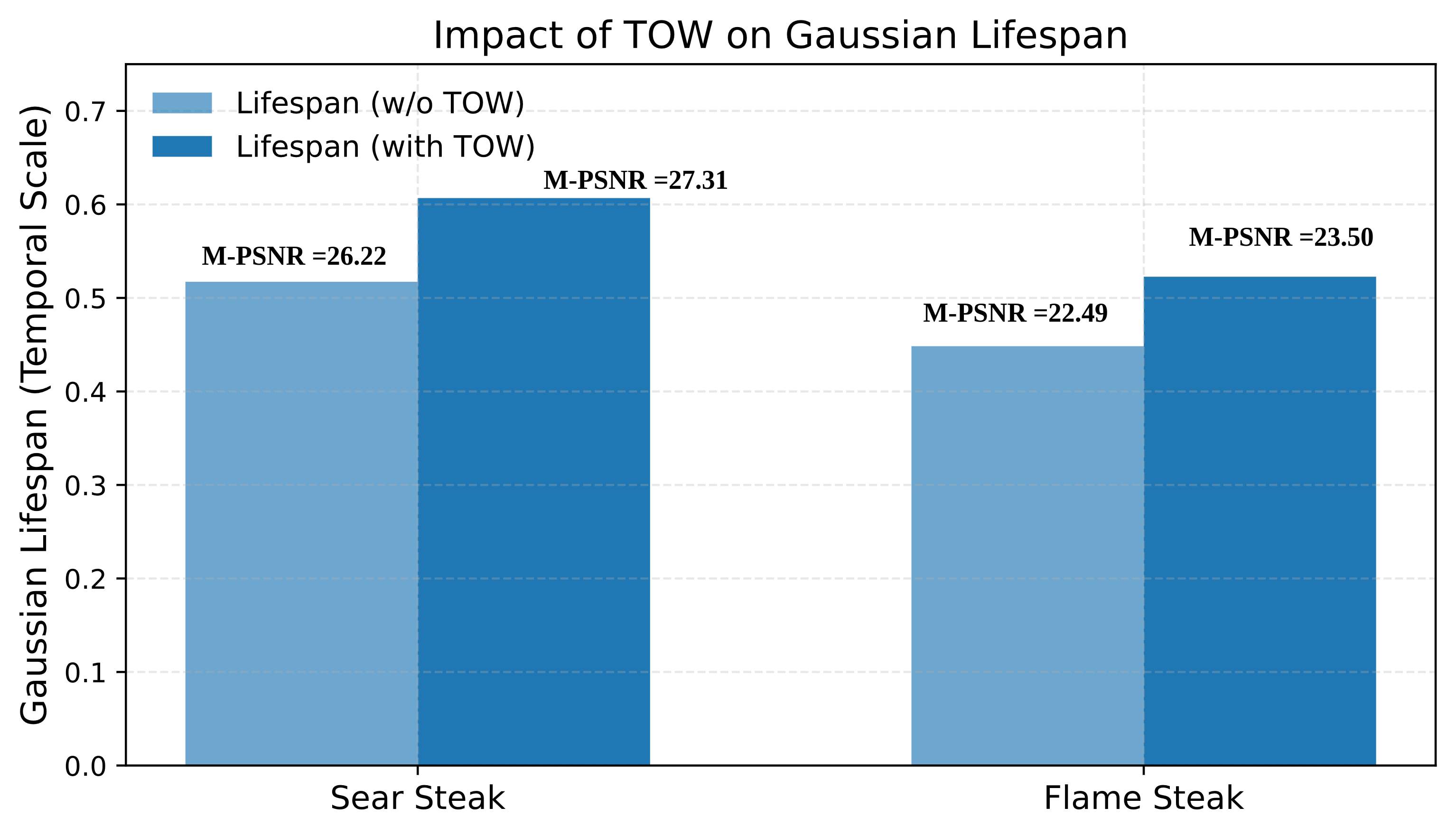}
    \vspace{1mm}
    \caption{\textbf{Impact of TOW on Gaussian lifespan and Model performance.}
    TOW increases the temporal lifespan of Gaussians, allowing them to remain active over longer motion ranges. This extended lifespan enables more effective densification in highly dynamic regions, ultimately yielding higher Masked-PSNR as shown in both sear steak and flame steak scenes.}
    \label{fig:tow_lifespan}
\end{minipage}

\vspace{-2mm}
\end{figure}

\noindent\textbf{Ablation on TOW hyperparameters:}
Figure~\ref{fig:tow_ablation} shows that TOW remains stable under moderate variations of its hyperparameters, with optimal reconstruction achieved when the temporal capacity is set near $75\%$ (or $\rho=0.75$) and the focus window size around $100$ frames (or $\lambda=100/T$). Beyond these ranges, performance gradually degrades, indicating that excessively narrow or overly diffuse temporal influence weakens the effectiveness of temporal offset warping. In our method, we adopt $\rho=0.75$ and $\lambda=50/T$, which lie in the stable, high-performance region of this sweep. For ease of interpretation, $\rho$ is reported as a percentage and $\lambda$ in number of frames in Figure~\ref{fig:tow_ablation}.

\noindent\textbf{Impact of TOW on Gaussian's lifespan:} We represent a Gaussian's lifespan using its temporal scale, a normalized value in $[0,1]$ indicating how much of the sequence it stays active ($1$ means it spans the full 300 frames). In Figure~\ref{fig:tow_lifespan}, we show the average temporal scale across all Gaussians for the \textit{sear steak} and \textit{flame steak} scenes. With TOW, Gaussians live longer and accumulate more stable gradients, leading to frequent densification and better performance especially in highly dynamic regions.

\noindent\textbf{Comparison with 50-frame segment training setup:}
A few works such as STG and Ex4DGS adopt a short-segment protocol, training separate models on 50-frame chunks and aggregating results over the full 300-frame sequence. For completeness, we benchmark our method against this setting as shown in Table~\ref{tab:shortsegment_merged}. However, such segmented training is suboptimal for dynamic scenes, as it increases overall training time, model overhead and often introduces temporal inconsistencies or flicker at segment boundaries due to the lack of long-range temporal coupling.

    \begin{table}[t]
    \centering
    \caption{\textbf{Short-segment (50-frame) training comparison on Interdigital and N3DV datasets.}
    We report PSNR, Masked PSNR (M-PSNR), Masked SSIM (M-SSIM), LPIPS~$\downarrow$, and Training Time (TT, in hours:minutes) on an NVIDIA A4000 16GB GPU.
    \colorbox{red!30}{Red} and \colorbox{orange!30}{Orange} denote the best and second-best results.}

    \vspace{1mm}
    \small
    \setlength{\tabcolsep}{4.5pt}
    \renewcommand{\arraystretch}{1.1}

    \begin{tabular}{lccccc}
    \toprule
    \textbf{Method} &
    PSNR~$\uparrow$ &
    M-PSNR~$\uparrow$ &
    M-SSIM~$\uparrow$ &
    LPIPS~$\downarrow$ &
    TT$\downarrow$ \\
    \midrule

    \multicolumn{6}{c}{\textbf{Interdigital}} \\
    \midrule

    STG~\cite{li2023spacetime}
    & \colorbox{orange!30}{33.80}
    & \colorbox{orange!30}{28.81}
    & \colorbox{orange!30}{0.898}
    & \colorbox{orange!30}{0.044}
    & \colorbox{orange!30}{7:19} \\

    Ex4DGS~\cite{lee2024ex4dgs}
    & 33.64
    & 28.63
    & 0.897
    & 0.052
    & 9:42 \\

    \textbf{Ours-50}
    & \colorbox{red!30}{34.36}
    & \colorbox{red!30}{29.18}
    & \colorbox{red!30}{0.911}
    & \colorbox{red!30}{0.042}
    & \colorbox{red!30}{4:27} \\

    \midrule

    \multicolumn{6}{c}{\textbf{Neural 3D Video}} \\
    \midrule

    STG~\cite{li2023spacetime}
    & \colorbox{orange!30}{32.05}
    & \colorbox{orange!30}{25.11}
    & 0.870
    & \colorbox{orange!30}{0.062}
    & 7:51 \\

    Ex4DGS~\cite{lee2024ex4dgs}
    & 32.01
    & 25.01
    & \colorbox{orange!30}{0.871}
    & 0.064
    & \colorbox{orange!30}{7:19} \\

    \textbf{Ours-50}
    & \colorbox{red!30}{32.51}
    & \colorbox{red!30}{26.03}
    & \colorbox{red!30}{0.889}
    & \colorbox{red!30}{0.057}
    & \colorbox{red!30}{5:17} \\

    \bottomrule
    \end{tabular}

    \label{tab:shortsegment_merged}
    \vspace{-2mm}
    \end{table}

\noindent\textbf{Impact of each Component on Training Time and Model Size:} We analyze how each of our three proposed components contributes to training time and model overhead. Since all three components promote densification in dynamic regions, they naturally introduce additional Gaussians which increases the final model size. This increase in Gaussian count correspondingly raises training time, as more primitives must be processed during the differentiable rendering pass at each iteration. Table~\ref{tab:size_tt} provides a detailed breakdown of how each component incrementally affects model size and training time on the Neural 3D Video dataset.

\begin{table}[h]
    \centering
    \caption{\textbf{Impact of each component on training time and model size.}
    Results are averaged across all scenes in the Neural 3D Video dataset.}
    \begin{tabular}{lcc}
        \toprule
        \textbf{Configuration} & \textbf{Model Size (MB)} & \textbf{Training Time (min)} \\
        \midrule
        Baseline (deformation only) & 155 & 49 mins \\
        + VAD                       & 182 & 55 mins \\
        + VAD + TAT                 & 189 & 57 mins \\
        + VAD + TAT + TOW (Full)    & 204 & 62 mins \\
        \bottomrule
    \end{tabular}
    \label{tab:size_tt}
\end{table}

\begin{figure*}
\centering
\includegraphics[width=0.75\textwidth]{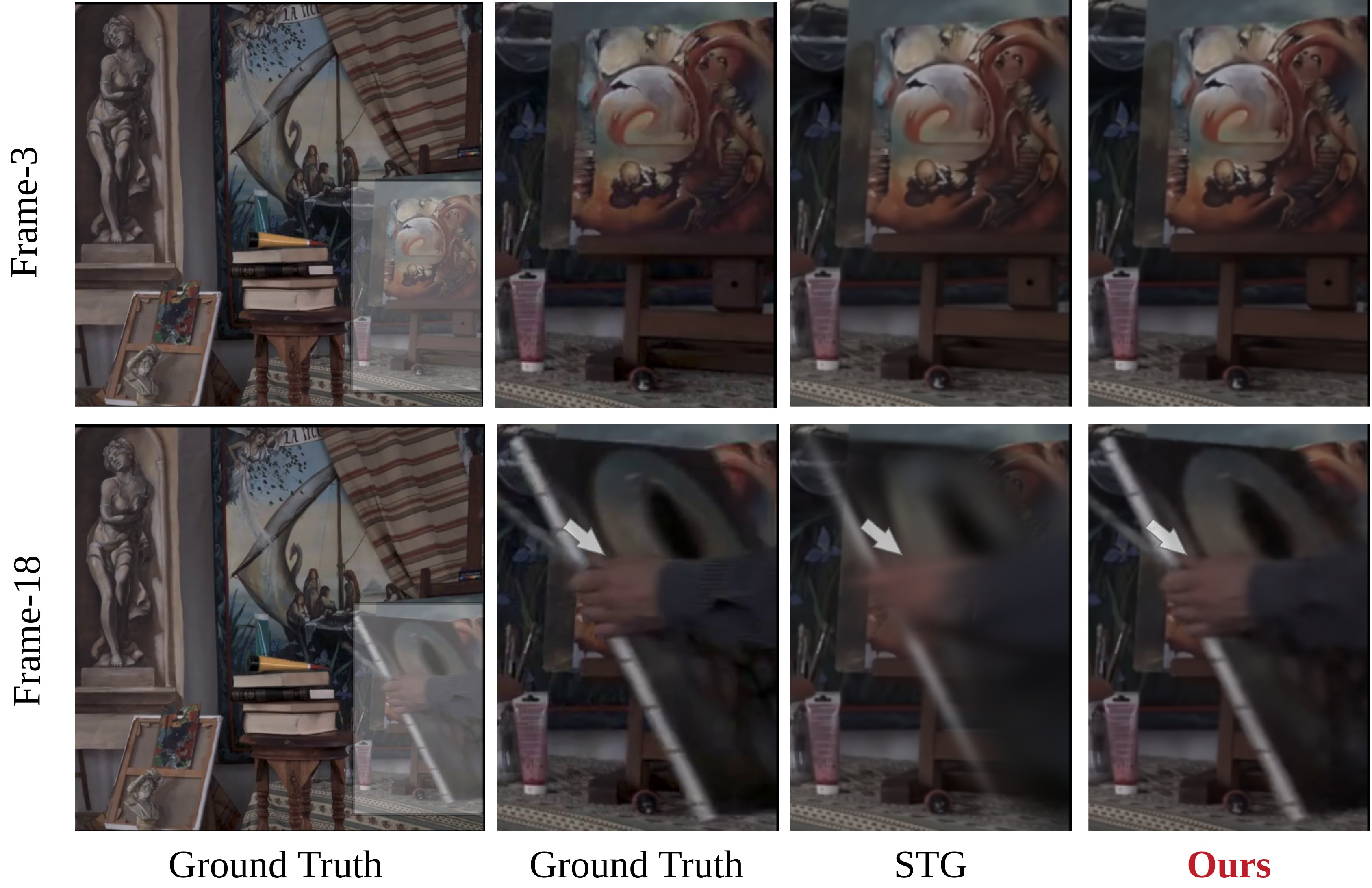}
\caption{\textbf{Sudden Appearance.} Interdigital \textit{Painter} scene.
STG produces blurry artifacts when objects abruptly enter the camera view,
whereas our method preserves finer details in these dynamic regions.}
\label{fig:occ_supp}
\end{figure*}

\noindent\textbf{Sudden Appearance/Occlusion:}
Figure~\ref{fig:occ_supp} illustrates a scenario where an object initially outside the camera view suddenly enters the field of view.
We compare our method with STG~\cite{li2023spacetime} on this challenging case. Such objects typically correspond to short-lived Gaussians due to their limited temporal visibility. Our method effectively handles these situations by enabling proper densification of these short-lived Gaussians, leading to better reconstruction quality.

\begin{figure*}
\centering
\includegraphics[width=0.9\textwidth]{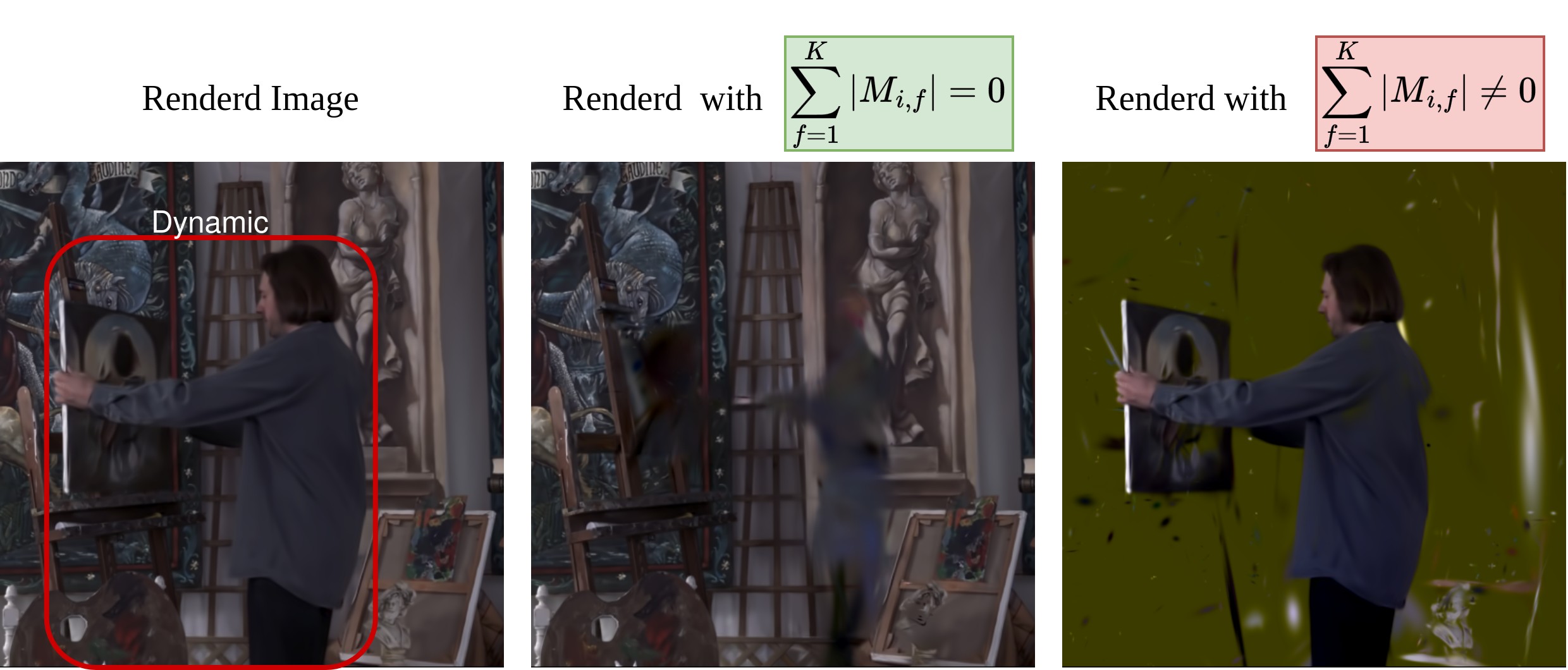}
\caption{
Visualization of Gaussians with zero and non-zero deformation coefficients. Gaussians with $\mathbf{M}_{i,f}=0$ predominantly correspond to static regions.
}
\label{fig:static_dynamic}
\end{figure*}

\noindent\textbf{Impact of TOW on static Gaussians:}
For static Gaussians, the model naturally learns near-zero deformation coefficients in Eqs.~(4)--(6) of the main paper, including the Fourier motion coefficients $\mathbf{M}_{i,f}$. The Gaussian parameters become effectively time-invariant, e.g., $\mu_i(t) \approx \mu_i$ in Eq.~(4), regardless of the warped temporal coordinate induced by TOW. Thus, TOW has negligible impact on static Gaussians and does not introduce artifacts or flicker in static regions, as also observed in the supplementary videos. To validate, we decompose the scene into Gaussians with zero and non-zero deformation coefficients in Figure~\ref{fig:static_dynamic}.

\noindent\textbf{Runtime Analysis:}
The total rendering time consists of preprocessing time for deforming Gaussians and rasterization time. We observe that preprocessing typically dominates the runtime. Methods such as 4DGaussian, Swift4D, and STG employ computationally heavy HexPlane and/or MLP-based deformation fields, leading to high preprocessing overhead. In contrast, our method uses lightweight Fourier- and polynomial-based deformations, substantially reducing preprocessing time. As a result, despite using more Gaussians, our method achieves significantly higher FPS. Table~\ref{tab:runtime_breakdown} provides a runtime breakdown on a 16 GB Nvidia A4000 GPU.

\begin{table}[h]
    \centering
    \footnotesize
    \setlength{\tabcolsep}{5pt}
    \renewcommand{\arraystretch}{1.05}
    \caption{Runtime breakdown on the \emph{Sear Steak} scene from N3DV.}
    \label{tab:runtime_breakdown}
    \begin{tabular}{lcccc}
    \toprule
    \textbf{Method} & \textbf{\# Gaussians} & \textbf{Preprocess (s)} & \textbf{Rasterize (s)} & \textbf{FPS}$\uparrow$ \\
    \midrule
    4DGaussian
    & \cellcolor{lightblue}{$\sim$1.1M}
    & 0.0150
    & \cellcolor{lightblue}{0.0002}
    & 66 \\

    \textbf{Ours}
    & $\sim$4.5M
    & \cellcolor{lightblue}{0.0030}
    & 0.0040
    & \cellcolor{lightblue}{143} \\
    \bottomrule
    \end{tabular}
\end{table}

\begin{figure}[h]
    \centering
    \includegraphics[width=\columnwidth]{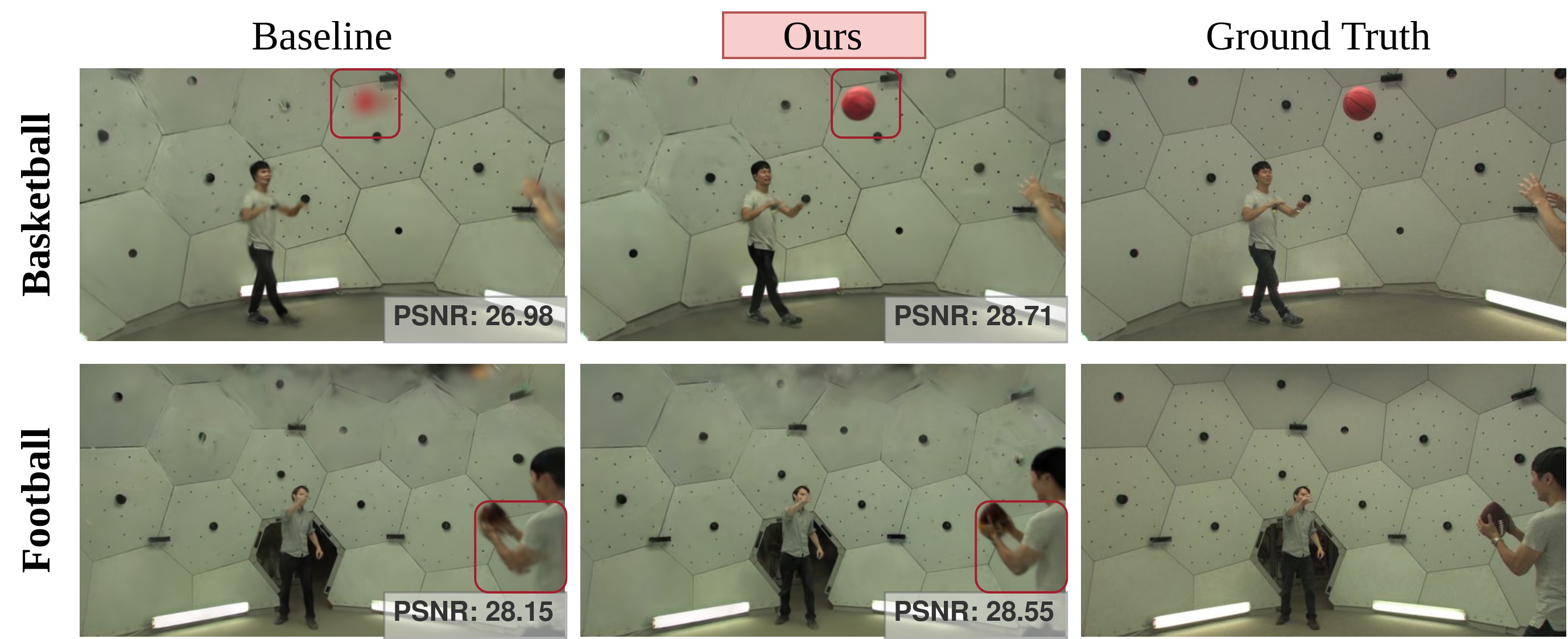}
    \caption{
    Qualitative comparison on the Panoptic Sports dataset. Our method better reconstructs fast-moving objects shown in \textcolor{red}{red} boxes and improves the overall PSNR.
    }
    \label{fig:panotic}
    \vspace{-3mm}
\end{figure}

 \noindent\textbf{Experiments on the Panoptic Sports Dataset:}
 We compare our method against the baseline on the Basketball and Football scenes in Fig.~\ref{fig:panotic}.

\noindent\textbf{Hyperparameter Analysis of TAT and TOW:}
We present a hyperparameter analysis of the TOW and TAT modules in Table~\ref{tab:hyp_ablation} on the Interdigital dataset.

\begin{table}[h]
    \centering
    \footnotesize
    \setlength{\tabcolsep}{4pt}
    \renewcommand{\arraystretch}{1.0}

    \begin{tabular}{c|cccc|cccc}
    \toprule
    \multicolumn{9}{c}{\textbf{TOW Hyperparameters}} \\
    \midrule

    \multicolumn{5}{c|}{\textbf{Temporal capacity $\rho$} ($\lambda = 0.15$)}
    & \multicolumn{4}{c}{\textbf{Focus window size $\lambda$} ($\rho = 0.75$)} \\

    \midrule

    $\rho / \lambda$
    & 0.25 & 0.50 & \cellcolor{lightblue}0.75 & 1.00
    & 0.25 & 0.33 & 0.42 & 0.50 \\

    M-PSNR
    & 28.5 & 28.7 & \cellcolor{lightblue}28.9 & 28.6
    & 28.9 & 29.0 & 28.7 & 28.4 \\

    \bottomrule
    \end{tabular}

    \vspace{0.3em}

    \begin{tabular}{c|ccccc}
    \toprule
    \multicolumn{6}{c}{\textbf{Sensitivity parameter $\beta$ (TAT)}} \\
    \midrule
    $\beta$ & 0.1 & \cellcolor{lightblue}0.3 & 0.5 & 0.7 & 0.9 \\
    M-PSNR & 28.6 & \cellcolor{lightblue}28.9 & 29.1 & 28.8 & 28.8 \\
    \bottomrule
    \end{tabular}
    \vspace{2mm}
    \caption{The performance with hyperparameters is stable. Our configurations are highlighted in light blue.}
    \label{tab:hyp_ablation}
\end{table}

\section{Qualitative Comparisons}

\begin{figure*}
    \captionsetup[subfigure]{labelformat=empty}
    \centering
    \begin{tabular}{@{}c@{\hspace{4pt}}c@{\hspace{4pt}}c@{\hspace{4pt}}c@{}}
        \subfloat[STG]{\includegraphics[width=0.18\textwidth, keepaspectratio]{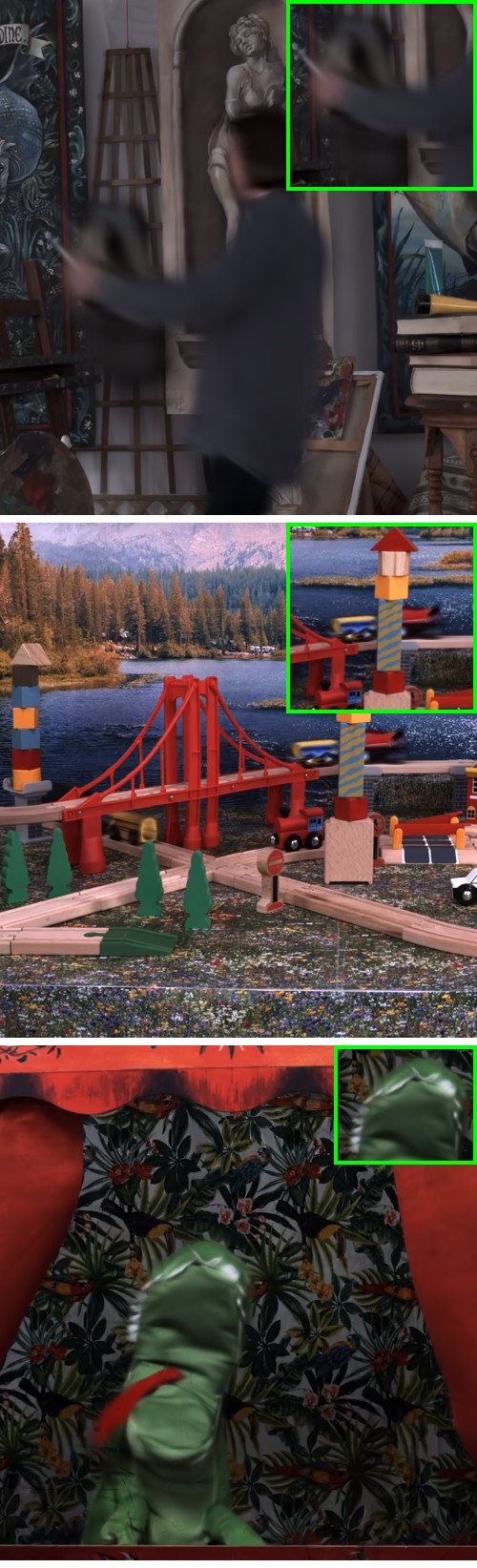}} &
        \subfloat[Ex4DGS]{\includegraphics[width=0.18\textwidth, keepaspectratio]{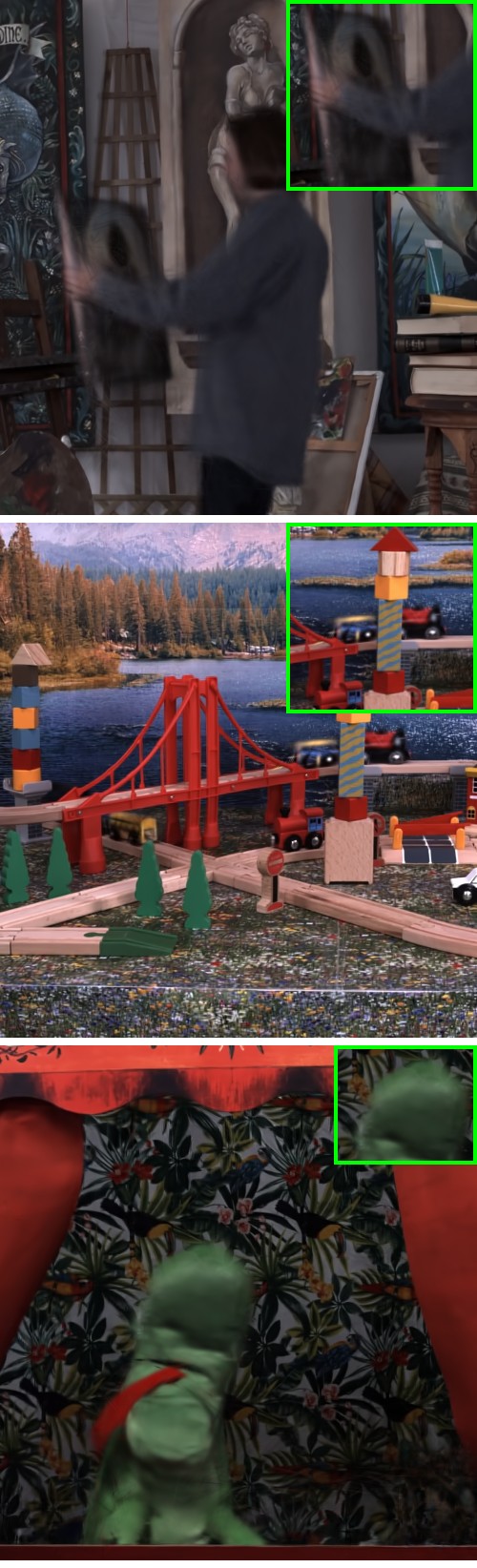}} &
        \subfloat[\textbf{\textcolor{red}{Ours}}]{\includegraphics[width=0.18\textwidth, keepaspectratio]{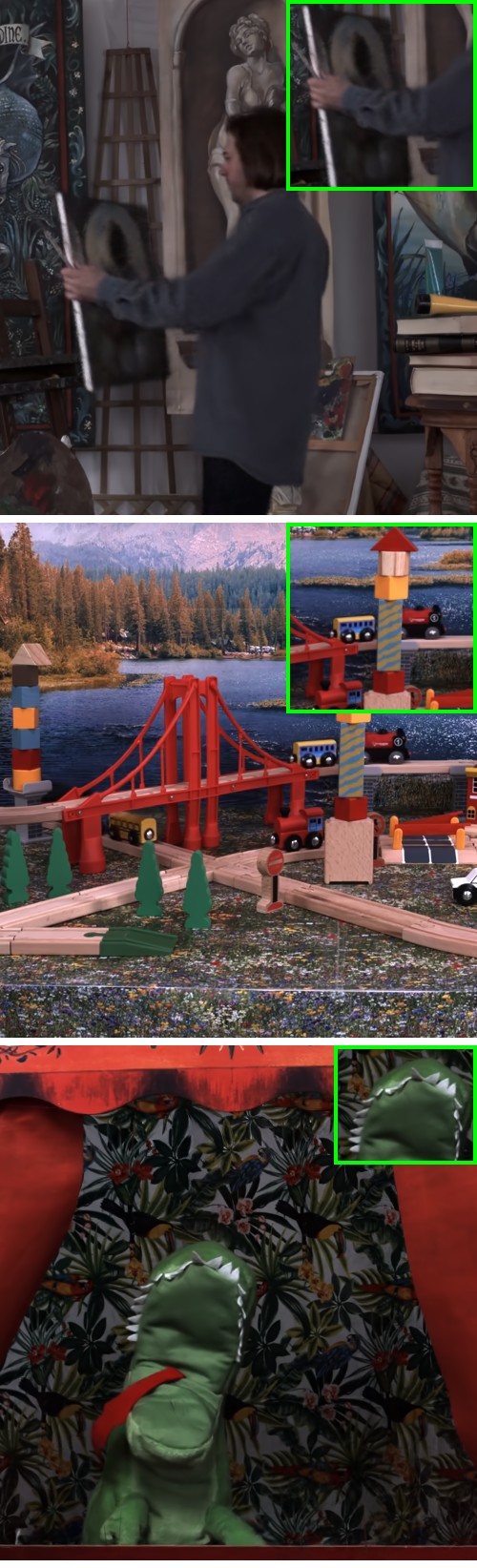}} &
        \subfloat[Ground truth]{\includegraphics[width=0.18\textwidth, keepaspectratio]{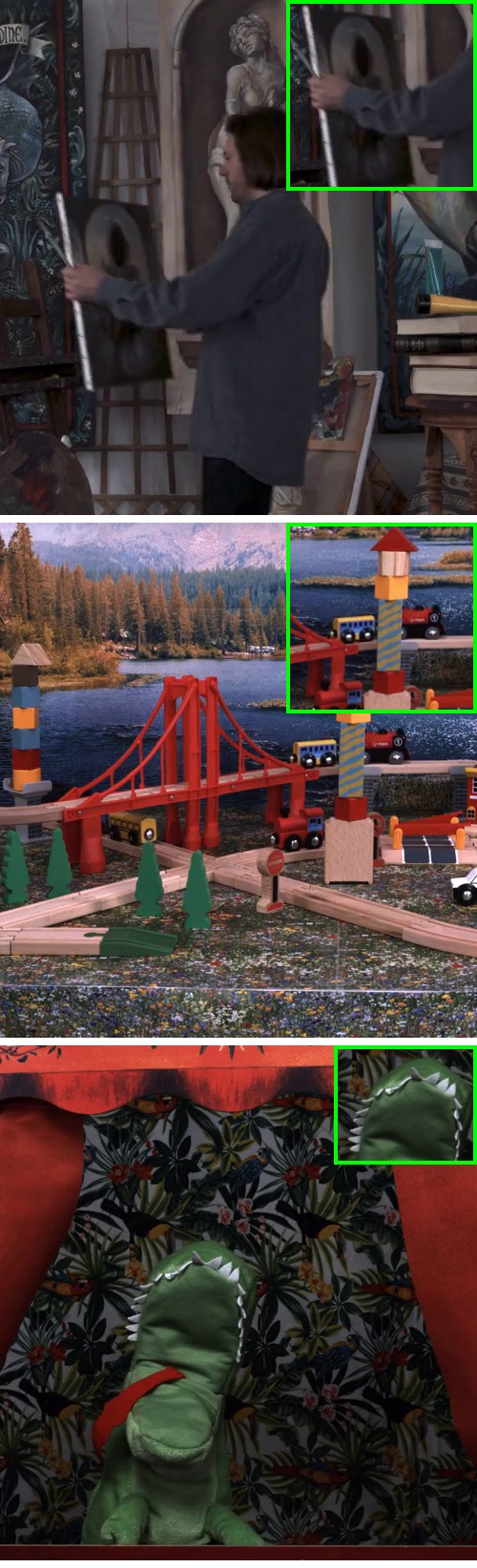}}
    \end{tabular}
    \caption{More qualitative results on the Interdigital Dataset~\cite{Sabater2017}. Scenes top to bottom: i) \textit{Painter}, ii) \textit{Train}, iii) \textit{Theater}. Please note the zoom-in crop of the dynamic regions shown in \textcolor{green}{green} box.}
    \label{fig:interdigital_figure_only}
    \end{figure*}

\begin{figure*}
        \captionsetup[subfigure]{labelformat=empty}
        \centering
        \begin{tabular}{@{}c@{\hspace{2pt}}c@{\hspace{2pt}}c@{\hspace{2pt}}c@{\hspace{2pt}}c@{}}
        \subfloat[\textbf{\textcolor{red}{Ours}}]
            {\includegraphics[width=0.18\linewidth]{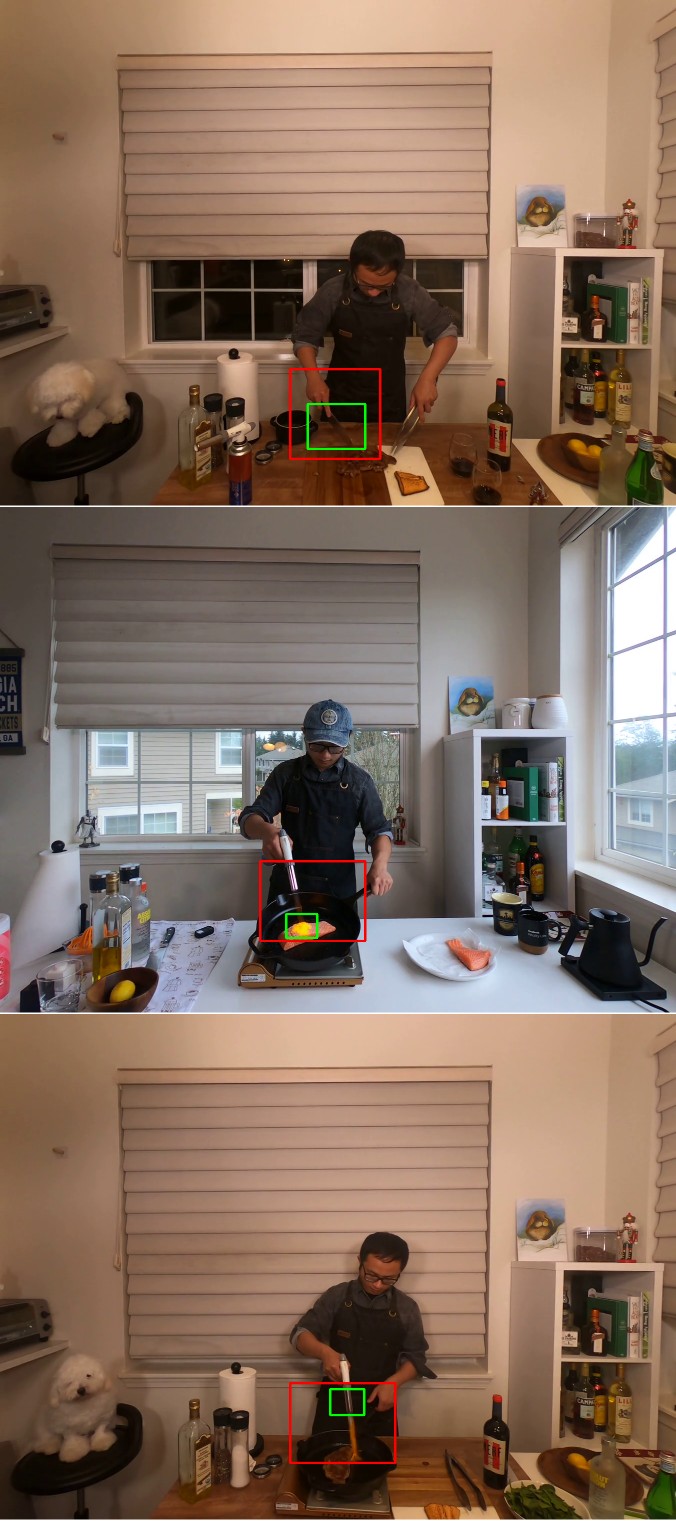}} &
        \subfloat[4DGaussian]
            {\includegraphics[width=0.18\linewidth]{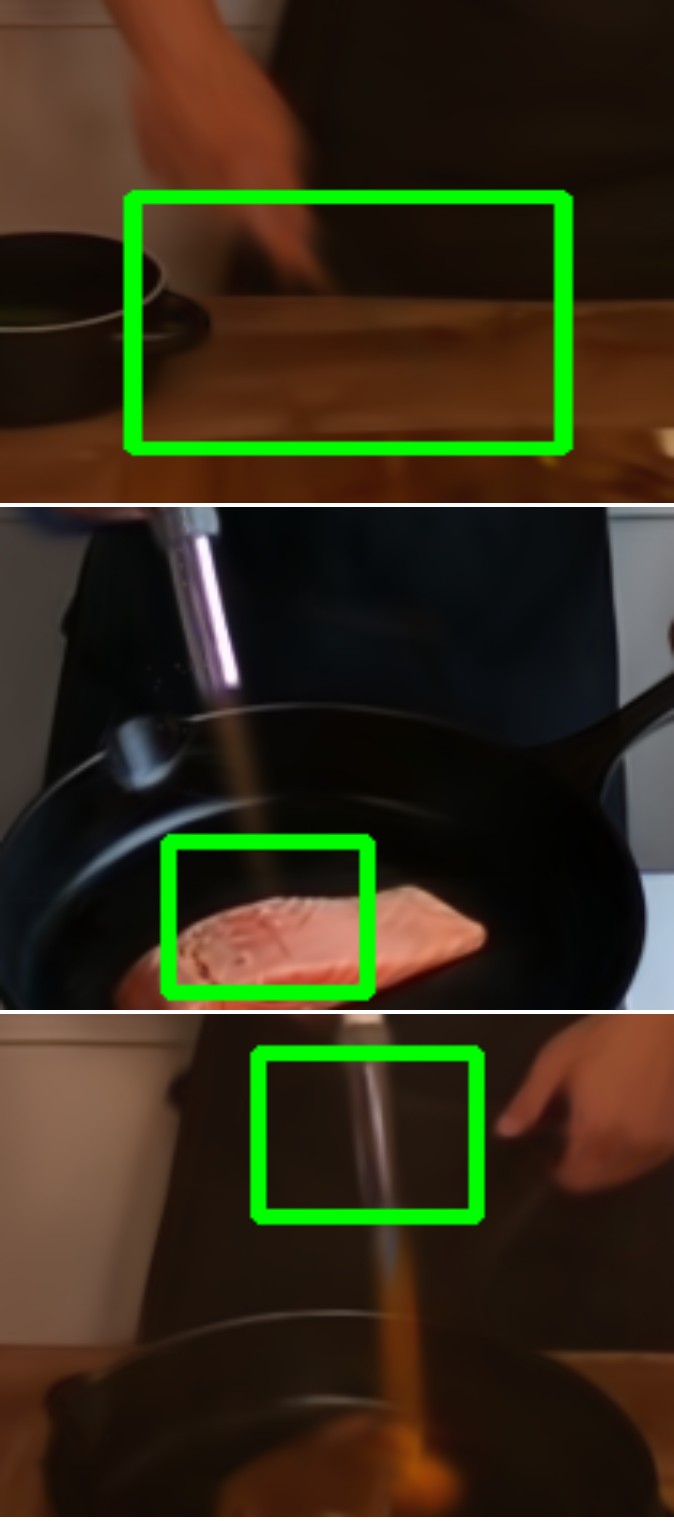}} &
        \subfloat[SaroGS]
            {\includegraphics[width=0.18\linewidth]{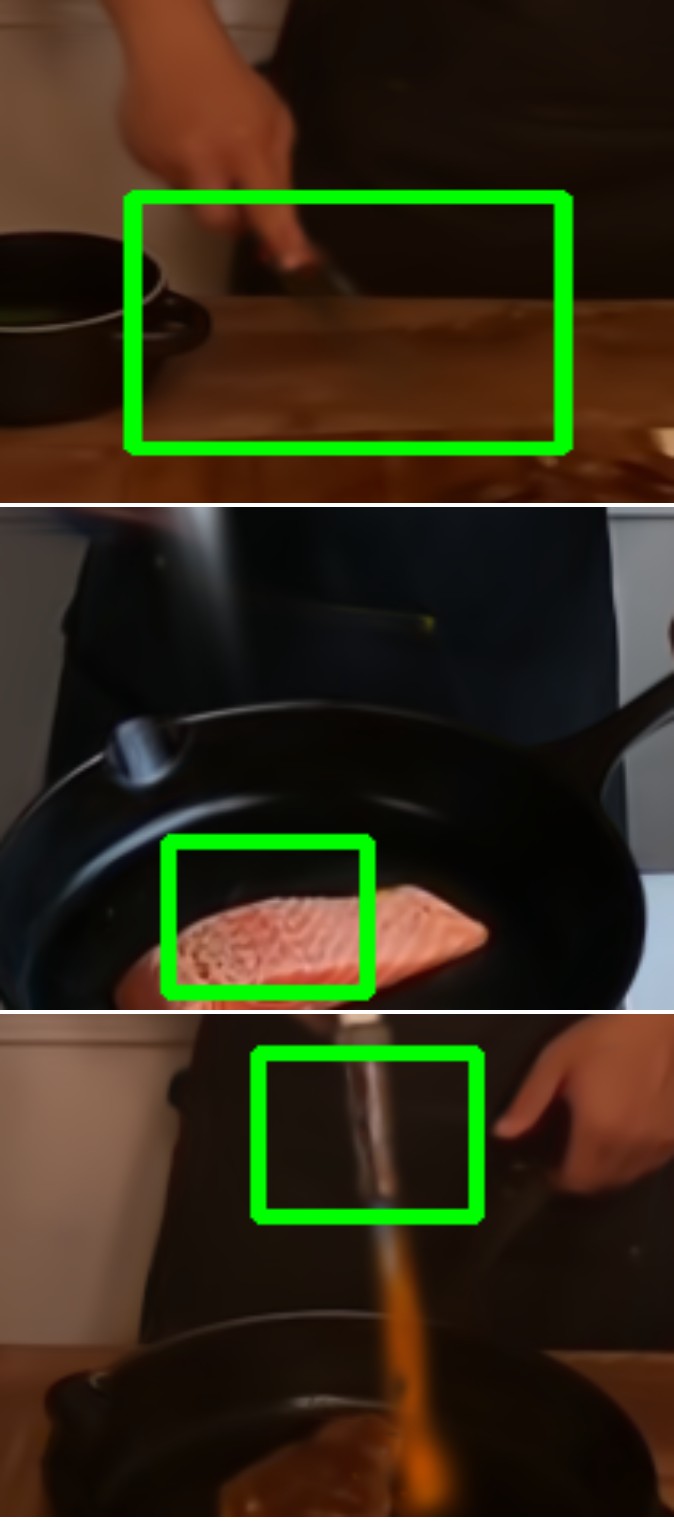}} &
        \subfloat[\textbf{\textcolor{red}{Ours}}]
            {\includegraphics[width=0.18\linewidth]{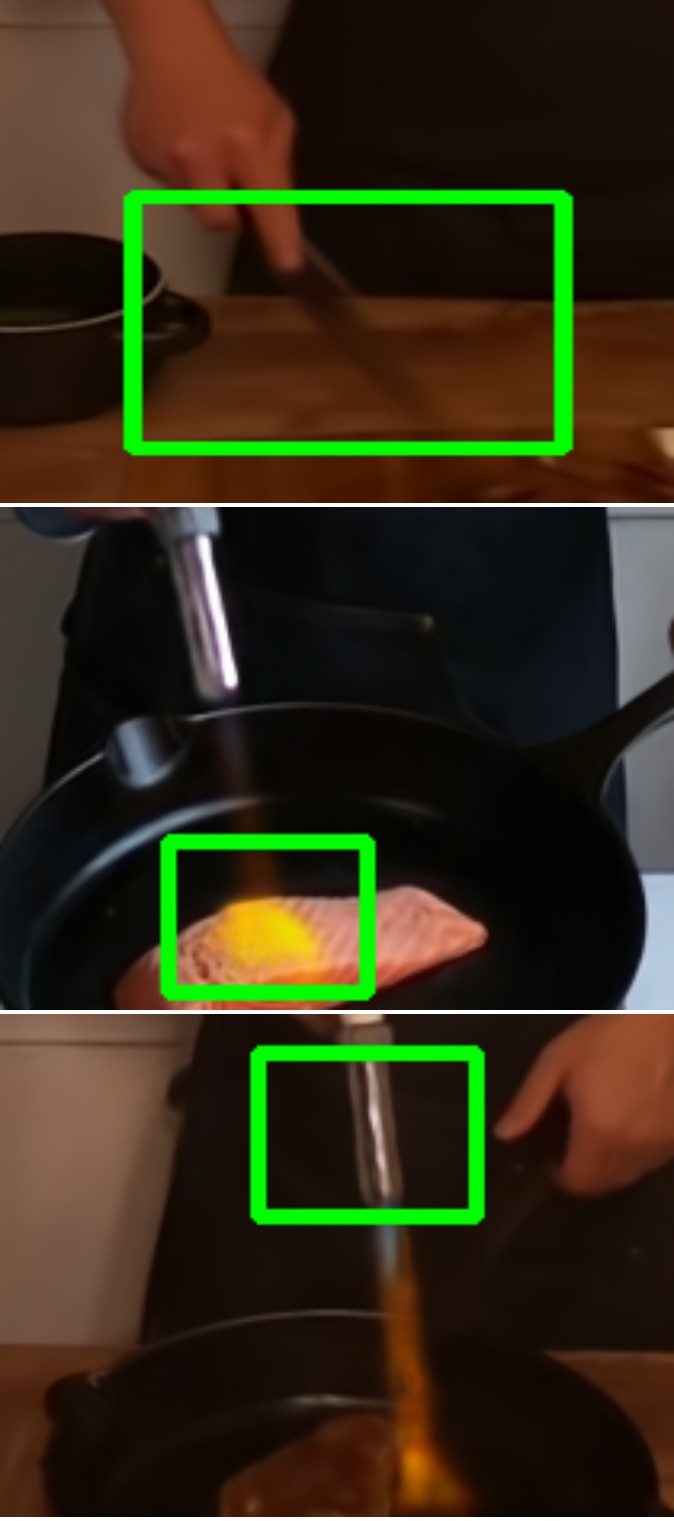}} &
        \subfloat[Ground truth]
            {\includegraphics[width=0.18\linewidth]{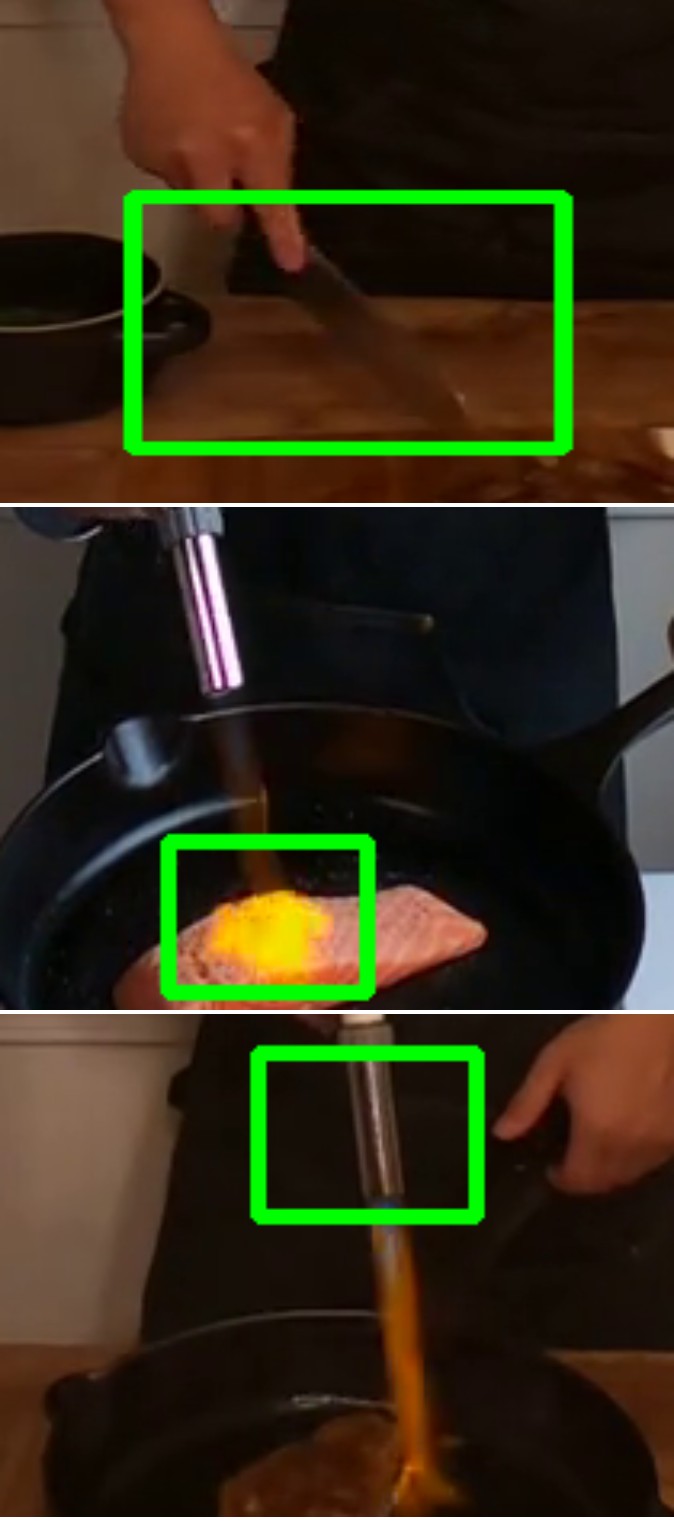}}
        \end{tabular}
        \caption{More qualitative comparison on Neural 3D Video~\cite{li2022neural}. Scene: i) \textit{cut roasted beef} ii) \textit{flame salmon} iii) \textit{flame steak}}
        \label{fig:n3d_supp_figure}
    \end{figure*}

\begin{figure*}
    \captionsetup[subfigure]{labelformat=empty}
    \centering

    \resizebox{0.95\textwidth}{!}{
    \begin{tabular}{@{}c@{\hspace{1pt}}c@{\hspace{1pt}}c@{\hspace{1pt}}c@{\hspace{1pt}}c@{\hspace{1pt}}c@{}}

    \subfloat[\textbf{\textcolor{red}{Full}}]{%
    \includegraphics[height=6.0cm,keepaspectratio]{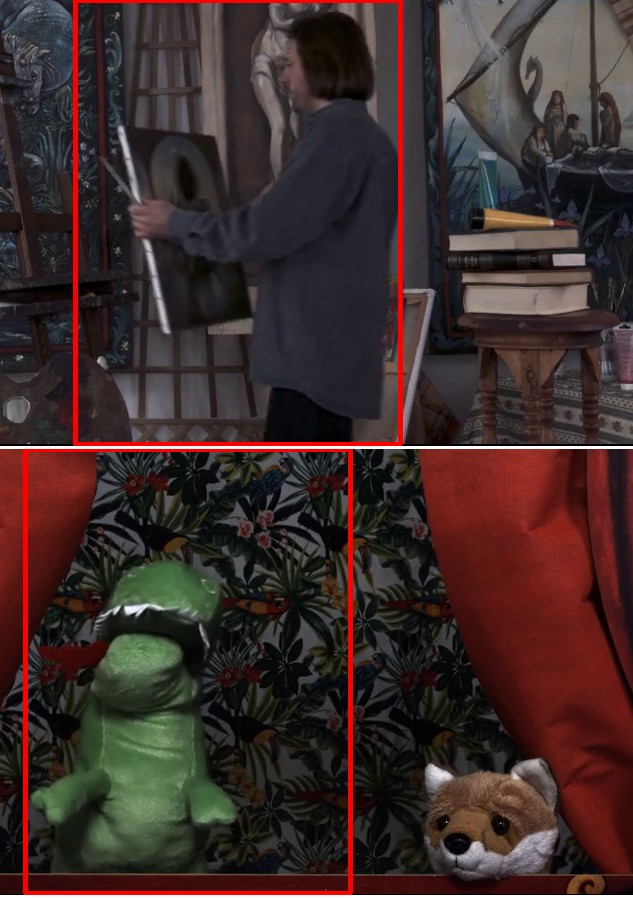}
    } &

    \subfloat[Baseline]{%
    \includegraphics[height=6.0cm,keepaspectratio]{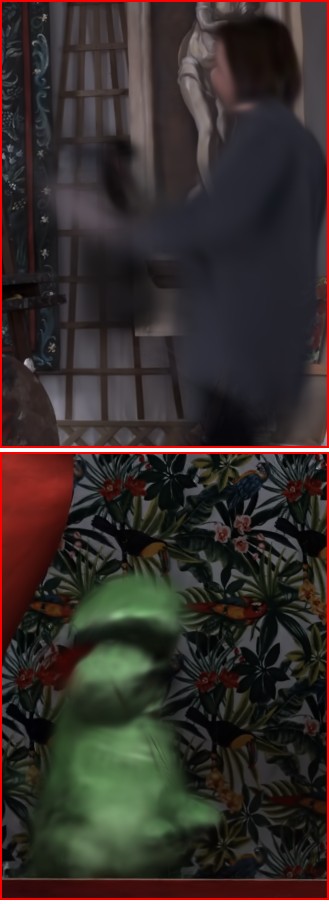}
    } &

    \subfloat[VAD]{%
    \includegraphics[height=6.0cm,keepaspectratio]{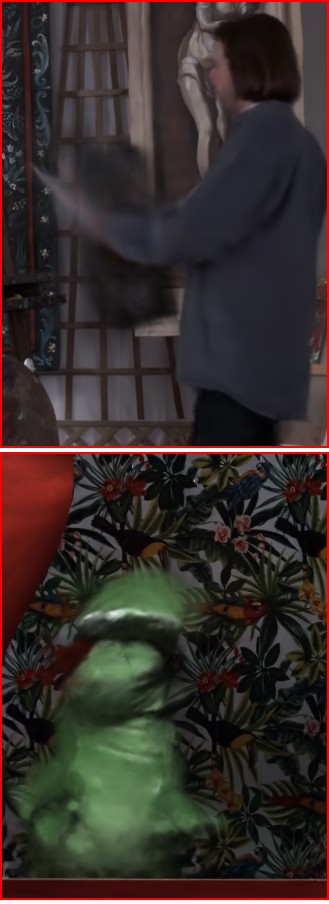}
    } &

    \subfloat[VAD + TAT]{%
    \includegraphics[height=6.0cm,keepaspectratio]{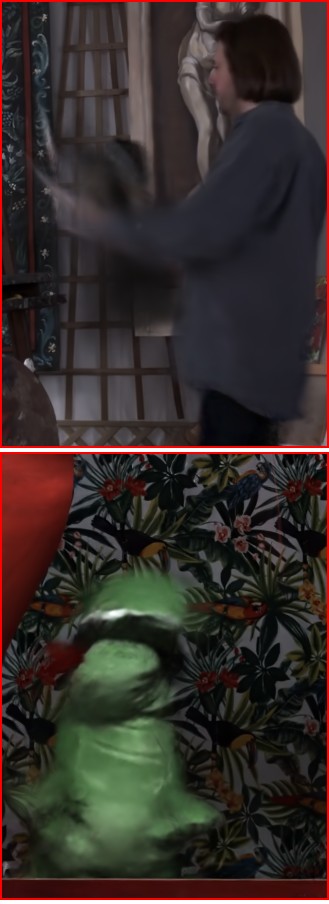}
    } &

    \subfloat[\textbf{\textcolor{red}{Full}}]{%
    \includegraphics[height=6.0cm,keepaspectratio]{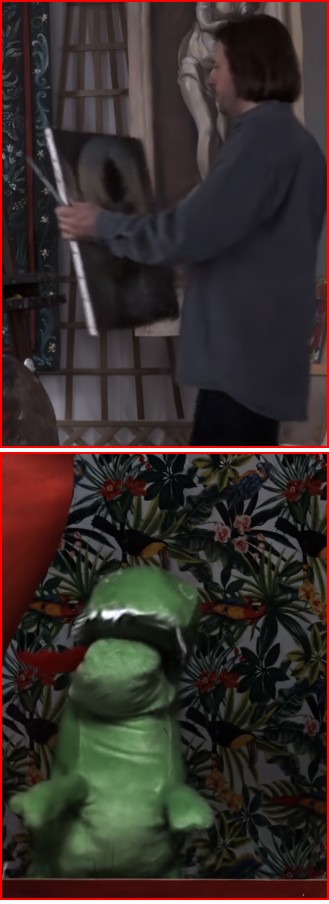}
    } &

    \subfloat[Ground truth]{%
    \includegraphics[height=6.0cm,keepaspectratio]{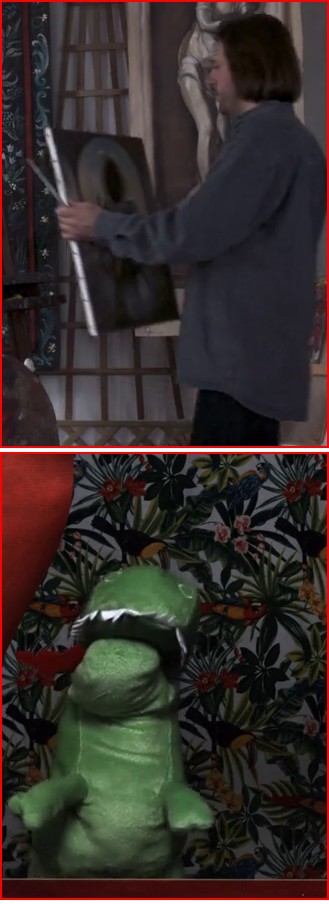}
    }

    \end{tabular}
    }

    \caption{Qualitative ablation from the baseline to the full model with the addition of VAD, TAT, and TOW. Each component improves reconstruction in dynamic regions. Scenes shown top to bottom: i) \textit{Painter}, ii) \textit{Theater}.}
    \label{fig:ablation-1-figonly}

    \end{figure*}

    We present additional qualitative results on the Interdigital and N3DV datasets.
    Figure~\ref{fig:interdigital_figure_only} shows scene-wise qualitative comparisons on the Interdigital dataset, highlighting zoomed-in crops of dynamic regions indicated by the green boxes. Figure~\ref{fig:n3d_supp_figure} shows qualitative results on the \textit{cut-roasted-beef}, \textit{flame-salmon}, and \textit{flame-steak} scenes from the N3DV dataset for the 4DGaussian and SaroGS~\cite{yan20244d} methods.
    Furthermore, Figure~\ref{fig:ablation-1-figonly} illustrates the impact of each of the three proposed components on dynamic scene reconstruction in the Interdigital dataset.

\section{Video Comparisons}

    We present video comparisons across the Interdigital \cite{Sabater2017}, Neural 3D Video \cite{li2020neural}, and VRU Basketball \cite{wu2025swift4d} datasets for all evaluated methods. On Interdigital, we compare our approach with Swift4D on the Train scene and with STG on Painter. For Neural 3D Video, we benchmark against SaroGS on flame-salmon, 4DGaussian on sear-steak, and Swift4D on flame-steak. For VRU Basketball, we compare our method with STG on the DG scene. We additionally include visualizations that highlight how the proposed Visibility-Aware Densification (VAD) module generalizes when integrated into other baselines. All videos are encoded using the H.264 codec, with YUV420p as the pixel format, and a frame rate of 30 FPS.

\section{Scene-wise Quantitative Comparisons}

    This section reports scene-wise metrics---PSNR, SSIM \cite{1284395}, M-PSNR, M-SSIM, LPIPS \cite{zhang2018perceptual} for all methods across all datasets. We observe that our method consistently achieves better performance across most scenes on all 3 datasets as shown in Table~\ref{tab:interdigital_perscene}, Table~\ref{tab:n3d_perscene}, and Table~\ref{tab:vru_perscene}.

    \begin{table*}
    \caption{
    \textbf{Per-scene quantitative comparisons on the Interdigital Dataset~\cite{Sabater2017}. All models are trained and evaluated for 300 frames.}
    }
    \label{tab:interdigital_perscene}
    \centering
    \resizebox{0.75\linewidth}{!}{%
    \begin{tabular}{l|c|ccccc}
    \toprule
    Method & Average & \textit{Birthday} & \textit{Painter} & \textit{Remy} & \textit{Theater} & \textit{Train} \\
    \midrule

    \multicolumn{7}{c}{\textbf{PSNR$\uparrow$}} \\
    \midrule
    4DGaussian      & 26.74 & 22.96 & 30.09 & 32.48 & 23.40 & 24.79 \\
    STG             & 33.45 & 31.88 & 36.78 & 36.12 & 30.98 & 31.54 \\
    Swift4D         & 31.63 & 29.28 & 34.33 & 35.67 & 29.06 & 29.79 \\
    Ex-4DGS         & 32.45 & 29.56 & 35.86 & 35.57 & 30.88 & 30.35 \\
    Ours            & 34.14 & 32.07 & 37.90 & 36.73 & 31.36 & 32.66 \\
    \midrule

    \multicolumn{7}{c}{\textbf{M-PSNR$\uparrow$}} \\
    \midrule
    4DGaussian      & 18.09 & 17.50 & 19.62 & 24.30 & 15.53 & 13.50 \\
    STG             & 27.14 & 26.60 & 30.24 & 28.75 & 24.07 & 26.04 \\
    Swift4D         & 23.01 & 22.92 & 23.60 & 29.08 & 21.50 & 17.93 \\
    Ex-4DGS         & 26.15 & 24.54 & 28.03 & 29.04 & 25.06 & 24.09 \\
    Ours            & 28.87 & 26.99 & 32.72 & 29.96 & 26.79 & 27.88 \\
    \midrule

    \multicolumn{7}{c}{\textbf{SSIM$\uparrow$}} \\
    \midrule
    4DGaussian      & 0.833 & 0.763 & 0.893 & 0.905 & 0.795 & 0.810 \\
    STG             & 0.938 & 0.944 & 0.962 & 0.942 & 0.906 & 0.937 \\
    Swift4D         & 0.930 & 0.925 & 0.951 & 0.941 & 0.897 & 0.935 \\
    Ex-4DGS         & 0.930 & 0.924 & 0.956 & 0.943 & 0.909 & 0.918 \\
    Ours            & 0.946 & 0.947 & 0.967 & 0.954 & 0.917 & 0.948 \\
    \midrule

    \multicolumn{7}{c}{\textbf{M-SSIM$\uparrow$}} \\
    \midrule
    4DGaussian      & 0.520 & 0.569 & 0.563 & 0.685 & 0.472 & 0.311 \\
    STG             & 0.860 & 0.916 & 0.901 & 0.833 & 0.787 & 0.877 \\
    Swift4D         & 0.741 & 0.838 & 0.720 & 0.845 & 0.718 & 0.586 \\
    Ex-4DGS         & 0.844 & 0.868 & 0.860 & 0.849 & 0.826 & 0.818 \\
    Ours            & 0.901 & 0.923 & 0.930 & 0.892 & 0.850 & 0.900 \\
    \midrule

    \multicolumn{7}{c}{\textbf{LPIPS$_{Alex}$$\downarrow$}} \\
    \midrule
    4DGaussian      & 0.173 & 0.209 & 0.134 & 0.168 & 0.189 & 0.163 \\
    STG             & 0.060 & 0.026 & 0.043 & 0.094 & 0.090 & 0.029 \\
    Swift4D         & 0.072 & 0.048 & 0.073 & 0.096 & 0.106 & 0.038 \\
    Ex-4DGS         & 0.070 & 0.052 & 0.059 & 0.084 & 0.087 & 0.066 \\
    Ours            & 0.044 & 0.024 & 0.037 & 0.068 & 0.070 & 0.021 \\

    \bottomrule
    \end{tabular}
    }
    \end{table*}

    \begin{table*}
    \caption{
    \textbf{Per-scene quantitative comparisons on the Neural 3D Video Dataset~\cite{li2022neural}. All models are trained and evaluated for 300 frames.}
    }
    \label{tab:n3d_perscene}
    \centering
    \resizebox{1\linewidth}{!}{%
    \begin{tabular}{l|c|cccccc}
    \toprule
    Method & Avg. & \textit{Coffee Martini} & \textit{Cook Spinach} & \textit{Cut Roasted Beef} & \textit{Flame Salmon} & \textit{Flame Steak} & \textit{Sear Steak} \\
    \midrule

    \multicolumn{8}{c}{\textbf{PSNR$\uparrow$}} \\
    \midrule
    4DGaussian & 31.21 & 28.52 & 32.15 & 32.36 & 29.01 & 32.73 & 32.60 \\
    STG        & 31.40 & 27.65 & 32.50 & 33.20 & 28.17 & 33.19 & 33.79 \\
    SaroGS     & 32.08 & 28.86 & 33.12 & 33.83 & 29.06 & 33.78 & 33.80 \\
    Swift4D    & 32.12 & 29.29 & 32.89 & 33.58 & 29.62 & 33.51 & 33.83 \\
    Ex-4DGS    & 31.45 & 27.67 & 32.87 & 33.08 & 28.24 & 33.13 & 33.74 \\
    Ours       & 32.42 & 29.30 & 33.61 & 33.99 & 29.70 & 34.04 & 33.89 \\
    \midrule

    \multicolumn{8}{c}{\textbf{M-PSNR$\uparrow$}} \\
    \midrule
    4DGaussian & 22.66 & 22.46 & 23.51 & 25.29 & 19.44 & 22.35 & 22.95 \\
    STG        & 22.61 & 22.10 & 22.54 & 25.31 & 18.50 & 22.18 & 25.06 \\
    SaroGS     & 23.62 & 23.80 & 24.30 & 26.89 & 19.32 & 23.10 & 24.31 \\
    Swift4D    & 23.74 & 23.54 & 23.09 & 26.45 & 21.03 & 22.19 & 26.24 \\
    Ex-4DGS    & 23.41 & 21.15 & 23.70 & 26.59 & 19.65 & 23.01 & 26.39 \\
    Ours       & 24.68 & 23.67 & 24.60 & 27.64 & 21.36 & 23.50 & 27.31 \\
    \midrule

    \multicolumn{8}{c}{\textbf{SSIM$\uparrow$}} \\
    \midrule
    4DGaussian & 0.942 & 0.920 & 0.946 & 0.951 & 0.924 & 0.957 & 0.954 \\
    STG        & 0.948 & 0.918 & 0.956 & 0.958 & 0.923 & 0.964 & 0.966 \\
    SaroGS     & 0.950 & 0.926 & 0.958 & 0.959 & 0.926 & 0.964 & 0.964 \\
    Swift4D    & 0.944 & 0.917 & 0.950 & 0.953 & 0.922 & 0.958 & 0.960 \\
    Ex-4DGS    & 0.939 & 0.902 & 0.951 & 0.953 & 0.932 & 0.958 & 0.958 \\
    Ours       & 0.955 & 0.930 & 0.962 & 0.964 & 0.932 & 0.968 & 0.969 \\
    \midrule

    \multicolumn{8}{c}{\textbf{M-SSIM$\uparrow$}} \\
    \midrule
    4DGaussian & 0.784 & 0.803 & 0.783 & 0.833 & 0.768 & 0.789 & 0.722 \\
    STG        & 0.792 & 0.814 & 0.767 & 0.825 & 0.732 & 0.786 & 0.825 \\
    SaroGS     & 0.821 & 0.861 & 0.819 & 0.874 & 0.765 & 0.809 & 0.794 \\
    Swift4D    & 0.836 & 0.871 & 0.788 & 0.878 & 0.834 & 0.802 & 0.842 \\
    Ex-4DGS    & 0.815 & 0.761 & 0.817 & 0.876 & 0.740 & 0.823 & 0.857 \\
    Ours       & 0.863 & 0.867 & 0.838 & 0.902 & 0.824 & 0.849 & 0.895 \\
    \midrule

    \multicolumn{8}{c}{\textbf{LPIPS$_{Alex}$$\downarrow$}} \\
    \midrule
    4DGaussian & 0.071 & 0.085 & 0.071 & 0.070 & 0.083 & 0.061 & 0.054 \\
    STG        & 0.069 & 0.085 & 0.068 & 0.068 & 0.084 & 0.057 & 0.052 \\
    SaroGS     & 0.064 & 0.078 & 0.059 & 0.059 & 0.075 & 0.053 & 0.055 \\
    Swift4D    & 0.061 & 0.075 & 0.056 & 0.059 & 0.077 & 0.047 & 0.052 \\
    Ex-4DGS    & 0.079 & 0.111 & 0.071 & 0.069 & 0.098 & 0.062 & 0.061 \\
    Ours       & 0.059 & 0.076 & 0.054 & 0.055 & 0.074 & 0.047 & 0.048 \\
    \bottomrule
    \end{tabular}
    }
    \end{table*}

    \begin{table*}
    \centering
    \caption{
    \textbf{Per-scene quantitative comparisons on the VRU dataset.
    All models are trained and evaluated for 250 frames.}
    }
    \label{tab:vru_perscene}

    \setlength{\tabcolsep}{10pt}
    \renewcommand{\arraystretch}{1.1}

    \begin{tabular}{lccc}
    \toprule
    Method & Avg. & \textit{DG} & \textit{GZ} \\
    \midrule

    \multicolumn{4}{c}{\textbf{PSNR$\uparrow$}} \\
    \midrule
    STG      & 26.03 & 26.46 & 25.61 \\
    Swift4D  & 25.72 & 26.51 & 24.94 \\
    Ours     & 28.28 & 28.13 & 28.44 \\
    \midrule

    \multicolumn{4}{c}{\textbf{M-PSNR$\uparrow$}} \\
    \midrule
    STG      & 21.33 & 22.23 & 20.43 \\
    Swift4D  & 19.52 & 21.42 & 17.63 \\
    Ours     & 25.39 & 25.30 & 25.48 \\
    \midrule

    \multicolumn{4}{c}{\textbf{SSIM$\uparrow$}} \\
    \midrule
    STG      & 0.897 & 0.883 & 0.911 \\
    Swift4D  & 0.892 & 0.884 & 0.900 \\
    Ours     & 0.921 & 0.905 & 0.937 \\
    \midrule

    \multicolumn{4}{c}{\textbf{M-SSIM$\uparrow$}} \\
    \midrule
    STG      & 0.736 & 0.757 & 0.714 \\
    Swift4D  & 0.705 & 0.764 & 0.645 \\
    Ours     & 0.881 & 0.875 & 0.886 \\
    \midrule

    \multicolumn{4}{c}{\textbf{LPIPS$_{Alex}\downarrow$}} \\
    \midrule
    STG      & 0.129 & 0.097 & 0.160 \\
    Swift4D  & 0.127 & 0.121 & 0.134 \\
    Ours     & 0.090 & 0.090 & 0.090 \\

    \bottomrule
    \end{tabular}

    \vspace{-2mm}
    \end{table*}

\end{document}